\documentclass[10pt,journal,compsoc]{IEEEtran}
\usepackage{float}
\usepackage{diagbox}
\usepackage{amsthm}
\newtheorem{theorem}{Theorem}
\newtheorem{lemma}{Lemma}
\usepackage{graphicx}
\usepackage{epstopdf}
\usepackage{amsmath}
\usepackage{amssymb}
\usepackage{multirow}
\usepackage{hyperref}
\usepackage[ruled,linesnumbered]{algorithm2e}
\usepackage{color}

\usepackage{amsthm,amsmath,amssymb}
\usepackage{mathrsfs}
\usepackage{diagbox}
\usepackage{yhmath}
\usepackage [latin1]{inputenc}
\usepackage[table]{xcolor}

\ifCLASSOPTIONcompsoc
  \usepackage[nocompress]{cite}
\else
  \usepackage{cite}
\fi
\ifCLASSINFOpdf
\else
\fi
\begin{document}
%
\title{Gradient Harmonization in Unsupervised Domain Adaptation}
%
%
%
%

\author{Fuxiang Huang, 
        ~Suqi Song,
        ~Lei Zhang

\IEEEcompsocitemizethanks{\IEEEcompsocthanksitem This work was partially supported by National Key R\&D Program of China (2021YFB3100800), National Natural Science Fund of China (62271090), Chongqing Natural Science Fund (cstc2021jcyj-jqX0023). This work is also supported by Huawei computational power of Chongqing Artificial Intelligence Innovation Center. \textit{(Corresponding author: Lei Zhang)}

\IEEEcompsocthanksitem Fuxiang Huang, Suqi Song and Lei Zhang are with the School of Microelectronics and Communication Engineering, Chongqing University, Chongqing 400044, China. (E-mail: huangfuxiang@cqu.edu.cn, songsuqi@stu.cqu.edu.cn, leizhang@cqu.edu.cn).


}

\thanks{Manuscript received April 19, 2005; revised August 26, 2015.
}}

%
%

\markboth{Journal of \LaTeX\ Class Files,~Vol.~14, No.~8, August~2015}%
{Shell \MakeLowercase{\textit{et al.}}: Bare Demo of IEEEtran.cls for Computer Society Journals}
%



\IEEEtitleabstractindextext{%
\begin{abstract}
Unsupervised domain adaptation (UDA) intends to transfer knowledge from a labeled source domain to an unlabeled target domain. Many current methods focus on learning feature representations that are both discriminative for classification and invariant across domains by simultaneously optimizing domain alignment and classification tasks. However, these methods often overlook a crucial challenge: the inherent conflict between these two tasks during gradient-based optimization.
In this paper, we delve into this issue and introduce two effective solutions known as Gradient Harmonization, including GH and GH++, to mitigate the conflict between domain alignment and classification tasks. GH operates by altering the gradient angle between different tasks from an obtuse angle to an acute angle, thus resolving the conflict and trade-offing the two tasks in a coordinated manner. Yet, this would cause both tasks to deviate from their original optimization directions. We thus further propose an improved version, GH++, which adjusts the gradient angle between tasks from an obtuse angle to a vertical angle. This not only eliminates the conflict but also minimizes deviation from the original gradient directions.
Finally, for optimization convenience and efficiency, we evolve the gradient harmonization strategies into a dynamically weighted loss function using an integral operator on the harmonized gradient.
Notably, GH/GH++ are orthogonal to UDA and can be seamlessly integrated into most existing UDA models. Theoretical insights and experimental analyses demonstrate that the proposed approaches not only enhance popular UDA baselines but also improve recent state-of-the-art models.
\end{abstract}
}

\maketitle

\IEEEdisplaynontitleabstractindextext

%
\IEEEpeerreviewmaketitle

\IEEEraisesectionheading{\section{Introduction}\label{sec:introduction}}

%
%
%
%

\IEEEPARstart{D}{eep} convolutional neural networks and transformers, driven by extensive labeled samples, have achieved remarkable success in various computer vision tasks such as classification, semantic segmentation and object detection.  However, these models often demonstrate high vulnerability when deployed in novel application scenarios due to data distribution discrepancies. The process of collecting and annotating data across various domains is expensive, labor-intensive and time-consuming. Consequently, Unsupervised Domain Adaptation (UDA) arises to transfer the knowledge from a labeled source domain to an unlabeled target domain~\cite{r:1,long2016deepdeep,jiang2018stacked,dai2022graph}. 
\begin{figure}[t]
\centering
\includegraphics[width=0.33\textwidth]{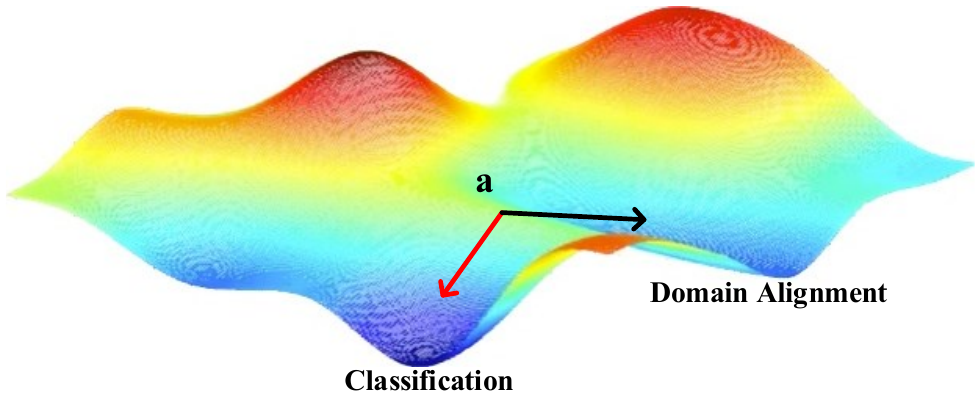} 
\caption{Motivation of the proposed Gradient Harmonization. At point $\textbf{a}$, the black and red arrows point to the optimal gradient descent direction of domain alignment and classification, respectively. The obtuse angle formed by the two gradients of both tasks leads to optimization conflict and further destroy the multi-task optimality.}
\label{motivation}
\end{figure}

In recent years, UDA algorithms have made significant progress in enhancing classification performance~\cite{48guide,4CDAN,c:12,c:13, c:14,51,52}. The primary focus of these approaches is to acquire domain-invariant feature representations, thereby achieving domain alignment and narrowing the probability distributions across domains. Currently, domain alignment methods fall into two main categories: distance metrics-based methods~\cite{r:15,c:16,c:17,44subspace,45transfer,c:22} and adversarial learning-based methods~\cite{46decaf,c:19,c:3GVB,47li,c:18}. The distance metrics-based approaches align the source and target domains by mapping them into a shared feature space while minimizing the distribution disparities between them. Inspired by generative adversarial networks~\cite{42}, adversarial learning techniques have been introduced to tackle domain adaptation challenges~\cite{c:23, 4CDAN}. For instance, ~\cite{c:23} is among the pioneering methods to achieve domain alignment by introducing a game between the feature extractor and domain classifier. \cite{4CDAN} adjusts the adversarial domain adaptation models on discriminative information conveyed in the classifier predictions. These methods usually jointly optimize domain alignment and classification tasks in the course of training.

\emph{However, due to their objective differences between the two tasks, the optimal gradient descent directions from the two tasks may be uncoordinated or imbalanced. Therefore, directly sharing network parameters (i.e., feature generator) may result in optimization conflict between tasks and affect domain-invariant feature learning.} During optimization, the angle between the gradients of the two tasks is sometimes an \emph{obtuse} angle, indicating an obvious gradient conflict. As shown in Fig. \ref{motivation}, the black arrow and red arrow point to the optimal gradient descent direction of the domain alignment task and classification task, respectively. In the joint optimization process, obtuse angle formed by the two gradients of both tasks leads to optimization conflict, which reveals the sub-optimality of the two tasks.

\begin{figure}[t]
\centering
 \begin{minipage}{4.2cm}
 \centerline{\includegraphics[scale=0.28]{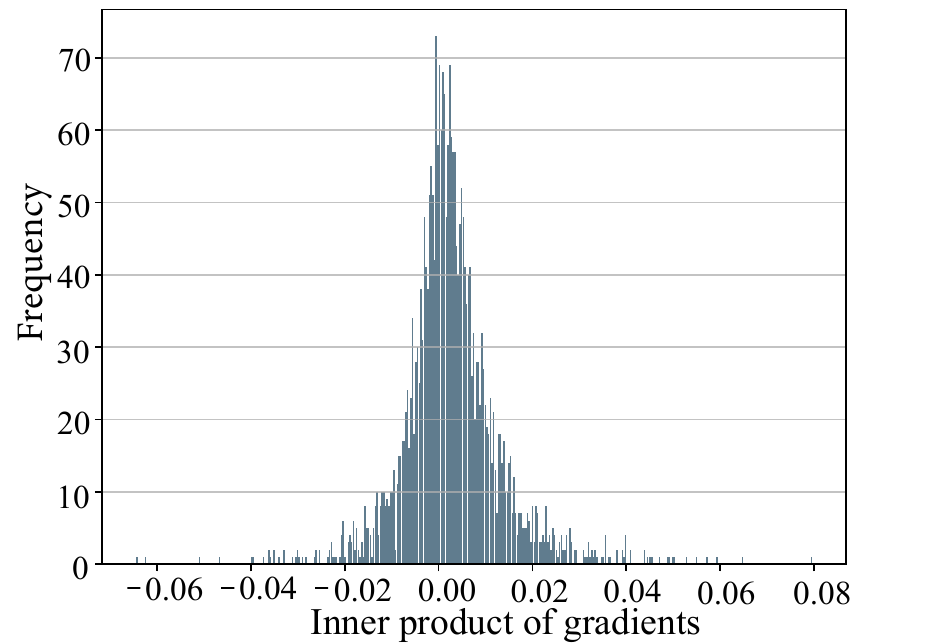}}
 \centerline{(a) MCD}
\end{minipage}
\begin{minipage}{4.2cm}
 \centerline{\includegraphics[scale=0.28]{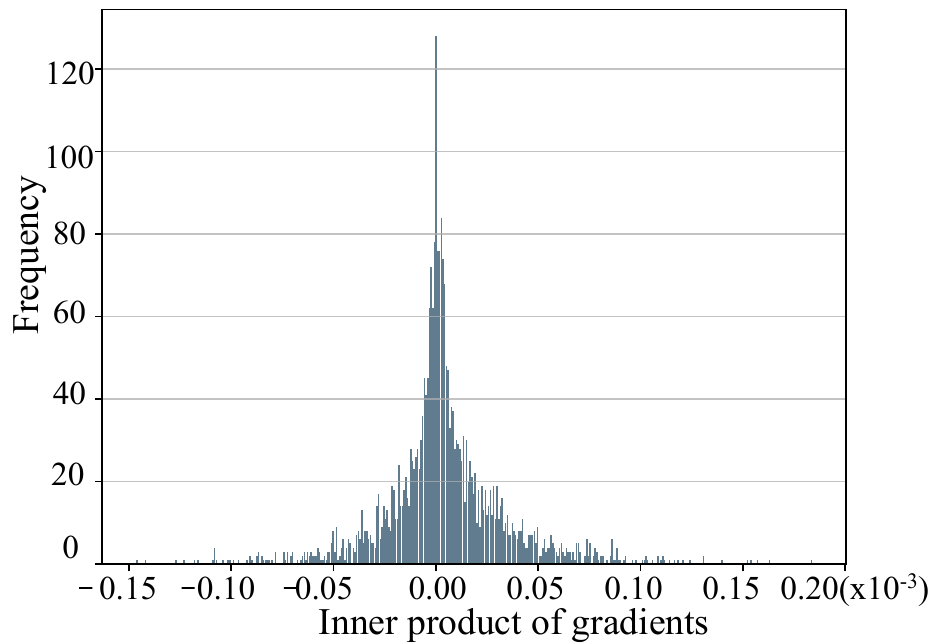}}
 \centerline{(b) DWL}
\end{minipage}
\hfill
  \caption{Inner product distributions (histogram) of the two baselines MCD~\cite{c:2MCD} and DWL~\cite{c:5DWL} in the training process. The horizontal axis represents the inner product of the two gradients, and the vertical axis represents frequency (i.e., number of occurrences of inner product of two gradients). Obviously, both (a) and (b) exist obtuse angles, i.e. optimization conflict, and the optimization conflicts of (a) are more serious.}
  \label{figfig}
\end{figure}

To further verify the above observations, we conduct some experiments on real cross-domain classification scenarios. Fig. \ref{figfig} displays inner product distributions of two gradients between two tasks (alignment vs. classification) throughout the training process on task of MNIST $\rightarrow$ USPS. From Fig. \ref{figfig}, we can obtain the following two observations. First, it implies that the inner product between two gradients is approximately normally distributed with a mean slightly around above zero. Second, both acute and obtuse angles exist on the two models, where obtuse angles account for about 42\% of the total number in MCD~\cite{c:2MCD} and 37\% in DWL~\cite{c:5DWL} (DWL relieves the imbalance by trade-offing the transferability and discriminability to some extent but ignores optimization conflicts, whereas MCD does not take into account the conflict problem). In the joint optimization process, the aggregated gradient direction generated by two obtuse gradients will be seriously deviated from their respective optimal gradient descent directions, thereby resulting in sub-optimal alignment and classification. Further, we empirically observe that the obtuse angle is always present during training over time and cannot be eliminated automatically. Therefore, \textit{how to reasonably eliminate optimization conflicts while maintaining the task-specific optimality during training is a challenging and practical work}. 

Multi-task learning (MTL), as explored in studies such as \cite{desideri2012multiple, Multitasklearning2017, vandenhende2021multi, sener2018multi, mahapatra2020multi}, focuses on simultaneously optimizing multiple conflicting criteria. These approaches typically seek one or more Pareto optimal solutions that offer different trade-offs. Inspired by MTL, ParetoDA \cite{liang2021pareto} introduces a target-classification-mimicking (TCM) loss based on held-out target data and dynamically seeks a desirable Pareto optimal solution for the target domain. However, Pareto optimization relies on additional information to make decisions and requires significant computational resources and time in high-dimensional or complex systems.

In this paper, to address the aforementioned challenges, we propose an idea of \emph{de-conflict on the gradients}. Technically, we introduce two intuitive yet highly effective approaches called Gradient Harmonization, involving GH and GH++, which aim to alleviate optimization conflicts that emerge between the domain alignment and classification tasks by leveraging insights from geometric awareness in model optimization. Specifically, we first calculate the original gradients of two tasks to find out if there is a conflict, i.e., the angle between the gradients of the tasks is obtuse (negatively correlated). Subsequently, we employ GH to turn the angle between the original gradients of two tasks from an obtuse angle to an acute angle, i.e., positively correlated. This harmonization of gradients enables both tasks to progress in a coordinated manner. However, this process may cause both tasks to deviate from their original directions to compromise performance.
To further mitigate gradient deviation while maintaining the task-specific optimality, we introduce an improved version, GH++, which turns the angle between the original gradients of two tasks from an obtuse angle to a vertical angle, i.e., orthogonal. Then both tasks can evolve harmoniously during joint training while preserving their individual task-specific optimality.
Detailed theoretical support is also provided for the proposed approaches.

\begin{figure*}[t]
\centering
\includegraphics[width=0.95\textwidth]{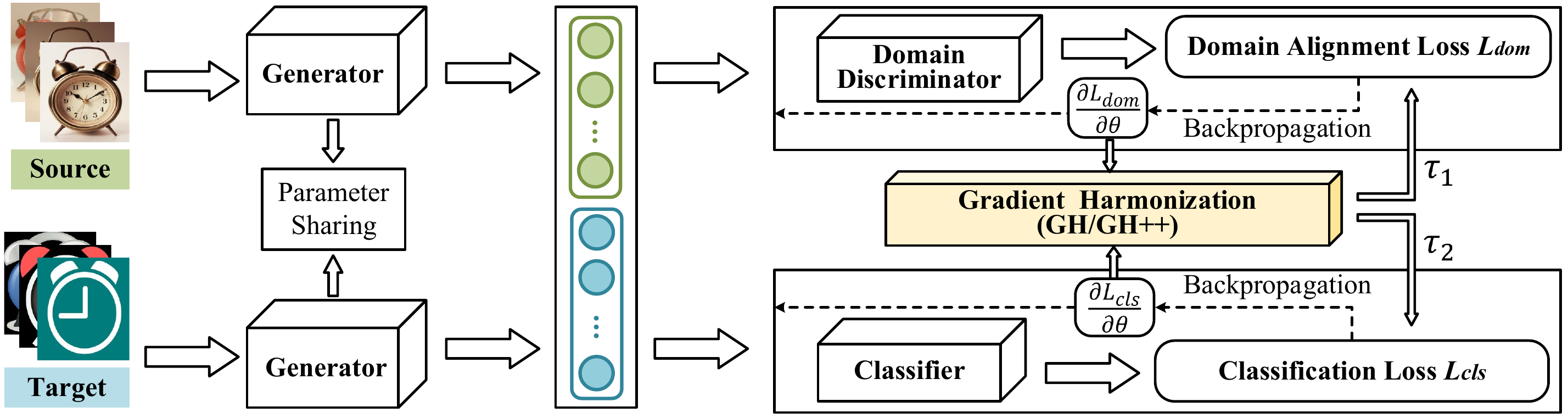} 
\setlength{\abovecaptionskip}{0.cm}
\setlength{\belowcaptionskip}{-0.2cm}
\caption{The usage illustration of our GH module. Optimization objectives include universal domain alignment loss and classification loss. GH module is responsible for harmonization process for the gradients of the two losses. Then the coefficients $\tau_1$ and $\tau_2$ are deduced from GH with the loss gradients $g_1$ and $g_2$ to reweight the two losses. Finally, the reweighted loss functions are backpropagated to update network parameters.}
\label{fig2}
\end{figure*}

Furthermore, in order to facilitate optimization and improve computation efficiency, we derive an equivalent but more efficient model of GH/GH++ for unsupervised domain adaptation. The equivalent model is derived as a dynamic objective function, namely UDA+GH/GH++, which can achieve fast elimination of optimization conflicts without sacrificing the task-specific objective. The proposed GH/GH++ is a universal plug-and-play approach for between-task balance learning and can be easily embedded in most alignment-based unsupervised domain adaptation methods. This work is a new upgrade and more general form of our previous CVPR paper~\cite{c:5DWL} from the perspective of gradient harmonization.
The main contributions and novelties of this paper are summarized as follows:
\begin{itemize}
\item We verify and visualize the gradient conflict in UDA methods and develop two novel Gradient Harmonization approaches, including  GH and GH++. The proposed approaches alleviate the gradient conflict problem between the domain alignment task and classification task by adjusting their gradient angle from obtuse to acute/vertical angle adaptively. Fig. \ref{fig2} is a representative and illustrative example to depict the UDA model based on GH/GH++.
\item  The proposed approaches are orthogonal to most existing UDA methods. To facilitate optimization and improve computation efficiency, we further derive the equivalent models, i.e., \textbf{UDA with GH/GH++}. Specifically, the proposed approaches can be evolved into dynamically weighted loss functions via an integral operation, and promote optimization.
\item Extensive experiments validate the effectiveness and universality of our approaches on various benchmarks in several mainstream UDA models. More insights and analyses justify the reasonability of the proposed approaches.
\end{itemize}

\section{Related Work}

\subsection{Unsupervised Domain Adaptation (UDA)}


Unsupervised Domain Adaptation (UDA) aims to leverage the knowledge learned from a labeled source dataset to solve similar tasks in a new unlabeled domain. Recent works~\cite{liu2018structure, lee2019sliced, c:12, 2123, xu2019larger, lee2019sliced, li2021meta, yan2021transferable, na2021fixbi, li2022dynamic} have focused on UDA based on domain alignment and discriminative feature learning methods.

\textbf{Domain Alignment.} The mainstream approach for UDA aims to realize domain alignment by learning domain-invariant feature representations across domains, which can be mainly summarized into two categories: distance metric based methods and adversarial learning based methods. First, the methods based on distance metrics mostly use several existing distance measurement indicators to assess the alignment degree across domains, such as MMD~\cite{r:38}, MDD~\cite{r:39} and their variants. With the popularity of deep learning methods, more and more researchers apply deep neural networks to domain adaptation. Compared with traditional non-deep domain adaptation methods, deep methods directly improve the learning effect on different classification tasks. Second, some domain adaptation methods based on adversarial training can learn the domain-invariant feature representations well and better implement knowledge transfer. For example, CDAN~\cite{4CDAN} proposes an adversarial adaptation model for the discriminative information transmitted in the prediction of the classifier. GVB~\cite{c:3GVB} learns the domain-invariant feature representations by applying the gradually vanishing bridge mechanism on the feature generator.
Besides, considering the equilibrium problem of adversarial learning, FGDA \cite{gao2021gradient} reduces the distribution discrepancy by constraining feature gradient. CGDM \cite{CGMD2021} explicitly minimizes the discrepancy of gradients generated by source samples and target samples to improve the accuracy of target samples. FixBi~\cite{na2021fixbi} introduces a fixed ratio-based mixup to augment multiple intermediate domains between the source and target domain.

\textbf{Classification.} Other approaches focus on improving the performance of the classification task by enhancing class discriminability feature learning. ETD~\cite{c:14} puts forward an attention-aware optimal transport distance to measure the domain discrepancy under the guidance of the prediction feedback and enables the model to learn distinguished feature representations. MCD~\cite{c:2MCD} proposes to maximize the prediction discrepancy of two classifiers to obtain strong discriminant feature representations. BSP~\cite{c:25} penalizes the largest singular values so that other eigenvectors can be relatively strengthened to boost the feature discriminability. CAN~\cite{kang2019contrastive} optimizes the metric for minimizing the domain discrepancy, which explicitly models the intra-class domain discrepancy and the inter-class domain discrepancy. RSDA~\cite{gu2020spherical} introduces a spherical classifier for label prediction and a spherical domain discriminator for discriminating domain labels and utilizes robust pseudo-label loss in the spherical feature space. DMRL~\cite{wu2020dual} focuses on guiding the classifier to enhance consistent predictions between samples and enriches the intrinsic structures of the latent space. To combine the source and target domains, DMRL introduces two mixup regularizations based on randomness. Recently, SSRT~\cite{sun2022safe} and TVT~\cite{yang2023tvt} learn both transferable and discriminative features by applying Vision Transformer (ViT).

\textbf{Balancing Domain Alignment and Classification.}
Previous approaches concentrate on learning domain-invariant and class-discriminative feature representations by jointly optimizing domain alignment and classification task but ignore imbalance or uncoordinated optimization problem between tasks.
Recently, some methods have started to pay attention to this problem and adopt strategies to alleviate it. To adapt to different cross-domain classification scenarios, DWL~\cite{c:5DWL} proposes dynamic weighted learning to avoid the discriminability vanishing problem caused by excessive alignment and domain misalignment problem caused by excessive discriminant learning from a macro perspective, which ignores gradient conflict between two tasks during optimization.
Meta~\cite{c:6MetaAlign} maximizes inner product of the gradients of the two tasks during training. However, it cannot guarantee the adjusted gradient direction is close to the original optimal gradient descent direction, and even appears serious deviation. Since Meta processes the gradients of the two tasks in an extreme way, it may produce some negative feedback, such as sacrificing partial alignment or classification performance.
Recently, in order to mitigate the effects of conflicting gradients, ParetoDA \cite{liang2021pareto} adopts Pareto optimization \cite{mahapatra2020multi} to cooperatively optimize all training objectives, which searches for the desirable Pareto optimal solution on the entire Pareto front. However, it is computationally inefficient. In this paper, we propose two intuitive yet efficient approaches, including GH and GH++, to handle the gradient conflict.

\textbf{Differences.} The proposed approaches represent a fundamental departure from previous methods like DWL~\cite{c:5DWL}, Meta~\cite{c:6MetaAlign}, and ParetoDA~\cite{liang2021pareto}. Our primary goal is to implicitly address the optimization conflict throughout the entire training process by active gradient harmonization, thereby ensuring a balanced learning between tasks. However, as depicted in Figure \ref{figfig}, the optimization conflict still persists in DWL. Additionally, ParetoDA introduces Pareto optimization and an additional target-classification-mimicking (TCM) loss, which necessitates the involvement of all class-wise discriminators. In contrast, the proposed GH/GH++ is evolved into a dynamically weighted loss function via an integral operator on the harmonized gradient. 

\vspace{-3mm}
\subsection{Multi Task Learning}
Multi-task learning (MTL) \cite{Multitasklearning1993, misra2016cross, Multitasklearning2017, vandenhende2021multi} aims at learning multiple tasks in a unified model to achieve mutual improvement among tasks considering their shared knowledge. By sharing parameters across tasks, MTL methods learn more efficiently with an overall smaller model size compared to learning with separate models.
Prior MTL approaches formulate the total objective as a weighted sum of task-specific objectives, such as DWA \cite{liu2019end} and GradNorm \cite{chen2018gradnorm}, which optimize weights based on task-specific learning rates or by random weighting. However, these \emph{weighted optimization based MLT} may not achieve satisfactory performance due to the presence of gradient conflicts among different tasks. \emph{Gradient optimization based MLT} \cite{desideri2012multiple, sener2018multi, yu2020gradient, mahapatra2020multi, liu2021conflict, liu2021towards} overcome this limitation, mitigating effects of conflicting or dominating gradients. MGDA \cite{desideri2012multiple} proposes to simply update the shared network parameters along a descent direction which leads to solutions that dominate the previous one. PCGrad \cite{yu2020gradient} proposes a ``gradient surgery'' to avoid gradient conflicts. CAGrad \cite{liu2021conflict} seeks the gradient direction within the neighborhood of the average gradient and maximizes the worst local improvement of any objective. \cite{bi2022mtrec} merge the gradients of auxiliary tasks and applied a scaling factor to adaptively adjust their impact on the main tasks, followed by applying gradient sharing between the main tasks and the merged auxiliary task. GradDrop \cite{chen2020just} proposes a probabilistic masking procedure, which samples gradients at an activation layer based on their level of consistency. Aligned-MTL \cite{senushkin2023independent} eliminates instability in the training process by aligning the orthogonal components of the linear system of gradients.

In this study, we introduce two simple yet highly efficient approaches, GH and GH++,  to effectively address gradient conflicts between any two distinct tasks. Detailed theoretical derivations and fundamental insights regarding the proposed approaches are elaborated in Sections \ref{section3}. Additionally, under thorough theoretical analysis and a wealth of experiments, we have substantiated the soundness and effectiveness of the proposed methodologies.


\section{Proposed Approach}\label{section3}
\subsection{Problem Definition}
Given a labeled source domain $\mathcal{D}_s=\{x_i^{s},y_i^{s}\}_{i=1}^{n_s}$ with $n_s$ samples and an unlabeled target domain $\mathcal{D}_t=\{x_j^{t}\}_{j=1}^{n_t}$ with $n_t$ samples, where $y_i^{s}$ is the class label of the $i^{th}$ source sample $x_i^{s}$. $\mathcal{D}_s$ and $\mathcal{D}_t$ share the same feature space and category space, but have different data distributions. Our purpose is to utilize the labeled data $\mathcal{D}_s$ and unlabeled data $\mathcal{D}_t$ to learn a deep model, which can accurately predict the class label of samples in the target domain.
\vspace{-3mm}
\subsection{A General Framework of UDA}
\label{section3.1}
Adversarial learning has proven to be an effective method for domain alignment, starting from Domain Adversarial Neural Network (DANN)~\cite{c:23}. The basic idea is to trick the domain discriminator $D$ by generating features via a feature generator $G$. Then the domain discriminator predicts whether the generated feature by $G$ is from the source domain or the target domain. The training of domain alignment is achieved through the game between generator and domain discriminator. The parameter $\theta_g$ of generator $G$ and the parameter $\theta_d$ of domain discriminator $D$ are optimized by the following domain alignment objective function.
\begin{equation}\label{align}
\begin{aligned}
\setlength{\abovedisplayskip}{3pt}
\setlength{\belowdisplayskip}{3pt}
 \begin{split}
    \mathcal{L}_{dom}(\theta_g,\theta_d)=&~\mathbb{E}_{x_i^{s}\sim\mathcal{D}_s}log[D(G(x_i^{s}))]+\\
    &~~~~~\mathbb{E}_{x_j^{t}\sim\mathcal{D}_t}log[1-D(G(x_j^{t}))].
 \end{split}
\end{aligned}
\end{equation}

In order to improve the classification performance of the target domain samples, we must first ensure that the classifier $C$ can correctly classify the samples from the source domain. Thus, the supervised classification loss can be described as
\begin{equation}\label{cls}
\setlength{\abovedisplayskip}{3pt}
\setlength{\belowdisplayskip}{3pt}
    \mathcal{L}_{cls}(\theta_g,\theta_c)=\frac{1}{N_s}\sum_{i=1}^{N_s}\mathcal{L}_{ce}(C(G(x_i^{s};\theta_g);\theta_c),y_i^{s}),
\end{equation}
where $\mathcal{L}_{ce}$ is the standard cross-entropy loss function.

During training stage, the existing methods usually jointly optimize the two objective functions ($\mathcal{L}_{dom}$ and $\mathcal{L}_{cls}$) to obtain domain-invariant and class-discriminant feature representation. The overall minimax objective function is
\begin{equation}\label{gc}
\setlength{\abovedisplayskip}{3pt}
\setlength{\belowdisplayskip}{3pt}
    \underset{\theta_g,\theta_c}{\min}~~\underset{\theta_d}{\max}~~\mathcal{L}_{dom}+\mathcal{L}_{cls},
\end{equation}
where $\theta_g$, $\theta_d$, $\theta_c$ denote the parameters of feature generator, domain discriminator and classifier, respectively.
\begin{figure*}[t]
\centering
\includegraphics[width=0.9\textwidth]{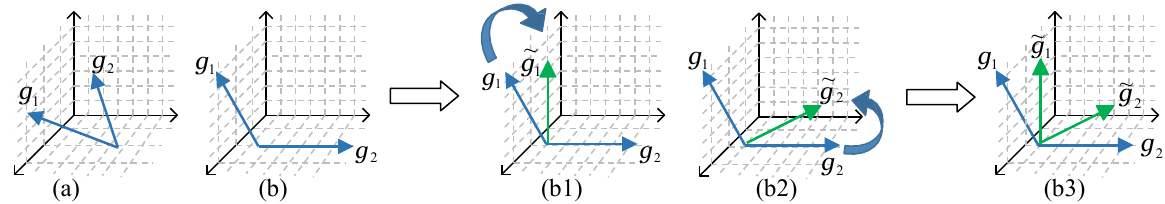} 
\setlength{\abovecaptionskip}{-0.cm}
\setlength{\belowcaptionskip}{-0.3cm}
\caption{Overall idea of de-conflict for the gradients $g_1$ and $g_2$ of two tasks. (a) displays the angle between two gradients is an acute angle. (b) displays the angle between two gradients is an obtuse angle. Following GH, (b) needs to be processed and (b1), (b2) and (b3) are harmonization process. (b1) and (b2) are the details of performing our gradient harmonization on $g_1$ and $g_2$, resp. (b3) is final harmonization results. $\tilde{g}_1$ and $\tilde{g}_2$ represent the gradients after harmonization, resp. Finally, after applying GH, the angle between $g_1$ and $g_2$ has changed from obtuse angle to acute angle.}
\label{fig3}
\end{figure*}

\vspace{-3mm}
\subsection{Gradient Harmonization (GH)}
\label{section3.2}
Domain alignment and classification are two different tasks. Their optimal gradient descent directions may not be coordinated, which results in the optimization conflict of the two loss functions in the training process and deteriorates the final domain adaptation performance.

In order to ensure that the two target tasks can be optimized in a coordinated manner, we propose an idea of \emph{de-conflict} on the gradients, which then formulates the proposed GH technique, i.e., altering the gradient angle between different tasks from an obtuse angle to an acute angle. The de-conflict process for the gradients (i.e., $g_1$ and $g_2$) is schematically shown in Fig. \ref{fig3}. Specifically, we first provide three \textbf{Lemmas} to support our idea. Then the proposed GH is summarized as \textbf{Theorem 1}. For convenience, we define $\mathcal{L}_1(\Theta)$ and $\mathcal{L}_2(\Theta)$ as the general form of any two conflicted loss functions. In UDA, $\mathcal{L}_1(\Theta)$ and $\mathcal{L}_2(\Theta)$ represent the domain alignment loss $\mathcal{L}_{dom}(\theta_g,\theta_d)$ and the classification loss $\mathcal{L}_{cls}(\theta_g,\theta_c)$, respectively. $\Theta=\{\theta_g, \theta_{d},\theta_{c}\}$ indicates model parameters of generator, discriminator and classifier. In fact, the three Lemmas aim to deduce the gradient harmonization formula during model optimization (training) phase, by exhausted mathematical solving process.

\begin{lemma}
Given two objective functions $\mathcal{L}_1(\Theta)$ and $\mathcal{L}_2(\Theta)$, we define $g_1$ and $g_2$ as their gradient, respectively, and $\tilde{g}_1$ is the result of harmonizing the gradient $g_1$. For minimizing the objective $\mathcal{L}_1(\Theta-\tilde{g}_1)+\mathcal{L}_2(\Theta-\tilde{g}_1)$, we consider to first minimize the objective $\mathcal{L}_1(\Theta-\tilde{g}_1)$, and then there is,
\begin{equation}\label{g1}
\setlength{\abovedisplayskip}{2pt}
\setlength{\belowdisplayskip}{2pt}
    \tilde{g}_1=g_1-\delta(g_1^Tg_2<0)\frac{g_1^Tg_2}{\lVert g_2 \rVert^2}g_2,
\end{equation}
where $\delta(\cdot)$ represents the indicator function whose value is 0 or 1 and the mathematical expression is:
\begin{equation}\label{aa}
\setlength{\abovedisplayskip}{4pt}
\setlength{\belowdisplayskip}{4pt}
\delta(A) =
\begin{cases}
1,  & if~A~is~true \\
0,  & if~A~is~false \\
\end{cases}
\end{equation}
\label{lemma1}
\end{lemma}

\vspace{-10mm}
\begin{proof}\renewcommand{\qedsymbol}{}
In order to minimize the loss function $\mathcal{L}_1(\Theta-\tilde{g}_1)$, the problem is transformed into solving the following model:
\begin{equation}
\setlength{\abovedisplayskip}{4pt}
\setlength{\belowdisplayskip}{4pt}
\begin{aligned} \label{A}
&~~~~~~~~~~~\underset{\tilde{g}_1}{\min}~~\mathcal{L}_1(\Theta-\tilde{g}_1)  \\
s.t.~&\mathcal{L}_1(\Theta-\tilde{g}_1)\leq\mathcal{L}_1(\Theta),~~\mathcal{L}_2(\Theta-\tilde{g}_1)\leq\mathcal{L}_2(\Theta).
\end{aligned}
\end{equation}
Due to the gradient $g_1=\nabla_{\Theta}\mathcal{L}_1(\Theta)$ is the fastest descent direction of the objective function $\mathcal{L}_1(\Theta)$, if the harmonic gradient $\tilde{g}_1$ is the same as $g_1$ as much as possible, the loss function $\mathcal{L}_1(\Theta)$ will be optimized along the gradient descent direction. If the angle between the harmonic gradient $\tilde{g}_1$ and $g_1$ is an acute angle, then $\tilde{g}_1$ points to the direction of gradient descent. Naturally, the angle between $\tilde{g}_1$ and $g_2=\nabla_{\Theta}\mathcal{L}_2(\Theta)$ is also expected to be an acute angle. Therefore, the above model is transformed into:
\begin{equation}\label{B}
\setlength{\abovedisplayskip}{4pt}
\setlength{\belowdisplayskip}{4pt}
\begin{aligned}
&\underset{\tilde{g}_1}{\min}~~\frac{1}{2}\lVert g_1-\tilde{g}_1\rVert^2  \\
&s.t.~\tilde{g}_1^Tg_1\geq0, \tilde{g}_1^Tg_2\geq0.
\end{aligned}
\end{equation}
The problem (\ref{B}) is a convex optimization problem. To solve the problem (\ref{B}), we define the Lagrangian function
\begin{equation}\label{F}
\setlength{\abovedisplayskip}{4pt}
\setlength{\belowdisplayskip}{4pt}
    \mathcal{L}(\tilde{g}_1,\alpha_1,\alpha_2)=\frac{1}{2}\lVert g_1-\tilde{g}_1\rVert^2
    -\alpha_1\tilde{g}_1^Tg_1-\alpha_2\tilde{g}_1^Tg_2,
\end{equation}
where $\alpha_1$, $\alpha_2$ are the Lagrangian multipliers associated with the inequality constraint of problem (\ref{B}). To simplify the solution, first note that
\begin{equation}
\begin{aligned}
\setlength{\abovedisplayskip}{4pt}
\setlength{\belowdisplayskip}{4pt}
P= &\max_{\alpha_1\geq0,\alpha_2\geq0} \mathcal{L}(\tilde{g}_1,\alpha_1,\alpha_2) \\
=& \left\{
\begin{array}{lcl}
\frac{1}{2}\lVert g_1-\tilde{g}_1\rVert^2,&& \tilde{g}_1^Tg_1\geq0, \tilde{g}_1^Tg_2\geq0\\
+\infty,&& otherwise\\
\end{array} \right.
\end{aligned}
\label{p1}
\end{equation}
This means that we can express the optimal value of the primal problem (\ref{B}) as:
\begin{equation}\label{p2}
\setlength{\abovedisplayskip}{4pt}
\setlength{\belowdisplayskip}{4pt}
    p^{\ast} = {\min_{\tilde{g}_1}}~~\max_{\alpha_1\geq0,\alpha_2\geq0}~~\mathcal{L}(\tilde{g}_1,\alpha_1,\alpha_2).
\end{equation}
By the definition of the dual function, we can express the optimal value of the dual problem:
\begin{equation}\label{D}
\setlength{\abovedisplayskip}{4pt}
\setlength{\belowdisplayskip}{4pt}
    d^{\ast}=\underset{\alpha_1\geq0,\alpha_2\geq0}{\max}~~\underset{\tilde{g}_1}{\min}~~\mathcal{L}(\tilde{g}_1,\alpha_1,\alpha_2),
\end{equation}
where $\alpha_1\geq0,\alpha_2\geq0$ is the dual feasibility condition. It is easy to get the weak duality $d^{\ast} \leq p^{\ast}$. When the strong duality holds, i.e., $d^{\ast} = p^{\ast}$, we have
\begin{equation}\label{p3}
\setlength{\abovedisplayskip}{4pt}
\setlength{\belowdisplayskip}{4pt}
    \alpha_1\tilde{g}_1^Tg_1=0, \alpha_2\tilde{g}_1^Tg_2=0.
\end{equation}
Eq. (\ref{p3}) is known as complementary slackness condition. To summarize, for convex  optimization problem (\ref{B}), we have
\begin{equation}\label{p4}
\setlength{\abovedisplayskip}{4pt}
\setlength{\belowdisplayskip}{4pt}
\begin{aligned} \left\{
\begin{array}{lcl}
\nabla_{\tilde{g}_1} \mathcal{L}(\tilde{g}_1,\alpha_1,\alpha_2)=0,&&\textcircled{1}\\
\tilde{g}_1^Tg_1\geq0, \tilde{g}_1^Tg_2\geq0,&&\textcircled{2}\\
\alpha_1\geq0,\alpha_2\geq0,&&\textcircled{3} \\
\alpha_1\tilde{g}_1^Tg_1=0, \alpha_2\tilde{g}_1^Tg_2=0.&&\textcircled{4}\\
\end{array} \right.
\end{aligned}
\end{equation}
Conditions (\ref{p4}) are called the \emph{Karush-Kuhn-Tucker} (KKT) conditions\cite{2004Convex}, which is the the necessary condition for pair of primal and dual optimal points. 
We can solve the dual problem Eq. (\ref{D}). As for the inner optimization problem, i.e., $D=\underset{\tilde{g}_1}{\min}~~\mathcal{L}(\tilde{g}_1,\alpha_1,\alpha_2)$, we then obtain its optimal solution by setting its derivative with respect to $\tilde{g}_1$ as zero. Then,
\begin{equation}\label{E}
\setlength{\abovedisplayskip}{4pt}
\setlength{\belowdisplayskip}{4pt}
    \tilde{g}_1=g_1+\alpha_1g_1+\alpha_2g_2.
\end{equation}
We can calculate the extremum of $D$ by substituting Eq. (\ref{E}) into Eq. (\ref{F}) :
\begin{equation}\label{G}
\setlength{\abovedisplayskip}{4pt}
\setlength{\belowdisplayskip}{4pt}
\begin{aligned}
    D&=\underset{\tilde{g}_1}{\min}~~\mathcal{L}(\tilde{g}_1,\alpha_1,\alpha_2)\\
    &=\frac{1}{2}\tilde{g}_1^T\tilde{g}_1 -\tilde{g}_1^Tg_1-\alpha_1\tilde{g}_1^Tg_1-\alpha_2\tilde{g}_1^Tg_2+const\\
    &=\frac{1}{2}\tilde{g}_1^T\tilde{g}_1- \tilde{g}_1^T(g_1+\alpha_1g_1+\alpha_2g_2)+const\\
    &= -\frac{1}{2}\tilde{g}_1^T\tilde{g}_1+const\\
    &=-\frac{1}{2}\lVert g_1+\alpha_1g_1+\alpha_2g_2\rVert^2+const.\\
\end{aligned}
\end{equation}
We then get the outer optimization problem by substituting Eq. (\ref{G}) into Eq. (\ref{D}) as follows:
\begin{equation}\label{H}
\setlength{\abovedisplayskip}{4pt}
\setlength{\belowdisplayskip}{4pt}
    \underset{\alpha_1\geq0,\alpha_2\geq0}{\max}~~-\frac{1}{2}\lVert g_1+\alpha_1g_1+\alpha_2g_2\rVert^2.
\end{equation}
For the problem (\ref{H}), we can take the derivative of $\alpha_1$ and $\alpha_2$ respectively, and set the derivative to 0. Then we can obtain the optimal solution: $\alpha_1^*=-1$, $\alpha_2^*=0$. Obviously, the obtained $\alpha_1^*$ does not satisfy the dual feasibility condition. Therefore, it is necessary to find the boundary solution that satisfies the constraints, which is established under two possible cases:


Case~1: Suppose $\alpha_1^*=0$, the Eq. (\ref{H}) is transformed into:
\begin{equation}\label{I}
\setlength{\abovedisplayskip}{4pt}
\setlength{\belowdisplayskip}{4pt}
    \underset{\alpha_2}{\min}~~\frac{1}{2}\lVert g_1+\alpha_2g_2\rVert^2.
\end{equation}
We can set the derivative of $\alpha_2$ to 0, i.e., $(g_1+\alpha_2 g_2)^T g_2 = 0$. Then we can obtain the optimal solution:
\begin{equation}\label{J}
\setlength{\abovedisplayskip}{4pt}
\setlength{\belowdisplayskip}{4pt}
    \alpha_2^*=-\frac{g_1^Tg_2}{\lVert g_2 \rVert^2}.
\end{equation}
Obviously, Eq. (\ref{J}) satisfies the KKT conditions $\textcircled{1} \textcircled{2} \textcircled{4}$ in Eq. (\ref{p4}). According to dual feasibility, i.e., $\textcircled{3}$ in Eq. (\ref{p4}), $\alpha_2^*=-\frac{g_1^Tg_2}{\lVert g_2 \rVert^2}\geq0$, then we have $g_1^Tg_2\leq0$. Therefore, we can obtain a boundary solution
\begin{equation}
\setlength{\abovedisplayskip}{4pt}
\setlength{\belowdisplayskip}{4pt}
\begin{cases}
   \alpha_1^*=0,\\
   \alpha_2^*=-\frac{g_1^Tg_2}{\lVert g_2 \rVert^2}, \quad g_1^Tg_2\leq0
\end{cases}
\end{equation}

Case~2: Suppose $\alpha_2^*=0$, the Eq. (\ref{H}) is transformed into:
\begin{equation}\label{p5}
    \underset{\alpha_1}{\min}~~\frac{1}{2}\lVert g_1+\alpha_1g_1\rVert^2.
\end{equation}
Then we can obtain $\alpha_1^*=0$ in nature. Obviously, it satisfies the KKT conditions $\textcircled{1} \textcircled{3} \textcircled{4}$ in Eq. (\ref{p4}). According to primal feasibility, i.e., $\textcircled{2}$ in Eq. (\ref{p4}), we have $g_1^Tg_2\geq0$. Therefore, the boundary solution is
\begin{equation}
\setlength{\abovedisplayskip}{4pt}
\setlength{\belowdisplayskip}{4pt}
\begin{cases}
   \alpha_1^*=0,\\
   \alpha_2^*=0,  \quad g_1^Tg_2\geq0
\end{cases}
\end{equation}
In summary, the optimal solution of problem (\ref{H}) is:
\begin{equation}\label{L}
\setlength{\abovedisplayskip}{4pt}
\setlength{\belowdisplayskip}{4pt}
\begin{cases}
\alpha_1^*=0\\
\alpha_2^*=-\delta(g_1^Tg_2<0)\frac{g_1^Tg_2}{\lVert g_2 \rVert^2}.
\end{cases}
\end{equation}
Then the optimal value of the primal problem (\ref{B}) $\tilde{g}$ can be obtained by substituting Eq. (\ref{L}) into Eq. (\ref{E}).
\begin{equation}
\setlength{\abovedisplayskip}{4pt}
\setlength{\belowdisplayskip}{4pt}
    \tilde{g}_1=g_1-\delta(g_1^Tg_2<0)\frac{g_1^Tg_2}{\lVert g_2 \rVert^2}g_2,
\end{equation}
where $\tilde{g}_1$ is the result of harmonizing the gradient $g_1$ of $\mathcal{L}_1$. Then the proof of Eq. (\ref{g1}) in \textbf{Lemma 1} is completed.

The above process not only solves the problem (\ref{B}) but also harmonizes the gradient $g_1$ of $\mathcal{L}_1(\Theta)$, i.e., Fig. \ref{fig3} (b1).
\end{proof}

\begin{lemma}
Define $\tilde{g}_2$ as the result of harmonizing the gradient $g_2$, we consider to further minimize the optimization objective $\mathcal{L}_2(\Theta-\tilde{g}_2)$, then there is,
\begin{equation}\label{g2}
\setlength{\abovedisplayskip}{4pt}
\setlength{\belowdisplayskip}{4pt}
    \tilde{g}_2=g_2-\delta(g_1^Tg_2<0)\frac{g_2^Tg_1}{\lVert g_1 \rVert^2}g_1.
\end{equation}
\label{lemma2}
\end{lemma}
\vspace{-7mm}
\begin{proof}\renewcommand{\qedsymbol}{}
In order to minimize the loss function $\mathcal{L}_2(\Theta-\tilde{g}_2)$, similar to \textbf{Lemma 1}, we need to solve the following minimization problem.
\begin{equation}
\setlength{\abovedisplayskip}{4pt}
\setlength{\belowdisplayskip}{4pt}
\begin{aligned} \label{N}
&~~~~~~~~~~~\underset{\tilde{g}}{\min}~~\mathcal{L}_2(\Theta-\tilde{g}_2)  \\
s.t.~&\mathcal{L}_1(\Theta-\tilde{g}_2)\leq\mathcal{L}_1(\Theta), \mathcal{L}_2(\Theta-\tilde{g}_2)\leq\mathcal{L}_2(\Theta).
\end{aligned}
\end{equation}
The above model can be transformed into
\begin{equation}\label{a4}
\setlength{\abovedisplayskip}{4pt}
\setlength{\belowdisplayskip}{4pt}
\begin{aligned}
&\underset{\tilde{g}_2}{\min}~~\frac{1}{2}\lVert g_2-\tilde{g}_2\rVert^2  \\
&s.t.~\tilde{g}_2^Tg_1\geq0, \tilde{g}_2^Tg_2\geq0.
\end{aligned}
\end{equation}
Similar to problem (\ref{B}), by solving (\ref{a4}), we can obtain
\begin{equation}
\setlength{\abovedisplayskip}{4pt}
\setlength{\belowdisplayskip}{4pt}
    \tilde{g}_2=g_2-\delta(g_1^Tg_2<0)\frac{g_2^Tg_1}{\lVert g_1 \rVert^2}g_1,
\end{equation}
where $\tilde{g}_2$ is the result of harmonizing the gradient $g_2$ of $\mathcal{L}_2(\Theta)$, i.e., Fig. \ref{fig3} (b2). 

\begin{figure}[t]
\centering
\includegraphics[width=0.45\textwidth]{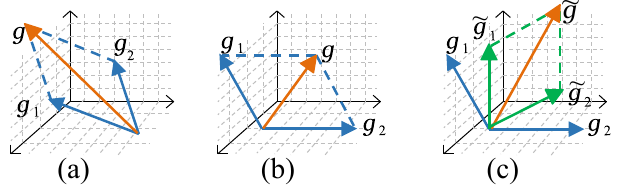}
\setlength{\abovecaptionskip}{-0cm}
\setlength{\belowcaptionskip}{-0.3cm}
\caption{Gradient aggregation. (a) Gradient aggregation when the angle of the original gradients $g_1$ and $g_2$ is an acute angle. (b) Gradient aggregation when the angle of the original gradients $g_1$ and $g_2$ is an obtuse angle. (c) Gradient aggregation after GH for case (b). Comparing (b) and (c), GH changes the magnitude and the direction of the aggregated/combined gradient $g$ and gradient harmonization is realized.}
\label{fig4}
\end{figure}

\end{proof}
\begin{lemma}
Define the overall loss function $\mathcal{L}=\mathcal{L}_1(\Theta)+\mathcal{L}_2(\Theta)$ of two tasks, and $\tilde{g}$ denotes the overall harmonized gradient. We consider the whole optimization objective $\mathcal{L}_1(\Theta-\tilde{g})+\mathcal{L}_2(\Theta-\tilde{g})$, then there is,
\begin{equation}\label{bbbb}
\begin{aligned}
\setlength{\abovedisplayskip}{4pt}
\setlength{\belowdisplayskip}{4pt}
    \!\!\tilde{g}\!=&\tilde{g}_1+\tilde{g}_2\\
    =&g_1\!+\!g_2\!-\!\delta(g_1^Tg_2\!<\!0)\frac{g_1^Tg_2}{\lVert g_2 \rVert^2}g_2\!-\!\delta(g_1^Tg_2\!<\!0)\frac{g_2^Tg_1}{\lVert g_1 \rVert^2}g_1.
\end{aligned}
\end{equation}
\label{lemma3}
\end{lemma}
\vspace{-8mm}
\begin{proof}\renewcommand{\qedsymbol}{}
According to \textbf{Lemma 1} and \textbf{Lemma 2}, the results of harmonizing the gradients $g_1$ and $g_2$, i.e., $\tilde{g}_1$ and $\tilde{g}_2$, can be obtained. By aggregating Eq. (\ref{g1}) in Lemma 1 and Eq. (\ref{g2}) in Lemma 2, the whole gradient after harmonization can be obtained as Eq. (\ref{bbbb}), and proof of Lemma 3 is completed.
\end{proof}

For brevity, we summarize the above process of GH in \textbf{Theorem 1}.

\begin{theorem}
For any two different tasks, optimization conflicts can be eliminated by Gradient Harmonization (GH). Define the overall loss function $\mathcal{L} (\Theta)=\mathcal{L}_1 (\Theta)+\mathcal{L}_2 (\Theta)$, composed of two sub-objectives (refer to domain alignment and classification in this paper), and $g_1$ and $g_2$ as the gradient of $\mathcal{L}_1 (\Theta)$ and $\mathcal{L}_2 (\Theta)$. Then the overall harmonized gradient $\tilde{g}$ of the whole loss $\mathcal{L} (\Theta)$ with GH can be formulated as:
\begin{equation}\label{ggg}
\setlength{\abovedisplayskip}{4pt}
\setlength{\belowdisplayskip}{4pt}
    \!\!\tilde{g}=g_1\!+\!g_2\!-\!\delta(g_1^Tg_2\!<\!0)\frac{g_1^Tg_2}{\lVert g_2 \rVert^2}g_2\!-\!\delta(g_1^Tg_2\!<\!0)\frac{g_2^Tg_1}{\lVert g_1 \rVert^2}g_1.
\end{equation}

\end{theorem}


Obviously, in the optimization process of standard SGD algorithm, the overall gradient of the loss $\mathcal{L}$ is $g\!=\!g_1\!+\!g_2$. In this paper, two cases 
are considered based on the correlation between gradients $g_1$ and $g_2$.

Case \uppercase\expandafter{\romannumeral1}:\ When $g_1$ and $g_2$ are positively correlated or unrelated, i.e., $\cos(g_1,g_2)\geq0$ or $g_1^Tg_2\geq0$, the angle between the two gradients is an \textbf{acute} or \textbf{vertical} angle. Then, we believe that there is no conflict between the two objectives in training, that is, two gradients are coordinated or balanced. Thus $g_1$ and $g_2$ do not need to be harmonized.

\begin{figure}[t]
\centering
\includegraphics[width=0.42\textwidth]{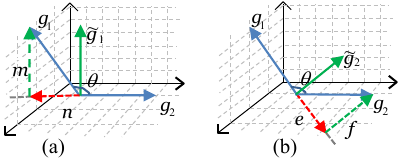} 
\setlength{\abovecaptionskip}{0.cm}
\setlength{\belowcaptionskip}{-0.4cm}
\caption{The Essence of Gradient Harmonization. (a) and (b) denote the essence illustration of performing GH on $g_1$ and $g_2$, respectively.}
\label{Essence}
\end{figure}

Case \uppercase\expandafter{\romannumeral2}:\ When $g_1$ and $g_2$ are negatively correlated, i.e., $\cos(g_1,g_2)<0$ or $g_1^Tg_2<0$, the angle between the two gradients is an \textbf{obtuse} angle. Then, we believe that there is a gradient optimization conflict between the two objectives. In this case, $g_1$ and $g_2$ need to be harmonized. For the overall loss function $\mathcal{L} (\Theta)=\mathcal{L}_1(\Theta)+\mathcal{L}_2(\Theta)$ of an arbitrary model, the proposed \textbf{UDA+GH} is to find a gradient $\tilde{g}$ that satisfies the following problem:

\begin{equation}\label{a33}
\setlength{\abovedisplayskip}{4pt}
\setlength{\belowdisplayskip}{4pt}
    \underset{\tilde{g}}{\min}~~\mathcal{L}_1(\Theta-\tilde{g})+\mathcal{L}_2(\Theta-\tilde{g}),
\end{equation}
where Eq. (\ref{a33}) can be solved by aforementioned three lemmas. Firstly, the harmonic gradient $\tilde{g}_1$ of $g_1$ can be obtained by \textbf{Lemma 1}. Then the harmonic gradient $\tilde{g}_2$ of $g_2$ can be obtained by \textbf{Lemma 2}.  Finally, as can be seen from \textbf{Lemma 3}, it is easy to acquire the whole aggregated gradient $g$ in Eq. (\ref{ggg}) after harmonization and the gradient aggregation process is visualized in Fig. \ref{fig4}.


\vspace{-3mm}
\subsection{Essence and Insights of GH}
\label{essence}
To understand the essence of GH more intuitively, we demonstrate the proposed gradient harmonization by backward inference in this section. Fig. \ref{Essence} (a) and (b) show the essential analysis by performing gradient harmonization on original gradients $g_1$ and $g_2$, respectively. Note that we mainly consider the case where gradient conflict exists, that is, the angle between $g_1$ and $g_2$ is an obtuse angle.

First, we analyze the essence of gradient harmonization for $g_1$ (i.e., Fig. \ref{Essence} (a)). $\theta$ denotes the angle between original gradients $g_1$ and $g_2$. $n$ is the projection of $g_1$ on the reversely extended line of $g_2$. $m$ is perpendicular to $g_2$. Note that each letter except $\theta$ on the figure represents a vector. Naturally, $m$ can be derived as Eq. (\ref{wjk}), from which we can see that the expression of the vector $m$ is the same as Eq. (\ref{g1}) when $g_1^Tg_2<0$. That is, the vector $m$ is the result of gradient harmonization of $g_1$ (i.e., $\tilde{g}_1$).
\begin{equation}\label{wjk}
\begin{aligned}
\setlength{\abovedisplayskip}{3pt}
\setlength{\belowdisplayskip}{3pt}
 \begin{split}
    m=g_1-n=&g_1-|g_1|\cdot cos(\pi-\theta)\cdot (-\frac{g_2}{|g_2|})\\
    =&g_1-|g_1|\cdot cos\theta\cdot \frac{g_2}{|g_2|}\\
    =&g_1-|g_1|\cdot \frac{\langle g_1,g_2\rangle}{|g_1|\cdot |g_2|}\cdot \frac{g_2}{|g_2|}\\
    =&g_1-\frac{\langle g_1,g_2\rangle}{\|g_2\|^2}g_2.\\
 \end{split}
\end{aligned}
\end{equation}
where $\langle\cdot\rangle$ is the inner product operator.

Second, we analyze the essence of gradient harmonization for $g_2$ (i.e., Fig. \ref{Essence} (b)). Similar to (a), the vector $e$ is the projection of $g_2$ on the reversely extended line of $g_1$. The vector $f$ is perpendicular to $g_1$. Through the derivation of Eq. (\ref{nzhk}), it can be found that the expression of the vector $f$ is the same as Eq. (\ref{g2}) when $g_1^Tg_2<0$. That is, the vector $f$ is the result of harmonizing the gradient $g_2$ (i.e., $\tilde{g}_2$).
\begin{equation}\label{nzhk}
\begin{aligned}
\setlength{\abovedisplayskip}{4pt}
\setlength{\belowdisplayskip}{4pt}
 \begin{split}
    f=g_2-e=&g_2-|g_2|\cdot cos(\pi-\theta)\cdot (-\frac{g_1}{|g_1|})\\
    =&g_2-|g_2|\cdot cos\theta\cdot \frac{g_1}{|g_1|}\\
    =&g_2-|g_2|\cdot \frac{\langle g_1,g_2\rangle}{|g_1|\cdot |g_2|}\cdot \frac{g_1}{|g_1|}\\
    =&g_2-\frac{\langle g_1,g_2\rangle}{\|g_1\|^2}g_1.\\
 \end{split}
\end{aligned}
\end{equation}

From the above derivations, the nature of GH is summarized as follows. 1) The harmonic gradient $\tilde{g}_1$ is essentially the projection of the raw gradient $g_1$ on the vertical line of $g_2$. 2) The harmonic gradient $\tilde{g}_2$ is essentially the projection of the raw gradient $g_2$ on the vertical line of $g_1$.

Therefore, with the above nature, the following three observations can be obtained. 1) The proposed gradient harmonization method not only ensures that the harmonic gradient ($\tilde{g}_1/\tilde{g}_2$) is close to the original gradient ($g_1$/$g_2$), but also ensures that the aggregated gradients before and after applying GH are close to each other. 2) If the angle between the original gradients $g_1$ and $g_2$ is $\theta$, then the angle between harmonized gradient $\tilde{g}_1$ and $\tilde{g}_2$ becomes $\pi-\theta$ after applying GH. That is, GH really realizes the transformation from \textit{obtuse} angle to \textit{acute} angle. 3) The proposed GH aims to move the harmonic gradients towards the direction favorable for optimization, rather than those arbitrary gradient directions with acute angles.

\vspace{-2mm}

\subsection{Improved Version: GH++}
\begin{figure}[t]
\centering
 \begin{minipage}{4.3cm}
 \centerline{\includegraphics[scale=0.3]{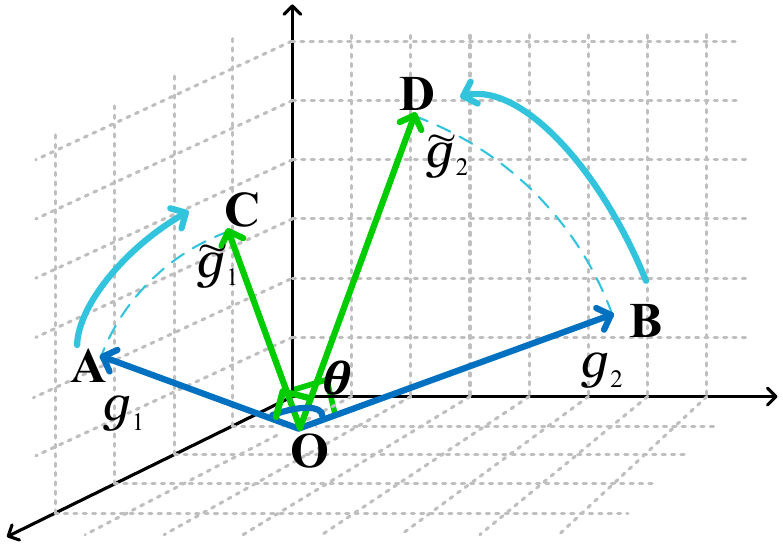}}
 \centerline{(a) GH}
\end{minipage}
\begin{minipage}{4.3cm}
 \centerline{\includegraphics[scale=0.3]{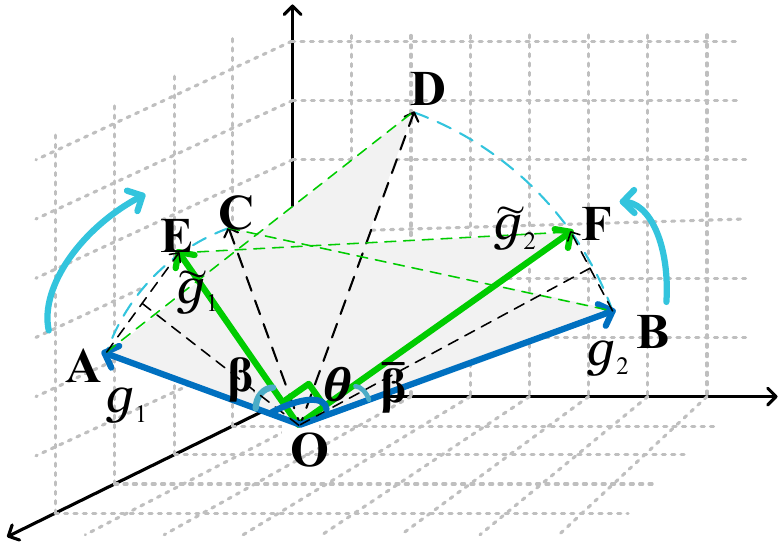}}
 \centerline{(b) GH++}
\end{minipage}
\hfill
  \caption{Illustration of the proposed GH/GH++. The blue and green arrows represent the directions of original gradients and harmonization gradients, respectively.
  Let $\theta$ denote the angle between original gradients, i.e., $g_1$ and $g_2$. (a) GH turns the angle between the original gradients from an obtuse angle to an acute angle. The sum of the gradient deviation angles is $2(\theta - \frac{\pi}{2})$, i.e., $\angle AOC+\angle DOB$. (b) GH++ turns the angle between the original gradients from an obtuse angle to a vertical angle. The sum of the gradient deviation angles is $(\theta - \frac{\pi}{2})$ i.e., $\angle AOE+\angle FOB$.}
  \label{figGH++}
\end{figure}

In this section, we propose an improved version called GH++, which aims to adjust the gradient angle between the two tasks from an obtuse angle to a vertical angle, i.e., making them orthogonal. This is to eliminate the conflict but simultaneously relieve the gradient deviation. Fig. \ref{figGH++} visually illustrates the concepts of the proposed GH and GH++, with a primary focus on scenarios where gradient conflict exists, i.e., when the angle between $g_1$ and $g_2$ is obtuse.

\textbf{Gradient deviation.} For clarity, we designate the original gradients as $g_1=\overrightarrow{OA}$ and $g_2=\overrightarrow{OB}$. Let $\theta$ represent the angle between $g_1$ and $g_2$. As illustrated in Fig. \ref{figGH++} (a), we denote the gradients after applying GH as $\tilde{g}_1=\overrightarrow{OC}$ and $\tilde{g}_2=\overrightarrow{OD}$, where $OC\perp OB$ and $OD\perp OA$. Apparently, the sum of the angles deviated from the original directions is $2(\theta - \frac{\pi}{2})$, i.e., $\angle AOC+\angle DOB$. In other words, although GH can promote the positive correlation between the two gradients, its optimization gradient is seriously deviated from the original gradient. Therefore, we propose an improved version, GH++, to eliminate the conflict and minimize the sum of the gradient deviations. As shown in Fig. \ref{figGH++} (b), GH++ adjusts the gradient angle between the two tasks from an obtuse angle to a vertical angle, i.e., $\angle EOF$. The sum of the gradient deviation angles is $(\theta - \frac{\pi}{2})$ i.e., $\angle AOE+\angle FOB$, which is half of the sum of the gradient deviations of GH.



Specifically, as shown in Fig. \ref{figGH++} (b), let denote the harmonized gradients of GH++ as $\tilde{g}_1=\overrightarrow{OE}$ and $\tilde{g}_2=\overrightarrow{OF}$.  In order to resolve the conflict and relieve the deviation from the original gradient directions, we designate $OE\perp OF$,  i.e., $\angle EOF = \frac{\pi}{2}$, where points $E$ and $F$  move along arcs $\wideparen{AC}$ and $\wideparen{BD}$, respectively. Intuitively, $\bigtriangleup EOF$ forms a rotatable right triangle. We define the direction of rotation from $g_1$ to $\tilde{g}_1$ as positive. $\beta$ represents the rotating angles from $g_1$ to $\tilde{g}_1$, which is a positive angle and $\bar{\beta}$ represents the rotating angles from $g_2$ to $\tilde{g}_2$, which is a negative angle. According to Fig. \ref{figGH++} (b), the harmonization gradients of GH++, i.e., $\tilde{g}_1$ and $\tilde{g}_2$, can be represented as

\begin{equation}\label{gh++1}
\setlength{\abovedisplayskip}{4pt}
\setlength{\belowdisplayskip}{4pt}
\begin{aligned}
\tilde{g}_1 = \overrightarrow{OE} = \overrightarrow{OA}+\overrightarrow{AE}= g_1+ \overrightarrow{AE},\\
\tilde{g}_2 = \overrightarrow{OF} = \overrightarrow{OB}+\overrightarrow{BF}= g_2+ \overrightarrow{BF}.
\end{aligned}
\end{equation}
Since the $\bigtriangleup AOE$ and $\bigtriangleup BOF$ are isosceles triangles, we can obtain $\overrightarrow{AE}$ and $\overrightarrow{BF}$ according to the trigonometric formula, whose expression can be written as
\begin{equation}\label{gh++2}
\setlength{\abovedisplayskip}{4pt}
\setlength{\belowdisplayskip}{4pt}
\begin{aligned}
\overrightarrow{AE} = 2 \cdot \overrightarrow{OA} \cdot \sin \frac{\angle AOE}{2}= 2 \cdot g_1 \cdot \sin \frac{\beta}{2}, \\
\overrightarrow{BF} = 2 \cdot \overrightarrow{OB} \cdot \sin \frac{\angle BOF}{2}= 2 \cdot g_2 \cdot \sin \frac{\bar{\beta }}{2}.
\end{aligned}
\end{equation}
Substituting Eq. (\ref{gh++2}) into Eq. (\ref{gh++1}), we can derive the harmonized gradients of GH++.
\begin{equation}\label{gh++3}
\setlength{\abovedisplayskip}{4pt}
\setlength{\belowdisplayskip}{4pt}
\begin{aligned}
\tilde{g}_1 = (1+2 \cdot \sin \frac{\beta}{2})\cdot g_1,\\
\tilde{g}_2 = (1+2 \cdot \sin \frac{\bar{\beta }}{2})\cdot g_2.
\end{aligned}
\end{equation}
Referring to Figure \ref{figGH++} (b), we have a variable angle $\beta = \angle AOE \in [0, \theta - \frac{\pi}{2}]$, where $\theta = \angle AOB = \arccos \frac{g_1^Tg_2}{|g_1|\cdot|g_2|} \in (\frac{\pi}{2}, \pi]$. To simplify, we can express $\beta$ as follows:
\begin{equation}\label{gh++4}
\beta = \lambda \angle AOC = \lambda (\theta-\frac{\pi}{2}) = \lambda(\arccos \frac{g_1^Tg_2}{|g_1|\cdot|g_2|}-\frac{\pi}{2}),
\end{equation}
where $\lambda \in [0, 1]$ serves as a trade-off parameter, controlling the magnitude of deviation from $\tilde{g}_1$ to $g_1$. When $\lambda = 0$, $\tilde{g}_1 = g_1= \overrightarrow{OA}$ remains unchanged, and $g_2 = \overrightarrow{OD}$. When $\lambda = 1$, $g_1= \overrightarrow{OC}$, and $\tilde{g}_1 = g_2 = \overrightarrow{OB}$ remains constant. Since $OE\perp OF$,  i.e., $\angle EOF = \frac{\pi}{2}$, $\bar \beta$ can be represented as
\begin{equation}\label{gh++50}
\begin{aligned}
\bar{\beta}& =\! \angle BOF =\! - \angle FOB = \!-(\theta \!-\! \frac{\pi}{2}\! - \!\beta) = \beta\!-\!(\theta \!-\! \frac{\pi}{2}).\\
\end{aligned}
\end{equation}
Substituting Eq. (\ref{gh++4}) into Eq. (\ref{gh++50}), we have
\begin{equation}\label{gh++5}
\setlength{\abovedisplayskip}{4pt}
\setlength{\belowdisplayskip}{4pt}
\begin{aligned}
\bar{\beta}& =  \lambda (\theta-\frac{\pi}{2}) + (\theta - \frac{\pi}{2})= (\lambda-1)(\theta - \frac{\pi}{2})\\
&=(\lambda-1)(\arccos \frac{g_1^Tg_2}{|g_1|\cdot|g_2|}-\frac{\pi}{2}).
\end{aligned}
\end{equation}
Finally, substituting Eq. (\ref{gh++4}) and Eq. (\ref{gh++5}) into Eq. (\ref{gh++3}), we can obtain the harmonization gradients of GH++ as follows:
\begin{equation}\label{gh++6}
\setlength{\abovedisplayskip}{4pt}
\setlength{\belowdisplayskip}{4pt}
\begin{aligned}
&\tilde{g}_1 = (1+2 \cdot \sin \frac{\lambda(\arccos \frac{g_1^Tg_2}{|g_1|\cdot|g_2|}-\frac{\pi}{2})}{2})\cdot g_1,\\
&\tilde{g}_2 = (1+2 \cdot \sin \frac{(\lambda-1)(\arccos \frac{g_1^Tg_2}{|g_1|\cdot|g_2|}-\frac{\pi}{2})}{2})\cdot g_2.
\end{aligned}
\end{equation}

Similar to GH, when there is no conflict, i.e., $g_1^Tg_2\geq 0$, no action. The proposed GH++ is indicated as \textbf{Theorem 2}.

\begin{theorem}
For any two different tasks, optimization conflicts can be eliminated by GH++. Define the overall loss function $\mathcal{L} (\Theta)=\mathcal{L}_1 (\Theta)+\mathcal{L}_2 (\Theta)$, composed of two sub-objectives (refer to domain alignment and classification in this paper). Define $g_1$ and $g_2$ as the gradient of $\mathcal{L}_1 (\Theta)$ and $\mathcal{L}_2 (\Theta)$, respectively, then the general expression of the overall harmonized gradient $\tilde{g}$ of the whole loss $\mathcal{L} (\Theta)$ with GH++ can be formulated as:
\begin{equation}\label{gh++7}
\setlength{\abovedisplayskip}{4pt}
\setlength{\belowdisplayskip}{4pt}
\begin{aligned}
\!\!\!\!\!\tilde{g}\!&=\!\tilde{g}_1 + \tilde{g}_2 \\
    \!&=\!(1\!+\!2 \delta(g_1^Tg_2\!<\!0) \sin \frac{\lambda(\arccos \frac{g_1^Tg_2}{|g_1|\cdot|g_2|}\!-\!\frac{\pi}{2})}{2}) g_1\\
    \!&+\!(1\!+\!2 \delta(g_1^Tg_2\!<\!0) \sin \frac{(\lambda\!-\!1)(\arccos \frac{g_1^Tg_2}{|g_1|\cdot|g_2|}\!-\!\frac{\pi}{2})}{2}) g_2.
\end{aligned}
\end{equation}

\end{theorem}
It is worth noting that GH++ presents a more favorable optimization scenario for the original task compared to GH. This advantage arises from the fact that the sum of the gradient deviation from the original direction of GH++ is only half of GH.



\vspace{-2mm}
\subsection{Equivalent Model of UDA with GH/GH++}

By combining the general UDA model (Eq. (\ref{gc}), described in Section \ref{section3.1}) with gradient harmonization strategies (Eq. (\ref{ggg})/Eq. (\ref{gh++7}), a well-balanced UDA model can be trained and implemented. However, the gradient aggregation operator in Eq. (\ref{ggg})/Eq. (\ref{gh++7}) is intricate and has an impact on optimization efficiency. Therefore, we introduce a computation-efficient alternative model that is functionally equivalent to UDA with GH/GH++. For convenience, we can express the proposed gradient harmonization approaches, i.e., Eq. (\ref{ggg})/Eq. (\ref{gh++7}), as follows:
\begin{equation}\label{a1}
\setlength{\abovedisplayskip}{4pt}
\setlength{\belowdisplayskip}{4pt}
    \tilde{g}=\tau_1g_1+\tau_2g_2,
\end{equation}
where $\tau_1$ and $\tau_2$ are constants that can be calculated by using the original gradients $g_1$ and $g_2$. Notably, the gradient $g_1$ of the original loss $\mathcal{L}_1(\Theta)$ and the gradient $g_2$ of the original loss $\mathcal{L}_2(\Theta)$ can be easily computed, only if the UDA model is fixed.  Then, if GH is chosen to resolve the conflict, $\tau_1$ and $\tau_2$ can be calculated as follows.
\begin{equation}\label{a2}
\setlength{\abovedisplayskip}{4pt}
\setlength{\belowdisplayskip}{4pt}
\begin{aligned}
&\tau_1=1-\delta(g_1^Tg_2<0)\frac{g_2^Tg_1}{\lVert g_1 \rVert^2},  \\
&\tau_2=1-\delta(g_1^Tg_2<0)\frac{g_1^Tg_2}{\lVert g_2 \rVert^2}.
\end{aligned}
\end{equation}
If GH++ is used to eliminate the conflict, $\tau_1$ and $\tau_2$ can be calculated as follows.
\begin{equation}\label{a3}
\setlength{\abovedisplayskip}{4pt}
\setlength{\belowdisplayskip}{4pt}
\begin{aligned}
&\tau_1\!=\!(1\!+\!2 \delta(g_1^Tg_2\!<\!0) \sin \frac{\lambda(\arccos \frac{g_1^Tg_2}{|g_1|\cdot|g_2|}\!-\!\frac{\pi}{2})}{2}),\\
&\tau_1\!=\!(1\!+\!2 \delta(g_1^Tg_2\!<\!0) \sin \frac{(\lambda\!-\!1)(\arccos \frac{g_1^Tg_2}{|g_1|\cdot|g_2|}\!-\!\frac{\pi}{2})}{2}).
\end{aligned}
\end{equation}
Ultimately, we can derive the equivalent UDA model embedded GH/GH++ by conducting the integral operation on the gradient in Eq. (\ref{a1}). The overall loss of \textbf{UDA with GH/GH++} model can be represented as

\begin{equation}\label{T}
\setlength{\abovedisplayskip}{4pt}
\setlength{\belowdisplayskip}{4pt}
    \tilde{\mathcal{L}}=\int(\tau_1g_1+\tau_2g_2)d \Theta=\tau_1\mathcal{L}_1(\Theta)+\tau_2\mathcal{L}_2(\Theta),
\end{equation}
where $\mathcal{L}_1(\Theta)$ and $\mathcal{L}_2(\Theta)$ can be the domain alignment loss and classification loss described in Eq. (\ref{align}) and Eq. (\ref{cls}), respectively, for unsupervised domain adaptation.
Finally, the objective function of the \textbf{UDA with GH/GH++} is:
\begin{equation}\label{U}
\setlength{\abovedisplayskip}{4pt}
\setlength{\belowdisplayskip}{4pt}
    \underset{\theta_g,\theta_c}{\min}~~\underset{\theta_d}{\max}~~\tilde{\mathcal{L}}=\tau_1 \mathcal{L}_{dom}(\theta_g,\theta_d)+\tau_2 \mathcal{L}_{cls}(\theta_g,\theta_c).
\end{equation}

It can be seen that the proposed approaches essentially reweight the original loss function $\mathcal{L}_1(\Theta)$ and $\mathcal{L}_2(\Theta)$. The computation of the weights $\tau_1$ and $\tau_2$ needs to use the gradient of the original loss function, as presented in  Eq. (\ref{a2})/Eq. (\ref{a3}). Finally, a balanced and efficient UDA model is formulated. Without loss of generality, when $g_1^Tg_2>0$, we can observe $\tau_1$ =1 and $\tau_2$ =1, and the overall loss function is degenerated into $\tilde{\mathcal{L}}=\mathcal{L}_1(\Theta)+\mathcal{L}_2(\Theta)$, i.e., the general UDA model. We describe the training procedure in Algorithm \ref{algorithm1}.

Notably, the proposed approaches are  plug-and-play modules and can be deployed in almost all models with different losses $\mathcal{L}_1(\Theta)$ and $\mathcal{L}_2(\Theta)$. In order to clearly observe the computational cost of the equivalent model, we discuss the training speed and recognition accuracy before and after applying GH in Table \ref{tabtraining speed}.

\begin{algorithm}[t]
  \SetKwData{Left}{left}\SetKwData{This}{this}\SetKwData{Up}{up}
  \SetKwFunction{Un ion}{Union}\SetKwFunction{FindCompress}{FindCompress}
  \SetKwInOut{Input}{input}\SetKwInOut{Output}{output}
  \SetKwRepeat{Do}{do}{while}
\caption{Balanced UDA with GH/GH++}\label{algorithm1}
\Input{Source samples $\{x_i^{s},y_i^{s}\}_{i=1}^{n_s}$, Target samples $\{x_j^{t}\}_{j=1}^{n_t}$, Optimal parameters $\Theta=\{\theta_g, \theta_{d},\theta_{c}\}$, learning rate $\eta$, max\_iteration, $\lambda$ (if GH++)}
\Output{Optimal parameters $\Theta$}
Initialization {Optimal parameters $\Theta$};
\Repeat{max\_iteration is reached}{
Compute the domain alignment loss $\mathcal{L}_{dom}(\theta_g,\theta_d)$, i.e., $\mathcal{L}_1 (\Theta)$ , and the classification loss $\mathcal{L}_{cls}(\theta_g,\theta_c)$, i.e., $\mathcal{L}_2 (\Theta)$;\\
Compute original gradients $g_1$ and $g_2$;\\
Calculate the inner product of two gradients, i.e., $g_1^{T}g_2$, and the indicator function by Eq. (\ref{aa});\\
Compute $\tau_1$ and $\tau_2$ by Eq. (\ref{a2})/Eq. (\ref{a3});\\
Compute updated total loss $\tilde{\mathcal{L}}=\tau_1 \mathcal{L}_{dom}(\theta_g,\theta_d)+\tau_2 \mathcal{L}_{cls}(\theta_g,\theta_c)$;\\
Update model parameters:\\
~~~~~~~~~~~~$\Theta^{t+1}\longleftarrow\Theta^{t}-\eta\bigtriangledown_{\Theta^{t}}\tilde{\mathcal{L}}$}
\end{algorithm}

\section{Experiments}
\subsection{Datasets}

\textbf{Office-31}~\cite{c:8office31} is a mainstream benchmark dataset for visual domain adaptation, which consists of three distinct domains: Amazon (A), DSLR (D), Webcam (W). It totally contains 4,652 images from 31 categories. We evaluate our method in all 6 different transfer tasks across domains.

\textbf{Office-Home}~\cite{deephashingnetwork} is a more challenging and harder benchmark than Office-31. It contains 15.5K images across 65 object categories from 4 different domains: Artistic images (Ar), Clip Art (Cl), Product images (Pr), and Real-World images (Rw). We evaluate our method in all 12 different transfer tasks across domains.

\textbf{Digits Datasets}. We mainly study three datasets: MNIST (M)~\cite{r:9mnist}, USPS (U)~\cite{r:10usps} and SVHN (S)~\cite{r:11svhn}. MNIST and USPS are two general handwriting recognition datasets involving 10 categories. SVHN is obtained from house numbers in Google Street View images. We conduct experiments in 3 universal tasks, including M$\rightarrow$U, U$\rightarrow$M and S$\rightarrow$M.

\textbf{VisDA-2017}~\cite{r:7visda} is a simulation-to-real dataset for domain adaptation with two distinct domains: synthetic object images rendered from 3D models and real object images. It contains over 280K images across 12 categories.

\textbf{DomainNet}~\cite{peng2019moment} is the largest domain adaptation dataset consisting of about 600K images from six distinct domains, including Clipart (clp), Infograph (inf), Painting (pnt), Quickdraw (qdr), Real (rel), Sketch (skt). There are 48K-172K images categorized into 345 classes per domain. We evaluate our method in all 30 cross-domain tasks.

\begin{table}[t]
\centering
\setlength{\abovecaptionskip}{0.1cm}
\setlength{\belowcaptionskip}{-0.1cm}
\caption{Accuracy ($\%$) on Office-31 for UDA (ResNet50). Avg$^{\ddag}$ represents the mean values except W $\leftrightarrow$ D. Note that TVT and SSRT exploit the ViT backbone pre-trained on ImageNet-21K, while TVT$^\ast$ and SSRT$^\ast$ indicate that their ViT backbone is pre-trained on ImageNet-1K.}
\setlength{\tabcolsep}{0.5mm}{
\resizebox{\linewidth}{!}{
\begin{tabular}{|l|c|c|c|c|c|c|c|c|}
\hline
Method &A $\rightarrow$ W &A $\rightarrow$ D &W $\rightarrow$ A &W $\rightarrow$ D &D $\rightarrow$ A  &D $\rightarrow$ W  &\quad Avg \quad &\ Avg$^{\ddag}\ $\\
\hline
ResNet-50~\cite{c:21} &68.4 &68.9 &60.7 &99.3 &62.5 &96.7 &76.1 &65.1\\
JAN~\cite{long2017deep} &85.4 &84.7 &70.0 &99.8 &68.6 &97.4 &84.3 &77.2\\
CAT~\cite{c:30} &91.1 &90.6 &66.5 &99.6 &70.4 &98.6 &86.1 &79.7\\
ETD~\cite{c:14} &92.1 &88.0 &67.8 &100.0 &71.0 &100.0 &86.2 &79.7\\
TAT~\cite{c:33} &92.5  &93.2  &72.1 &100.0 &73.1 &99.3  &88.4 &82.7\\
TADA~\cite{c:34} &94.3 &91.6 &73.0 &99.8 &72.9 &98.7 &88.4 &83.0\\
SymNets~\cite{c:35} &90.8 &93.9 &72.5 &100.0 &74.6 &98.8 &88.4 &83.0\\
BNM~\cite{c:36} &92.8 &92.9 &73.8 &100.0 &73.5 &98.8 &88.6 &83.3\\
ALDA~\cite{c:18} &95.6 &94.0 &72.5 &100.0 &72.2 &97.7 &88.7 &83.6\\
MDD~\cite{c:37} &94.5 &93.5 &72.2 &100.0 &74.6 &98.4 &88.9 &83.7\\
SAFN~\cite{xu2019larger}&88.8&87.7&67.9&99.8&69.8&98.4&85.4&78.6 \\
Meta~\cite{c:6MetaAlign} &93.9 &91.6 &74.1 &100.0 &73.7 &98.7 &88.7 &83.3\\
SHOT~~\cite{liang2020we}&90.1&94.0&74.3&100.0&74.7&98.4&88.6&83.3\\
CAN~\cite{kang2019contrastive}&94.5&95.0&77.0&100.0&78.0&99.1&90.6&86.1\\
FixBi~\cite{na2021fixbi}&96.1&95.0&79.4&100.0&78.7&99.3&91.4&87.3\\

{\color{black}FGDA \cite{gao2021gradient}} &{\color{black}93.3}&{\color{black}93.2}&{\color{black}72.7}&{\color{black}100.0}&{\color{black}73.2}&{\color{black}99.1}&{\color{black}88.6}&{\color{black}83.1}\\
{\color{black}ParetoDA \cite{liang2021pareto}} &{\color{black}95.0}&{\color{black}95.4}&{\color{black}75.7}&{\color{black}100.0}&{\color{black}77.6}&{\color{black}98.9}&{\color{black}90.4}&{\color{black}85.9}\\
TVT$^\ast$~\cite{yang2023tvt}&95.7&95.4&80.3&100.0&80.6&98.7&91.8&88.0\\
TVT~\cite{yang2023tvt}&96.4&96.4&86.1&100.0&84.9&99.4&93.9&90.9\\

\hline
{\color{black}CDAN~\cite{4CDAN} }&{\color{black} 94.1} & {\color{black}92.9} &{\color{black}69.3} &{\color{black}100.0}&{\color{black}71.0}&{\color{black}98.6}&{\color{black}87.7}&{\color{black}81.8} \\
\rowcolor{orange!10}{\color{black}CDAN+\textbf{GH}} &{\color{black} 94.7} & {\color{black}94.0} &{\color{black}72.0} &{\color{black}100.0}&{\color{black}72.6}&{\color{black}98.6}&{\color{black}88.6}&{\color{black}83.3}\\
\rowcolor{orange!20}{\color{black}CDAN+\textbf{GH++}} &{\color{black} \textbf{95.5}} & {\color{black} \textbf{94.4}} &{\color{black} \textbf{73.3}} &{\color{black}100.0}&{\color{black}\textbf{73.0}}&{\color{black}\textbf{98.8}}&{\color{black}\textbf{89.2}}&{\color{black}\textbf{84.0}}\\
\hline
{\color{black}MCD~\cite{c:2MCD}}&{\color{black} 88.6} & {\color{black}92.2} &{\color{black}69.7} &{\color{black}100.0}&{\color{black}69.5}&{\color{black}98.5}&{\color{black}86.5}&{\color{black}80.0}\\
\rowcolor{orange!10}{\color{black}MCD+\textbf{GH}}&{\color{black} \textbf{91.4}} & {\color{black}92.2} &{\color{black}70.6} &{\color{black}100.0}&{\color{black}69.9}&{\color{black}98.6}&{\color{black}87.1}&{\color{black}81.0}\\
\rowcolor{orange!20}{\color{black}MCD+\textbf{GH++}} &{\color{black} 90.8} & {\color{black}\textbf{92.8}} &{\color{black}\textbf{71.2}} &{\color{black}100.0}&{\color{black}\textbf{70.7}}&{\color{black}\textbf{98.6}}&{\color{black}\textbf{87.4}}&{\color{black}\textbf{81.4}}\\
\hline
{\color{black}DWL~\cite{c:5DWL}}&{\color{black}89.2}&{\color{black}91.2}&{\color{black}69.8}&{\color{black}100.0}&{\color{black}73.1}&{\color{black}99.2} &{\color{black}87.1} &{\color{black}80.8}\\
\rowcolor{orange!10}{\color{black}DWL+\textbf{GH}}&{\color{black}89.2}&{\color{black}91.1} &{\color{black}69.9}& {\color{black}100.0}&{\color{black}\textbf{73.7}}&{\color{black}99.3}&{\color{black}87.2}&{\color{black}81.0}\\
\rowcolor{orange!20}{\color{black}DWL+\textbf{GH++}} &{\color{black} \textbf{90.4}} & {\color{black}\textbf{91.4}} &{\color{black}\textbf{70.5}} &{\color{black}100.0}&{\color{black}\textbf{73.7}}&{\color{black}\textbf{99.3}}&{\color{black}\textbf{87.5}}&{\color{black}\textbf{81.5}}\\
\hline
{\color{black}GVB~\cite{c:3GVB}} &{\color{black} 94.8} & {\color{black}95.0} &{\color{black}73.7} &{\color{black}100.0}&{\color{black}73.4}&{\color{black}98.7}&{\color{black}89.3}&{\color{black}84.2}\\
\rowcolor{orange!10}{\color{black}GVB+\textbf{GH}} &{\color{black} 94.9} & {\color{black}\textbf{95.4}} &{\color{black}74.0} &{\color{black}100.0}&{\color{black}75.3}&{\color{black}99.0}&{\color{black}89.8}&{\color{black}84.9}\\
\rowcolor{orange!20}{\color{black}GVB+\textbf{GH++}} &{\color{black} \textbf{95.1}} & {\color{black}95.2} &{\color{black}\textbf{74.1}} &{\color{black}100.0}&{\color{black}\textbf{75.8}}&{\color{black}\textbf{99.1}}&{\color{black}\textbf{89.9}}&{\color{black}\textbf{85.0}}\\

\hline
{\color{black}SSRT$^\ast$~\cite{sun2022safe}}&{\color{black}97.7} &{\color{black}98.6} &{\color{black}82.2} &{\color{black}100.0} &{\color{black}83.5} &{\color{black}99.2} &{\color{black}93.5} &{\color{black}90.5}\\
\rowcolor{orange!10}{\color{black}SSRT$^\ast$ +\textbf{GH}}&{\color{black}98.5} &{\color{black}98.6} &{\color{black}\textbf{83.6}} &{\color{black}100.0} &{\color{black}84.2} &{\color{black}\textbf{99.3}} &{\color{black}94.0} &{\color{black}91.2}\\
\rowcolor{orange!20}{\color{black}SSRT$^\ast$ +\textbf{GH++}}&{\color{black}\textbf{98.9}} &{\color{black} \textbf{98.8}} &{\color{black}83.3} &{\color{black}100.0} &{\color{black}\textbf{84.6}} &{\color{black}\textbf{99.3}} &{\color{black}\textbf{94.1}} &{\color{black}\textbf{91.4}}\\
\hline
{\color{black}SSRT~\cite{sun2022safe}}&{\color{black}98.4} &{\color{black}99.2} &{\color{black}84.0} &{\color{black}100.0} &{\color{black}85.3} &{\color{black}99.2} &{\color{black}94.4} &{\color{black}91.7}\\
\rowcolor{orange!10}{\color{black}SSRT+\textbf{GH}}&{\color{black}\textbf{98.9}} &{\color{black}99.2} &{\color{black}84.2} &{\color{black}100.0} &{\color{black}85.6} &{\color{black}99.2} &{\color{black}94.5} &{\color{black}92.0}\\
\rowcolor{orange!20}{\color{black}SSRT+\textbf{GH++}}&{\color{black}98.7} &{\color{black} \textbf{99.6}} &{\color{black}\textbf{84.6}} &{\color{black}100.0} &{\color{black}\textbf{85.8}} &{\color{black}\textbf{99.4}} &{\color{black}\textbf{94.7}} &{\color{black}\textbf{92.2}}\\

\hline
\end{tabular}}}
\label{tab1}
\end{table}

\begin{table*}
\centering
\setlength{\abovecaptionskip}{0.cm}
\setlength{\belowcaptionskip}{-0.2cm}
\caption{Accuracy ($\%$) on Office-Home for UDA (ResNet-50). Avg$^{\ddag}$ represents the mean values except Pr $\leftrightarrow$ Rw. Note that TVT and SSRT exploit the ViT backbone pre-trained on ImageNet-21K, while TVT$^\ast$ and SSRT$^\ast$ indicate the ViT backbone is pre-trained on ImageNet-1K.}
\setlength{\tabcolsep}{0.0mm}{
\resizebox{\linewidth}{!}{
\begin{tabular}{|l|c|c|c|c|c|c|c|c|c|c|c|c|c|c|}
\hline
Method &Ar $\rightarrow$ Cl &Ar $\rightarrow$ Pr &Ar $\rightarrow$ Rw &Cl $\rightarrow$ Ar &Cl $\rightarrow$ Pr &Cl $\rightarrow$ Rw &Pr $\rightarrow$ Ar &Pr $\rightarrow$ Cl &Pr $\rightarrow$ Rw &Rw $\rightarrow$ Ar &Rw $\rightarrow$ Cl &Rw $\rightarrow$ Pr & \quad Avg \quad \ & \quad Avg$^{\ddag}\quad $\\ 
\hline
ResNet-50~\cite{c:21} &34.9 &50.0 &58.0 &37.4 &41.9 &46.2 &38.5 &31.2 &60.4 &53.9 &41.2 &59.9 &46.1 &43.3 \\
JAN~\cite{long2017deep} &45.9 &61.2 &68.9 &50.4 &59.7 &61.0  &45.8 &43.4 &70.3 &63.9 &52.4 &76.8 &58.3 &55.3\\
DAN~\cite{c:22} &43.6 &57.0 &67.9 &45.8 &56.5 &60.4  &44.0 &43.6 &67.7 &63.1 &51.5 &74.3 &56.3 &53.3\\
DANN~\cite{c:23} &45.6 &59.3 &70.1 &47.0 &58.5 &60.9 &46.1 &43.7 &68.5 &63.2 &51.8 &76.8 &57.6 &54.6\\
TAT~\cite{c:33} &51.6 &69.5 &75.4 &59.4 &69.5 &68.6  &59.5 &50.5 &76.8 &70.9 &56.6 &81.6 &65.8 &63.2\\
TADA~\cite{c:34} &53.1 &72.3 &77.2 &59.1 &71.2 &72.1  &59.7 &53.1 &78.4 &72.4 &60.0 &82.9 &67.6 &65.0\\
SymNets~\cite{c:35} &47.7 &72.9 &78.5 &64.2 &71.3 &74.2 &64.2 &48.8&79.5&74.5&52.6&82.7&67.6&64.9\\
ALDA~\cite{c:18} &53.7 &70.1 &76.4 &60.2 &72.6 &71.5  &56.8 &51.9 &77.1 &70.2 &56.3 &82.1 &66.6 &64.0\\
SAFN~\cite{xu2019larger}&58.9&76.2&81.4&70.4&73.0&77.8&72.4&55.3&80.4&75.8&60.4&79.9&71.8&70.2\\
SHOT~~\cite{liang2020we}&57.1&78.1&81.5&68.0&78.2&78.1&67.4&54.9&82.2&73.3&58.8&84.3&71.8&69.5\\
FixBi~\cite{na2021fixbi}&58.1&77.3&80.4&67.7&79.5&78.1&65.8&57.9&81.7&76.4&62.9&86.7&72.7&70.4\\
{\color{black}FGDA \cite{gao2021gradient}} &{\color{black}52.3}&{\color{black}77.0}&{\color{black}78.2}&{\color{black}64.6}&{\color{black}75.5}&{\color{black}73.7}&{\color{black}64.0}
&{\color{black}49.5}&{\color{black}80.7}&{\color{black}70.1}&{\color{black}52.3}&{\color{black}81.6}&{\color{black}68.3}&{\color{black}65.7}\\
{\color{black}ParetoDA \cite{liang2021pareto} } &{\color{black}56.8}&{\color{black}75.9}&{\color{black}80.5}&{\color{black}64.4}&{\color{black}73.5}&{\color{black}73.7}&{\color{black}65.6}
&{\color{black}55.5}&{\color{black}81.3}&{\color{black}75.2}&{\color{black}61.1}&{\color{black}83.9}&{\color{black}70.6}&{\color{black}68.2}\\
TVT$^\ast$~\cite{yang2023tvt}&67.1&83.5&87.3&77.4&85.0&85.6&75.6&64.9&86.6&79.1&67.2&88.0&78.9&77.3\\
TVT~\cite{yang2023tvt}&74.9&86.8&89.5&82.8&88.0&88.3&79.8&71.9&90.1&85.5&74.6&90.6&83.6&82.2\\
\hline
{\color{black}CDAN~\cite{4CDAN} } &{\color{black}50.7} &{\color{black}70.6} &{\color{black}76.0} &{\color{black}57.6} &{\color{black}70.0} &{\color{black}70.0} &{\color{black}57.4} &{\color{black}50.9} &{\color{black}77.3} &{\color{black}70.9} &{\color{black}56.7} &{\color{black}81.6} &{\color{black}65.8} &{\color{black}63.1}\\
\rowcolor{orange!10}{\color{black}CDAN+\textbf{GH} } &{\color{black}52.4} &{\color{black}\textbf{74.2}} &{\color{black}78.7} &{\color{black}62.3} &{\color{black}72.3} &{\color{black}73.4} &{\color{black}61.6} &{\color{black}51.4} &{\color{black}\textbf{80.7}} &{\color{black}73.1} &{\color{black}57.3} &{\color{black}82.3} &{\color{black}68.3} &{\color{black}65.7}\\
\rowcolor{orange!20}{\color{black}CDAN+\textbf{GH++}}  &{\color{black}\textbf{54.3}} &{\color{black}73.9} &{\color{black}\textbf{78.7}} &{\color{black}\textbf{62.4}} &{\color{black}\textbf{73.7}} &{\color{black}\textbf{73.6}} &{\color{black}\textbf{61.8}} &{\color{black}\textbf{53.5}} &{\color{black}80.6} &{\color{black}\textbf{73.3}} &{\color{black}\textbf{57.7}} &{\color{black}\textbf{83.5}} &{\color{black}\textbf{68.9}} &{\color{black}\textbf{66.3}}\\
\hline
{\color{black}MCD~\cite{c:2MCD}}&{\color{black}48.9} &{\color{black}68.3} &{\color{black}74.6} &{\color{black}61.3} &{\color{black}67.6} &{\color{black}68.8} &{\color{black}57.0} &{\color{black}47.1} &{\color{black}75.1} &{\color{black}69.1} &{\color{black}52.2} &{\color{black}79.6} &{\color{black}64.1} &{\color{black}61.5}\\
\rowcolor{orange!10}{\color{black}MCD+\textbf{GH}}&{\color{black}52.1} &{\color{black}71} &{\color{black}77.6} &{\color{black}\textbf{62.7}} &{\color{black}68.7} &{\color{black}\textbf{70.5}} &{\color{black}60.9} &{\color{black}51.9} &{\color{black}\textbf{78.8}} &{\color{black}74.2} &{\color{black}60.0} &{\color{black}\textbf{82.1}} &{\color{black}67.5} &{\color{black}65.0}\\
\rowcolor{orange!20}{\color{black}MCD+\textbf{GH++} } &{\color{black}\textbf{52.2}} &{\color{black}\textbf{72.3}} &{\color{black}\textbf{77.6}} &{\color{black}62.6} &{\color{black}\textbf{69.8}} &{\color{black}70.4} &{\color{black}\textbf{62.8}} &{\color{black}\textbf{53.1}} &{\color{black}78.6} &{\color{black}\textbf{74.7}} &{\color{black}\textbf{60.3}} &{\color{black}81.9} &{\color{black}\textbf{68.0}} &{\color{black}\textbf{65.6}}\\
\hline
{\color{black}DWL~\cite{c:5DWL}}&{\color{black}46.6} &{\color{black}67.9} &{\color{black}74.6} &{\color{black}57.7} &{\color{black}66.4} &{\color{black}68.4} &{\color{black}58.5} &{\color{black}46.0} &{\color{black}76.1} &{\color{black}69.9} &{\color{black}51.8} &{\color{black}78.7} &{\color{black}63.6} &{\color{black}60.8}\\
\rowcolor{orange!10}{\color{black}DWL+\textbf{GH}}&{\color{black}47.2} &{\color{black}69.5} &{\color{black}75.9} &{\color{black}59.4} &{\color{black}\textbf{67.3}} &{\color{black}68.4} &{\color{black}59.5} &{\color{black}46.9} &{\color{black}\textbf{76.6}} &{\color{black}70.1} &{\color{black}51.9} &{\color{black}\textbf{79.7}} &{\color{black}64.4} &{\color{black}61.6}\\
\rowcolor{orange!20}{\color{black}DWL+\textbf{GH++}}&{\color{black}\textbf{47.5}} &{\color{black}\textbf{69.5}} &{\color{black}\textbf{76.1}} &{\color{black}\textbf{59.5}} &{\color{black}67.2} &{\color{black}\textbf{69.1}} &{\color{black}\textbf{60.2}} &{\color{black}47.0} &{\color{black}76.4} &{\color{black}\textbf{70.5}} &{\color{black}\textbf{53.0}} &{\color{black}79.6} &{\color{black}\textbf{64.6}} &{\color{black}\textbf{62.0}}\\
\hline
{\color{black}GVB~\cite{c:3GVB}} &{\color{black}57.0} &{\color{black}74.7} &{\color{black}79.8} &{\color{black}64.6} &{\color{black}74.1} &{\color{black}74.6} &{\color{black}65.2} &{\color{black}55.1} &{\color{black}81.0} &{\color{black}74.6} &{\color{black}59.7} &{\color{black}84.3} &{\color{black}70.4} &{\color{black}67.9}\\
\rowcolor{orange!10}{\color{black}GVB+\textbf{GH}} &{\color{black}57.3} &{\color{black}75.4} &{\color{black}79.9} &{\color{black}\textbf{65.3}} &{\color{black}74.5} &{\color{black}75.1} &{\color{black}\textbf{65.9}} &{\color{black}55.4} &{\color{black}\textbf{81.6}} &{\color{black}74.6} &{\color{black}60.1} &{\color{black}84.3} &{\color{black}70.8} &{\color{black}68.3}\\
\rowcolor{orange!20}{\color{black}GVB+\textbf{GH++}}&{\color{black}\textbf{57.7}} &{\color{black}\textbf{75.5}} &{\color{black}\textbf{80.2}} &{\color{black}65.1} &{\color{black}\textbf{75.2}} &{\color{black}\textbf{75.4}} &{\color{black}65.6} &{\color{black}\textbf{55.4}} &{\color{black}81.2} &{\color{black}\textbf{74.7}} &{\color{black}\textbf{60.4}} &{\color{black}\textbf{84.4}} &{\color{black}\textbf{70.9}} &{\color{black}\textbf{68.5}}\\
\hline
{\color{black}SSRT$^\ast$~\cite{sun2022safe}}&{\color{black}75.2} &{\color{black}89.0} &{\color{black}91.1} &{\color{black}85.1} &{\color{black}88.3} &{\color{black}90.0} &{\color{black}85.0} &{\color{black}74.2} &{\color{black}91.3} &{\color{black}85.7} &{\color{black}78.6} &{\color{black}91.8} &{\color{black}85.4} &{\color{black}84.2}\\
\rowcolor{orange!10}{\color{black}SSRT$^\ast$+\textbf{GH}}&{\color{black}75.4} &{\color{black}\textbf{90.0}} &{\color{black}91.3} &{\color{black}85.5} &{\color{black}89.1} &{\color{black}90.1} &{\color{black}\textbf{85.8}} &{\color{black}75.1} &{\color{black}91.3} &{\color{black}86.8} &{\color{black}78.9} &{\color{black}92.2} &{\color{black}85.9} &{\color{black}84.8}\\
\rowcolor{orange!20}{\color{black}SSRT$^\ast$+\textbf{GH++}}&{\color{black}\textbf{75.5}} &{\color{black}89.6} &{\color{black}\textbf{91.4}} &{\color{black}\textbf{86.1}} &{\color{black}\textbf{89.4}} &{\color{black}\textbf{90.3}} &{\color{black}85.4} &{\color{black}\textbf{75.2}} &{\color{black}91.3} &{\color{black}\textbf{86.9}} &{\color{black}\textbf{79.7}} &{\color{black}\textbf{92.4}} &{\color{black}\textbf{86.1}} &{\color{black}\textbf{85.0}}\\
\hline
{\color{black}SSRT~\cite{sun2022safe}}&{\color{black}76.0} &{\color{black}88.7} &{\color{black}91.2} &{\color{black}85.2} &{\color{black}88.7} &{\color{black}90.2} &{\color{black}85.0} &{\color{black}74.6} &{\color{black}91.4} &{\color{black}86.3} &{\color{black}78.4} &{\color{black}92.5} &{\color{black}85.7} &{\color{black}84.4}\\
\rowcolor{orange!10}{\color{black}SSRT+\textbf{GH}}&{\color{black}\textbf{76.5}} &{\color{black}89.5} &{\color{black}91.3} &{\color{black}85.8} &{\color{black}89.2} &{\color{black}90.2} &{\color{black}85.2} &{\color{black}\textbf{75.3}} &{\color{black}\textbf{91.8}} &{\color{black}87.1} &{\color{black}78.7} &{\color{black}\textbf{92.8}} &{\color{black}86.1} &{\color{black}84.9}\\
\rowcolor{orange!20}{\color{black}SSRT+\textbf{GH++}}&{\color{black}76.3} &{\color{black}\textbf{89.7}} &{\color{black}\textbf{91.6}} &{\color{black}\textbf{86.5}} &{\color{black}\textbf{89.4}} &{\color{black}\textbf{90.4}} &{\color{black}\textbf{85.4}} &{\color{black}\textbf{75.3}} &{\color{black}91.3} &{\color{black}\textbf{87.5}} &{\color{black}\textbf{79.2}} &{\color{black}92.7} &{\color{black}\textbf{86.3}} &{\color{black}\textbf{85.1}}\\

\hline
\end{tabular}}}
\label{tab2}
\end{table*}

\subsection{Implementation Details}
We compare our method with the following state-of-the-art unsupervised domain adaptation methods:~\textbf{ResNet}~\cite{c:21}, \textbf{DAN} (Deep Adaptation Networks)~\cite{c:22}, \textbf{DANN} (Domain-adversarial Neural Networks)~\cite{c:23}, \textbf{JAN} (Joint Adaptation Networks)~\cite{long2017deep}, \textbf{DRCN} (Deep Reconstruction-Classification Networks)~\cite{c:26}, \textbf{CoGAN} (Coupled Generative Adversarial Networks)~\cite{c:27}, \textbf{ADDA} (Adversarial Discriminative Domain Adaptation)~\cite{c:28}, \textbf{CyCADA} ( Cycle-consistent Adversarial Domain Adaptation)~\cite{c:29}, \textbf{CAT} (Cluster Alignment with a Teacher)~\cite{c:30}, \textbf{TPN} (Transferrable prototypical networks)~\cite{c:31}, \textbf{LWC} (Light-weight Calibrator)~\cite{c:32}, \textbf{ETD} (Enhanced Transport Distance)~\cite{c:14}, \textbf{TAT} (Transferable Adversarial Training)~\cite{c:33}, \textbf{TADA} (Transferable Attention for Domain Adaptation)~\cite{c:34}, \textbf{SymNets} (Symmetric Networks)~\cite{c:35}, \textbf{BNM} (Batch Nuclear-norm Maximization)~\cite{c:36}, \textbf{ALDA} (Adversarial-learned Loss for Domain Adaptation)~\cite{c:18}, \textbf{MIMTFL} (Mutual Information Maximisation
and Transferable Feature Learning)~\cite{gao2020reducing}, \textbf{Meta} (Metaalign)~\cite{c:6MetaAlign}, \textbf{MDD} (Margin Disparity Discrepancy)~\cite{c:37}, \textbf{CDAN}~\cite{4CDAN}, \textbf{MCD} (Maximum Classification Discrepancy)~\cite{c:2MCD}, \textbf{GVB} (Gradually Vanishing Bridge)~\cite{c:3GVB} and \textbf{DWL} (Dynamic Weighted Learning)~\cite{c:5DWL}, \textbf{SAFN} (Stepwise Adaptive Feature Norm) \cite{xu2019larger}, \textbf{SWD} (Sliced Wasserstein Discrepancy) \cite{lee2019sliced}, \textbf{DMRL} (Dual Mixup Regularized Learning) \cite{wu2020dual}, \textbf{CAN} (Contrastive Adaptation Network) \cite{kang2019contrastive}, \textbf{FixBi}~\cite{na2021fixbi}, \textbf{FGDA} (Feature Gradient Distribution Alignment) \cite{gao2021gradient} and \textbf{ParetoDA} (Pareto Domain Adaptation) \cite{liang2021pareto}, CGDM (Cross-domain Gradient Discrepancy Minimization) \cite{CGMD2021}, \textbf{SSRT} (Safe Self-Refinement for Transformer-based domain adaptation)~\cite{sun2022safe} and \textbf{TVT} (Transferable Vision Transformer)~\cite{yang2023tvt}.  \textbf{SSRT} and \textbf{TVT} are transformer-based approaches, while other approaches are CNN-based. We use the results in their original papers for fair comparison. When there is no relevant experimental result in the original papers, we implement them according to their released source codes.

Furthermore, we investigate the effect of GH/GH++ on five popular UDA approaches, including four CNN-based approaches (i.e., \textbf{CDAN}~\cite{4CDAN}, \textbf{MCD}~\cite{c:2MCD}, \textbf{GVB}~\cite{c:3GVB} and \textbf{DWL}~\cite{c:5DWL}) and one transformer-based approach (i.e., \textbf{SSRT}~\cite{sun2022safe}). For CNN-based approaches, we use pre-trained \emph{ResNet-50}~\cite{c:21} on ImageNet-1K~\cite{43} as backbone network for Office-31 and Office-Home, and adopt the pre-trained \emph{ResNet-101}~\cite{c:21} as backbone network for VisDA-2017 and DomainNet. We employ \emph{LeNet}~\cite{lecun1998gradient} as backbone for Digits datasets. For transformer-based approach, SSRT uses the \emph{ViT-base} with $16 \times 16$ patch size pre-trained on ImageNet-21K and ImageNet-1K (denoted by SSRT$^\ast$) as the transformer backbones, respectively. In terms of optimization, we follow the original protocol of the baselines. Note that, in this paper, \textbf{CDAN} means CDAN+E in~\cite{4CDAN}) and \textbf{GVB} refers to GVB-GD in~\cite{c:3GVB}. For GH++, we empirically set $\lambda$ to 0.5.
MCD achieves domain alignment by the game between two classifiers and the generator. 
Therefore, the gradient of the discrepancy loss between the two classifiers is just the gradient of domain alignment loss. We implement our method in PyTorch by following UDA protocol~\cite{long2017deep} that source domain is labeled and target domain is unlabeled. 
For fair comparisons, we independently run all experiments five times and report the average target accuracy. 

\subsection{Results on UDA}



\begin{table}[h]
\renewcommand\arraystretch{1}
\setlength{\tabcolsep}{1.0mm}
\centering
\setlength{\abovecaptionskip}{0.cm}
\setlength{\belowcaptionskip}{-0.1cm}
\caption{Accuracy ($\%$) on Digits Datasets for UDA (LeNet). $^\ast$ indicates that the backbone is ViT pre-trained on ImageNet-1K.}
\small
\setlength{\tabcolsep}{3mm}{
\scalebox{0.9}{
\begin{tabular}{|l|c|c|c|c|}
\hline
Method &M $\rightarrow$ U &U $\rightarrow$ M &S $\rightarrow$ M & \ Avg\ \     \\
\hline
DAN~\cite{c:22} &80.3 &77.8 &73.5 &77.2\\
DRCN~\cite{c:26} &91.8 &73.7 &82.0 &82.5\\
CoGAN~\cite{c:27} &91.2  &89.1  &-  &-\\
ADDA~\cite{c:28} &89.4  &90.1  &76.0  &85.2\\
CyCADA~\cite{c:29} &95.6  &96.5  &90.4  &94.2\\
CAT~\cite{c:30} &90.6  &80.9  &98.1 &89.9\\
TPN~\cite{c:31} &92.1  &94.1  &93.0  &93.1\\
LWC~\cite{c:32} &95.6  &97.1 &97.1  &96.6\\
ETD~\cite{c:14} &96.4 &96.3  &97.9  &96.9\\
SWD~\cite{lee2019sliced}&98.1&97.1&98.9&98.0\\
SHOT~~\cite{liang2020we}&91.9&96.8&89.6&92.8\\
TVT$^\ast$~\cite{yang2023tvt}&97.7&98.9&98.0&98.2\\
TVT~\cite{yang2023tvt}&98.2&99.4&99.0&98.9\\

\hline

{\color{black}CDAN~\cite{4CDAN}} &{\color{black}95.6} &{\color{black}98.0} &{\color{black}89.2} &{\color{black}94.3}\\
\rowcolor{orange!10}{\color{black}CDAN+\textbf{GH}} &{\color{black}\textbf{96.8}} &{\color{black}98.1} &{\color{black}90.6} &{\color{black}\textbf{95.2}} \\
\rowcolor{orange!20}{\color{black}CDAN+\textbf{GH++} }&{\color{black}96.5}&{\color{black}\textbf{98.3}}&{\color{black}9\textbf{0.7}}&{\color{black}\textbf{95.2}}\\
\hline
MCD~\cite{c:2MCD} &94.2 &94.1 &96.2 &94.8\\
\rowcolor{orange!10}MCD+\textbf{GH} &{96.7} &{96.8}&\textbf{97.5} &{97.0}\\
\rowcolor{orange!20}{\color{black}MCD+\textbf{GH++}}&{\color{black}\textbf{97.3}}&{\color{black}\textbf{97.1}}&{\color{black}97.1}&{\color{black}\textbf{97.2}}\\
\hline
DWL~\cite{c:5DWL} &97.3 &97.4 &98.1 &97.6\\
\rowcolor{orange!10}DWL+\textbf{GH} &98.7  &98.5 &98.8 &98.7 \\
\rowcolor{orange!20}{\color{black}DWL+\textbf{GH++}}&{\color{black}\textbf{98.9}}&{\color{black}\textbf{98.6}}&{\color{black}\textbf{99.1}}&{\color{black}\textbf{98.9}}\\
\hline
{\color{black}GVB~\cite{c:3GVB}}&{\color{black}96.3}&{\color{black}95.1}&{\color{black}90.0}&{\color{black}93.8}\\%
\rowcolor{orange!10}{\color{black}GVB+\textbf{GH}} &{\color{black}96.7}&{\color{black}95.4}&{\color{black}91.0}&{\color{black}94.4}\\
\rowcolor{orange!20}{\color{black}GVB+\textbf{GH++}}&{\color{black}\textbf{96.9}}&{\color{black}\textbf{95.5}}&{\color{black}\textbf{91.7}}&{\color{black}\textbf{94.7}}\\
\hline
{\color{black}SSRT$^\ast$~\cite{sun2022safe}}&{\color{black}98.4}&{\color{black}99.4}&{99.1}&{\color{black}99.0}\\%
\rowcolor{orange!10}{\color{black}SSRT$^\ast$+\textbf{GH}} &{\color{black}\textbf{99.5}} &{\color{black}99.4} &{\color{black}\textbf{99.2}} &{\color{black}\textbf{99.3}}\\
\rowcolor{orange!20}{SSRT$^\ast$+\textbf{GH++}}&{99.4}&{\textbf{99.5}}&{\textbf{99.2}}&{\textbf{99.3}}\\
\hline
{\color{black}SSRT~\cite{sun2022safe}}&{\color{black}98.6}&{\color{black}99.4}&{99.2}&{\color{black}99.0}\\%
\rowcolor{orange!10}{\color{black}SSRT+\textbf{GH}} &{\color{black}98.9} &{\color{black}99.4} &{\color{black}\textbf{99.3}} &{\color{black}99.2}\\
\rowcolor{orange!20}{SSRT+\textbf{GH++}}&{\textbf{99.2}}&{\textbf{99.5}}&{\textbf{99.3}}&{\textbf{99.3}}\\
\hline
\end{tabular}}}
\label{tab3}
\end{table}

Tables \ref{tab1}, \ref{tab2}, \ref{tab3}, \ref{tab4} and \ref{tab5} present evaluation results on Office-31, Office-Home, VisDA-2017, Digits and DomainNet, respectively. Generally, the transformer-based results, such as TVT and SSRT, are much better than CNN-based results, which has been validated in previous work. For TVT and SSRT, the ViT backbone pre-trained on ImageNet-1K is slightly inferior than ImageNet-21K (i.e., TVT$^\ast$\&TVT, SSRT$^\ast$\&SSRT). Considering that the CNNs are pre-trained on ImageNet-1K, we take SSRT$^\ast$ for fair analysis as default in the following.  

In order to demonstrate the effectiveness and universality of the proposed approaches in UDA, we investigate the effect of GH/GH++ on five popular UDA approaches, including CNN-based approaches (i.e., CDAN, MCD, GVB and DWL) and transformer-based approach (i.e., SSRT). We can observe that the proposed approaches can further improve the baseline, whether based on transformer or CNN. This is attributed to the proposed approaches in alleviating gradient conflict between domain alignment task and classification task in optimization. Meanwhile, it verifies that the proposed approaches can be easily plugged and played in the existing UDA methods. In addition, GH++ usually performs better than GH. The reason is that during removing the gradient conflict between the two tasks, GH++ minimizes the gradient deflection, such that the two tasks can evolve harmoniously during joint training while preserving their individual task-specific optimality.

Specifically, as shown in Table \ref{tab1}, when GH/GH++ is applied, CDAN, MCD, GVB, DWL and SSRT outperform the original models by 1.5\%/2.2\%, 1.0\%/1.4\%,  0.2\%/0.7\%, 0.7\%/0.8\% and 0.7\%/0.9\% on Avg$^{\ddag}$ of Office-31, respectively.
As shown in Table \ref{tab2}, CDAN, MCD, GVB, DWL and SSRT after applying GH/GH++ outperform the original models by 2.6\%/3.2\%, 4.5\%/5.1\%, 0.8\%/1.2\%, 0.4\%/0.6\% and 0.6\%/0.8\% on Avg$^{\ddag}$ of Office-Home, respectively. As shown in Table \ref{tab3}, we can observe that CDAN, MCD, GVB, DWL and SSRT after applying GH/GH++ outperform the original models by 0.9\%/0.9\%, 2.2\%/2.4\%,  1.1\%/1.3\%, 0.6\%/0.9\% and 0.3\%/0.3\% on Avg of Digits, respectively.
As shown in Table \ref{tab4}, CDAN, MCD, GVB, DWL and SSRT after applying GH/GH++ outperform the original models by 1.4\%/3.9\%, 4.1\%/4.5\%, 0.5\%/0.8\%, 0.7\%/0.3\% and 0.7\%/0.6\% on VisDA-2017, respectively.
From the results in Table \ref{tab5}, we can observe that CDAN, MCD, GVB, DWL and SSRT after applying GH/GH++ outperform the original models by 3.2\%/3.7\%, 7.1\%/7.3\%, 0.7\%/1.0\%, 1.0\%/1.3\% and 1.1\%/1.4\% on Avg of DomainNet, respectively. The proposed approaches generally improve CDAN and MCD more than other baselines because the original models of CDAN and MCD have more obvious optimization conflicts than other baselines. Fig. \ref{figfig} also confirms this perspective. Therefore, the proposed approaches have better performance when the inherent gradient conflict is serious. Besides, it is worth noting that the proposed approaches usually achieve greater performance gains on large-scale datasets, i.e., VisDA-2017 and DomainNet, which further indicate the effectiveness and versatility of our method.

\begin{table}[t]
\renewcommand\arraystretch{1}
\centering
\setlength{\abovecaptionskip}{0.cm}
\setlength{\belowcaptionskip}{-0.2cm}
\caption{Accuracy ($\%$) on VisDA-2017 for UDA (ResNet-101). $^\ast$ indicates that the backbone is ViT pre-trained on ImageNet-1K.}
\setlength{\tabcolsep}{8mm}{
\scalebox{0.96}{
\begin{tabular}{|l|c|}
\hline
Method &Synthetic $\rightarrow$ Real (Avg)\\
\hline
ResNet-101~\cite{c:21} &52.4\\
JAN~\cite{long2017deep} &61.6\\
DAN~\cite{c:22} &61.1\\
DANN~\cite{c:23} &57.4\\
SAFN~\cite{xu2019larger}&76.1\\
SWD~\cite{lee2019sliced}&76.4\\
DMRL~\cite{wu2020dual}&75.5\\
SHOT~~\cite{liang2020we}&82.9\\
CAN~\cite{kang2019contrastive}&87.2\\
FixBi~\cite{na2021fixbi}&87.2\\
{\color{black}CGDM \cite{CGMD2021}}&{\color{black}82.3}\\
{\color{black}ParetoDA \cite{liang2021pareto}} &{\color{black}83.2}\\
TVT$^\ast$~\cite{yang2023tvt}&85.1\\
TVT~\cite{yang2023tvt}&86.7\\
\hline
{\color{black}CDAN~\cite{4CDAN} }&{\color{black}73.9}\\
\rowcolor{orange!10}{\color{black}CDAN+\textbf{GH} }&{\color{black}75.3}\\
\rowcolor{orange!20}{\color{black}CDAN+\textbf{GH++} }&{\color{black}\textbf{77.8}}\\
\hline
{\color{black}MCD~\cite{c:2MCD}}&{\color{black}71.9}\\
\rowcolor{orange!10}{\color{black}MCD+\textbf{GH}}&{\color{black}76.0}\\
\rowcolor{orange!20}{\color{black}MCD+\textbf{GH++}}&{\color{black}\textbf{76.4}}\\
\hline
{\color{black}DWL~\cite{c:5DWL}}&{\color{black}77.1}\\
\rowcolor{orange!10}{\color{black}DWL+\textbf{GH}}&{\color{black}77.6}\\
\rowcolor{orange!20}{\color{black}DWL+\textbf{GH++}}&{\color{black}\textbf{77.9}}\\
\hline
{\color{black}GVB}~\cite{c:3GVB} &{\color{black}77.0}\\
\rowcolor{orange!10}{\color{black}GVB+\textbf{GH}} &{\color{black}77.7}\\
\rowcolor{orange!20}{\color{black}GVB+\textbf{GH++}}&{\color{black}\textbf{78.3}}\\
\hline

{\color{black}SSRT$^\ast$~\cite{sun2022safe} }&{\color{black}88.8}\\
\rowcolor{orange!10}{\color{black}SSRT$^\ast$+\textbf{GH}} &{\color{black}\textbf{89.5}}\\
\rowcolor{orange!20}{\color{black}SSRT$^\ast$+\textbf{GH++}} &{\color{black}89.4}\\
\hline
{\color{black}SSRT~\cite{sun2022safe} }&{\color{black}88.9}\\
\rowcolor{orange!10}{\color{black}SSRT+\textbf{GH}} &{\color{black}89.2}\\
\rowcolor{orange!20}{\color{black}SSRT+\textbf{GH++}} &{\color{black}\textbf{89.6}}\\
\hline
\end{tabular}}}
\label{tab4}
\end{table}

\begin{table*}
\centering
\setlength{\abovecaptionskip}{0.cm}
\setlength{\belowcaptionskip}{-0.cm}
\caption{Accuracy ($\%$) on DomainNet for UDA (ResNet-101). In each sub-table, the column-wise domains are selected as the source domain and the row-wise domains are selected as the target domain. $^\ast$ indicates that the backbone is ViT pre-trained on ImageNet-1K.}
\setlength{\tabcolsep}{0.5mm}{
\resizebox{\textwidth}{!}{
\begin{tabular}{|c|ccccccc||c|ccccccc||c|ccccccc|}
\hline
{\textbf{ADDA}~\cite{c:28}} &{clp} &{inf} &{pnt} &{qdr} &{rel} &{skt} &{Avg} &
{\textbf{MIMTFL}~\cite{gao2020reducing}} &{clp} &{inf} &{pnt} &{qdr} &{rel} &{skt} &{Avg} &
{\textbf{MDD}~\cite{c:37}} &{clp} &{inf} &{pnt} &{qdr} &{rel} &{skt} &{Avg} \\
\hline
{clp} &{-} &{11.2} &{24.1} &{3.2} &{41.9} &{30.7} &{22.2} &
{clp} &{-} &{15.1} &{35.6} &{10.7} &{51.5} &{43.1} &{31.2} &
{clp} &{-} &{20.5} &{40.7} &{6.2} &{52.5} &{42.1} &{32.4} \\
{inf} &{19.1} &{-} &{16.4} &{3.2} &{26.9} &{14.6} &{16.0} &
{inf} &{32.1} &{-} &{31.0} &{2.9} &{48.5} &{31.0} &{29.1} &
{inf} &{33.0} &{-} &{33.8} &{2.6} &{46.2} &{24.5} &{28.0} \\
{pnt} &{31.2} &{9.5} &{-} &{8.4} &{39.1} &{25.4} &{22.7} &
{pnt} &{40.1} &{14.7} &{-} &{4.2} &{55.4} &{36.8} &{30.2} &
{pnt} &{43.7} &{20.4} &{-} &{2.8} &{51.2} &{41.7} &{32.0} \\
{qdr} &{15.7} &{2.6} &{5.4} &{-} &{9.9} &{11.9} &{9.1} &
{qdr} &{18.8} &{3.1} &{5.0} &{-} &{16.0} &{13.8} &{11.3} &
{qdr} &{18.4} &{3.0} &{8.1} &{-} &{12.9} &{11.8} &{10.8} \\
{rel} &{39.5} &{14.5} &{29.1} &{12.1} &{-} &{25.7} &{24.2} &
{rel} &{48.5} &{19.0} &{47.6} &{5.8} &{-} &{39.4} &{32.1} &
{rel} &{52.8} &{21.6} &{47.8} &{4.2} &{-} &{41.2} &{33.5} \\
{skt} &{35.3} &{8.9} &{25.2} &{14.9} &{37.6} &{-} &{25.4} &
{skt} &{51.7} &{16.5} &{40.3} &{12.3} &{53.5} &{-} &{34.9} &
{skt} &{54.3} &{17.5} &{43.1} &{5.7} &{54.2} &{-} &{35.0} \\
{Avg.} &{28.2} &{9.3} &{20.1} &{8.4} &{31.1} &{21.7} &{\cellcolor{orange!20}19.8} &
{Avg.} &{38.2} &{13.7} &{31.9} &{7.2} &{45.0} &{32.8} &{\cellcolor{orange!20}28.1} &
{Avg.} &{40.4} &{16.6} &{34.7} &{4.3} &{43.4} &{32.3} &{\cellcolor{orange!20}28.6} \\

\hline
{\textbf{CDAN}~\cite{4CDAN}} &{clp} &{inf} &{pnt} &{qdr} &{rel} &{skt} &{Avg.} &
{{\textbf{CDAN+GH}}} &{clp} &{inf} &{pnt} &{qdr} &{rel} &{skt} &{Avg.} &
{\textbf{CDAN+GH++}} &{clp} &{inf} &{pnt} &{qdr} &{rel} &{skt} &{Avg.} \\
\hline
{clp} &{-} &{20.4} &{36.6} &{9.0} &{50.7} &{42.3} &{31.8} &
{clp} &{-} &{20.9} &{40.8} &{11.4} &{57.2} &{45.8} &{35.3} &
{clp} &{-} &{21.4} &{40.3} &{9.4} &{56.3} &{46.4} &{35.5} \\
{inf} &{27.5} &{-} &{25.7} &{1.8} &{34.7} &{20.1} &{22.0} &
{inf} &{31.2} &{-} &{31.2} &{3.7} &{48.2} &{25.9} &{28.0} &
{inf} &{33.4} &{-} &{31.6} &{5.7} &{48.4} &{26} &{29.0} \\
{pnt} &{42.6} &{20.0} &{-} &{2.5} &{55.6} &{38.5} &{31.8} &
{pnt} &{44.3} &{20.0} &{-}  &{2.8} &{57.7} &{39.5} &{32.9} &
{pnt} &{45.1} &{20.2} &{-} &{6.2} &{58.1} &{40.9} &{34.1} \\
{qdr} &{21.0} &{4.5} &{8.1} &{-} &{14.3} &{15.7} &{12.7} &
{qdr} &{24.0} &{4.9} &{10.4} &{-} &{16.9} &{16.6} &{14.6} &
{qdr} &{24.2} &{4.8} &{10.3} &{-} &{16.2} &{16.7} &{14.6} \\
{rel} &{51.9} &{23.3} &{50.4} &{5.4} &{-} &{41.4} &{34.5} &
{rel} &{56.0} &{24.3} &{53.6} &{6.1} &{-} &{42.3} &{36.5} &
{rel} &{55.9} &{24.5} &{53.4} &{6.6} &{-} &{42.8} &{36.6} \\
{skt} &{50.8} &{20.3} &{43.0} &{2.9} &{50.8} &{-} &{33.6} &
{skt} &{56.9} &{21.6} &{46.1} &{11.7} &{55.1} &{-} &{38.3} &
{skt} &{57.4} &{21.5} &{46.3} &{12.3} &{55.9} &{-} &{38.7} \\
{Avg.} &{38.8} &{17.7} &{32.8} &{4.3} &{41.2} &{31.6} &{\cellcolor{orange!20}27.7} &
{Avg.} &{42.5} &{18.3} &{34.6} &{7.1} &{47.0} &{34.1} &{\cellcolor{orange!20}30.9} &
{Avg.} &{43.2} &{18.5} &{36.8} &{8.5} &{47.2} &{34.4} &{\cellcolor{orange!20}\textbf{31.4}} \\

\hline

{\textbf{MCD}~\cite{c:2MCD}} &{clp} &{inf} &{pnt} &{qdr} &{rel} &{skt} &{Avg.} &
{{\textbf{MCD+GH}}} &{clp} &{inf} &{pnt} &{qdr} &{rel} &{skt} &{Avg.} &
{\textbf{MCD+GH++}} &{clp} &{inf} &{pnt} &{qdr} &{rel} &{skt} &{Avg.} \\
\hline

{clp} &{-} &{14.2} &{26.1} &{1.6} &{45.0} &{33.8} &{24.1} &
{clp} &{-} &{18.3} &{37.2} &{8.4} &{52.1} &{43.6} &{31.9} &
{clp} &{-} &{19.6} &{37.2} &{8.5} &{52.2} &{44.3} &{32.4} \\
{inf} &{23.6} &{-} &{21.2} &{1.5} &{36.7} &{18.0} &{20.2} &
{inf} &{34.0} &{-} &{33.3} &{2.8} &{46.7} &{29.9} &{29.3} &
{inf} &{33.7} &{-} &{33.8} &{3.1} &{47.1} &{30.0} &{29.5} \\
{pnt} &{34.4} &{14.8} &{-} &{1.9} &{50.5} &{28.4} &{26.0} &
{pnt} &{44.3} &{18.9} &{-}  &{3.7} &{54.1} &{40.7} &{32.3} &
{pnt} &{44.0} &{19.4} &{-} &{3.9} &{53.9} &{40.9} &{32.3} \\
{qdr} &{15.0} &{3.0} &{7.0} &{-} &{11.5} &{10.2} &{9.3} &
{qdr} &{20.2} &{3.1} &{8.4} &{-} &{14.5} &{13.0} &{11.8} &
{qdr} &{20.9} &{3.6} &{8.6} &{-} &{15.1} &{13.5} &{12.3} \\
{rel} &{42.6} &{19.6} &{42.6} &{2.2} &{-} &{29.3} &{27.2} &
{rel} &{51.3} &{21.5} &{51.1} &{2.4} &{-} &{39.3} &{33.0} &
{rel} &{50.9} &{22.0} &{50.6} &{2.9} &{-} &{39.4} &{33.6} \\
{skt} &{41.2} &{13.7} &{27.6} &{3.8} &{34.8} &{-} &{24.2} &
{skt} &{54.4} &{19.1} &{44.1} &{8.6} &{51.1} &{-} &{35.5} &
{skt} &{54.5} &{19.2} &{43.8} &{7.8} &{51.3} &{-} &{35.5} \\
{Avg.} &{31.4} &{13.1} &{24.9} &{2.2} &{35.7} &{23.9} &{\cellcolor{orange!20}21.9} &
{Avg.} &{40.8} &{16.2} &{34.8} &{5.1} &{43.7} &{33.3} &{\cellcolor{orange!20}29.0} &
{Avg.} &{41.0} &{16.7} &{35.2} &{5.1} &{43.9} &{33.5} &{\cellcolor{orange!20}\textbf{29.2}} \\

\hline

{\textbf{DWL}~\cite{c:5DWL}} &{clp} &{inf} &{pnt} &{qdr} &{rel} &{skt} &{Avg.} &
{{\textbf{DWL+GH}}} &{clp} &{inf} &{pnt} &{qdr} &{rel} &{skt} &{Avg.} &
{\textbf{DWL+GH++}} &{clp} &{inf} &{pnt} &{qdr} &{rel} &{skt} &{Avg.} \\
\hline
{clp} &{-} &{17.0} &{29.5} &{9.3} &{48.5} &{37.2} &{28.3} &
{clp} &{-} &{17.3} &{31.3} &{10.2} &{49.3} &{38.2} &{29.3} &
{clp} &{-} &{17.4} &{31.3} &{11.4} &{49.2} &{38.7} &{29.6} \\
{inf} &{24.2} &{-} &{22.1} &{2.5} &{38.7} &{23.0} &{22.1} &
{inf} &{24.4} &{-} &{22.5} &{3.0} &{38.9} &{23.7} &{22.5} &
{inf} &{24.4} &{-} &{22.4} &{3.3} &{39.2} &{23.8} &{22.6} \\
{pnt} &{31.2} &{14.0} &{-} &{3.9} &{46.4} &{29.5} &{25.0} &
{pnt} &{31.8} &{14.7} &{-} &{5.4} &{46.7} &{31.0} &{25.9} &
{pnt} &{32.2} &{15.1} &{-} &{5.7} &{46.9} &{31.1} &{26.2} \\
{qdr} &{17.9} &{4.5} &{10.1} &{-} &{16.4} &{16.4} &{13.1} &
{qdr} &{18.7} &{4.6} &{10.3} &{-} &{17.9} &{16.6} &{13.6} &
{qdr} &{19.0} &{4.8} &{10.3} &{-} &{17.6} &{16.5} &{13.6} \\
{rel} &{43.4} &{20.5} &{41.4} &{5.3} &{-} &{33.8} &{28.9} &
{rel} &{43.6} &{21.9} &{41.7} &{5.4} &{-} &{34.5} &{29.4} &
{rel} &{44.7} &{21.5} &{42.0} &{5.6} &{-} &{34.7} &{29.7} \\
{skt} &{45.4} &{15.1} &{32.1} &{7.3} &{44.4} &{-} &{28.9} &
{skt} &{46.6} &{16.0} &{33.6} &{8.8} &{45.7} &{-} &{30.1} &
{skt} &{47.1} &{17.5} &{34.9} &{8.9} &{45.3} &{-} &{30.7} \\
{Avg.} &{32.4} &{14.2} &{27.0} &{5.7} &{38.9} &{28.0} &{\cellcolor{orange!20}24.4} &
{Avg.} &{33.0} &{14.9} &{27.9} &{6.6} &{39.7} &{28.8} &{\cellcolor{orange!20}25.1} &
{Avg.} &{33.5} &{15.3} &{28.2} &{7.0} &{39.6} &{29.0} &{\cellcolor{orange!20}25.4} \\

\hline

{\textbf{GVB}~\cite{c:3GVB}} &{clp} &{inf} &{pnt} &{qdr} &{rel} &{skt} &{Avg.} &
{\textbf{GVB+GH}} &{clp} &{inf} &{pnt} &{qdr} &{rel} &{skt} &{Avg.} &
{\textbf{GVB+GH++}} &{clp} &{inf} &{pnt} &{qdr} &{rel} &{skt} &{Avg.} \\
\hline
{clp} &{-} &{18.4} &{41.5} &{11.9} &{56.0} &{45.2} &{34.6} &
{clp} &{-} &{19.2} &{41.7} &{12.4} &{57.4} &{46.2} &{35.4} &
{clp} &{-} &{19.2} &{42.0} &{12.8} &{57.3} &{46.1} &{35.5} \\
{inf} &{33.5} &{-} &{36.4} &{2.3} &{47.0} &{27.2} &{29.3} &
{inf} &{35.4} &{-} &{39.6} &{3.1} &{49.8} &{29.8} &{31.5} &
{inf} &{36.4} &{-} &{39.6} &{3.5} &{50.0} &{29.2} &{31.7} \\
{pnt} &{46.0} &{19.0} &{-} &{5.1} &{57.4} &{41.8} &{33.9} &
{pnt} &{47.7} &{19.5} &{-} &{5.1} &{58.9} &{43.5} &{34.9} &
{pnt} &{47.9} &{19.6} &{-} &{5.6} &{59.4} &{43.7} &{35.2} \\
{qdr} &{31.2} &{3.0} &{14.4} &{-} &{22.1} &{21.6} &{18.5} &
{qdr} &{31.4} &{3.0} &{14.4} &{-} &{22.7} &{22.0} &{18.7} &
{qdr} &{32.0} &{3.3} &{15.2} &{-} &{24.0} &{22.3} &{19.4} \\
{rel} &{56.0} &{22.7} &{54.0} &{4.4} &{-} &{44.7} &{36.4} &
{rel} &{57.7} &{23.4} &{54.7} &{4.6} &{-} &{44.9} &{37.1} &
{rel} &{58.4} &{23.3} &{55.1} &{5.3} &{-} &{45.3} &{37.5} \\
{skt} &{57.6} &{18.7} &{48.5} &{11.1} &{55.9} &{-} &{38.4} &
{skt} &{57.6} &{19.5} &{48.5} &{12.6} &{56.9} &{-} &{39.0} &
{skt} &{57.9} &{19.7} &{48.6} &{12.1} &{57.2} &{-} &{39.1} \\
{Avg.} &{44.9} &{16.4} &{39.0} &{7.0} &{47.7} &{36.1} &{\cellcolor{orange!20}31.8} &
{Avg.} &{46.0} &{16.9} &{39.8} &{7.6} &{49.1} &{37.3} &{\cellcolor{orange!20}32.8} &
{Avg.} &{46.5} &{17.0} &{40.1} &{7.9} &{49.6} &{37.3} &{\cellcolor{orange!20}\textbf{33.1}} \\

\hline

{\textbf{SSRT$^\ast$}~\cite{sun2022safe}} &{clp} &{inf} &{pnt} &{qdr} &{rel} &{skt} &{Avg.} &
{\textbf{SSRT$^\ast$+GH}} &{clp} &{inf} &{pnt} &{qdr} &{rel} &{skt} &{Avg.} &
{\textbf{SSRT$^\ast$+GH++}} &{clp} &{inf} &{pnt} &{qdr} &{rel} &{skt} &{Avg.} \\
\hline
{clp} &{-} &{33.8} &{60.2} &{19.4} &{75.8} &{59.8} &{49.8} &
{clp} &{-} &{35.7} &{60.9} &{20.9} &{75.9} &{61.0} &{50.9} &
{clp} &{-} &{35.8} &{60.9} &{22.0} &{76.0} &{61.6} &{51.3} \\
{inf} &{55.5} &{-} &{54.0} &{9.0} &{68.2} &{44.7} &{46.3} &
{inf} &{55.9} &{-} &{55.1} &{12.2} &{68.4} &{50.2} &{48.4} &
{inf} &{55.7} &{-} &{55.2} &{11.7} &{69.0} &{49.6} &{48.2} \\
{pnt} &{61.7} &{28.5} &{-} &{8.4} &{71.4} &{55.2} &{45.0} &
{pnt} &{62.1} &{30.9} &{-} &{10.0} &{71.7} &{56.5} &{46.2} &
{pnt} &{61.7} &{33.2} &{-} &{10.5} &{71.6} &{56.3} &{46.7} \\
{qdr} &{42.5} &{8.8} &{24.2} &{-} &{37.6} &{33.6} &{29.3} &
{qdr} &{43.9} &{11.5} &{24.2} &{-} &{37.7} &{34.4} &{30.3} &
{qdr} &{43.3} &{11.8} &{24.5} &{-} &{38.9} &{35.1} &{30.7} \\
{rel} &{69.9} &{37.1} &{66.0} &{10.1} &{-} &{58.9} &{48.4} &
{rel} &{70.3} &{39.6} &{66.2} &{10.4} &{-} &{59.7} &{49.2} &
{rel} &{70.0} &{39.8} &{66.0} &{10.8} &{-} &{61.6} &{49.6} \\
{skt} &{70.6} &{32.8} &{62.2} &{21.7} &{73.2} &{-} &{52.1} &
{skt} &{71.0} &{34.7} &{62.9} &{22.8} &{73.4} &{-} &{53.0} &
{skt} &{71.6} &{35.1} &{62.5} &{23.2} &{73.4} &{-} &{53.2} \\
{Avg.} &{60.0} &{28.2} &{53.3} &{13.7} &{65.2} &{50.4} &{\cellcolor{orange!20}45.2} &
{Avg.} &{60.6} &{30.5} &{53.9} &{15.3} &{65.4} &{52.4} &{\cellcolor{orange!20}46.3} &
{Avg.} &{60.5} &{31.1} &{53.8} &{15.6} &{65.8} &{52.8} &{\cellcolor{orange!20}\textbf{46.6}} \\

\hline

{\textbf{SSRT}~\cite{sun2022safe}} &{clp} &{inf} &{pnt} &{qdr} &{rel} &{skt} &{Avg.} &
{\textbf{SSRT+GH}} &{clp} &{inf} &{pnt} &{qdr} &{rel} &{skt} &{Avg.} &
{\textbf{SSRT+GH++}} &{clp} &{inf} &{pnt} &{qdr} &{rel} &{skt} &{Avg.} \\
\hline
{clp} &{-} &{33.9} &{60.1} &{19.4} &{75.7} &{61.0} &{50.0} &
{clp} &{-} &{35.1} &{60.7} &{21.3} &{75.9} &{61.1} &{50.8} &
{clp} &{-} &{35.8} &{60.9} &{22.9} &{76.0} &{61.8} &{51.5} \\
{inf} &{55.7} &{-} &{54.9} &{9.3} &{68.2} &{44.8} &{46.6} &
{inf} &{56.3} &{-} &{55.6} &{11.3} &{68.5} &{50.2} &{48.4} &
{inf} &{56.0} &{-} &{55.2} &{11.4} &{69.1} &{49.8} &{48.3} \\
{pnt} &{62.0} &{31.5} &{-} &{9.3} &{71.2} &{55.5} &{45.9} &
{pnt} &{62.3} &{31.7} &{-} &{10.5} &{71.5} &{56.7} &{46.5} &
{pnt} &{62.3} &{33.3} &{-} &{10.7} &{71.6} &{56.4} &{46.9} \\
{qdr} &{42.6} &{12.6} &{22.2} &{-} &{37.0} &{34.3} &{29.7} &
{qdr} &{44.0} &{13.0} &{23.1} &{-} &{37.4} &{34.6} &{30.4} &
{qdr} &{43.5} &{13.7} &{23.3} &{-} &{38.7} &{35.3} &{30.9} \\
{rel} &{70.6} &{37.0} &{66.3} &{10.3} &{-} &{59.4} &{48.7} &
{rel} &{71.0} &{39.5} &{66.5} &{10.7} &{-} &{59.9} &{49.5} &
{rel} &{70.8} &{39.8} &{66.3} &{10.8} &{-} &{61.5} &{49.8} \\
{skt} &{70.7} &{31.9} &{62.4} &{21.9} &{73.2} &{-} &{52.0} &
{skt} &{71.0} &{33.9} &{63.0} &{22.9} &{73.5} &{-} &{52.9} &
{skt} &{71.7} &{34.8} &{62.6} &{23.4} &{73.4} &{-} &{53.2} \\
{Avg.} &{60.3} &{29.4} &{53.2} &{14.0} &{65.1} &{51.0} &{\cellcolor{orange!20}45.5} &
{Avg.} &{60.9} &{30.6} &{53.8} &{15.3} &{65.4} &{52.5} &{\cellcolor{orange!20}46.4} &
{Avg.} &{60.9} &{31.5} &{53.7} &{15.8} &{65.8} &{53.0} &{\cellcolor{orange!20}\textbf{46.8}} \\
\hline

\end{tabular}}}
\label{tab5}
\end{table*}

\section{Model Analysis and Discussion}\label{Model analysis}
\subsection{Feature Visualization}  
\begin{figure*}[t]
 \centering
\setlength{\abovecaptionskip}{0.1cm}
\setlength{\belowcaptionskip}{-0.2cm}
  \begin{minipage}{4.4cm}
 \centerline{\includegraphics[scale=0.2]{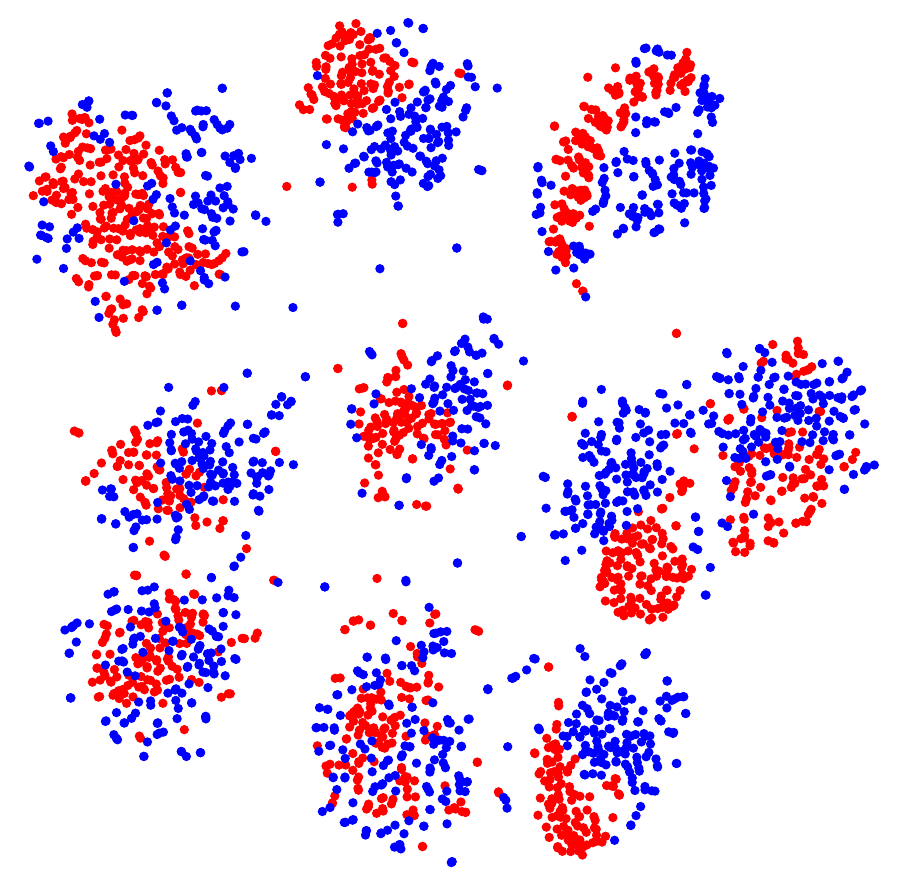}}
 \centerline{(a) MCD on U $\rightarrow$ M}
\end{minipage}
\begin{minipage}{4.4cm}
 \centerline{\includegraphics[scale=0.2]{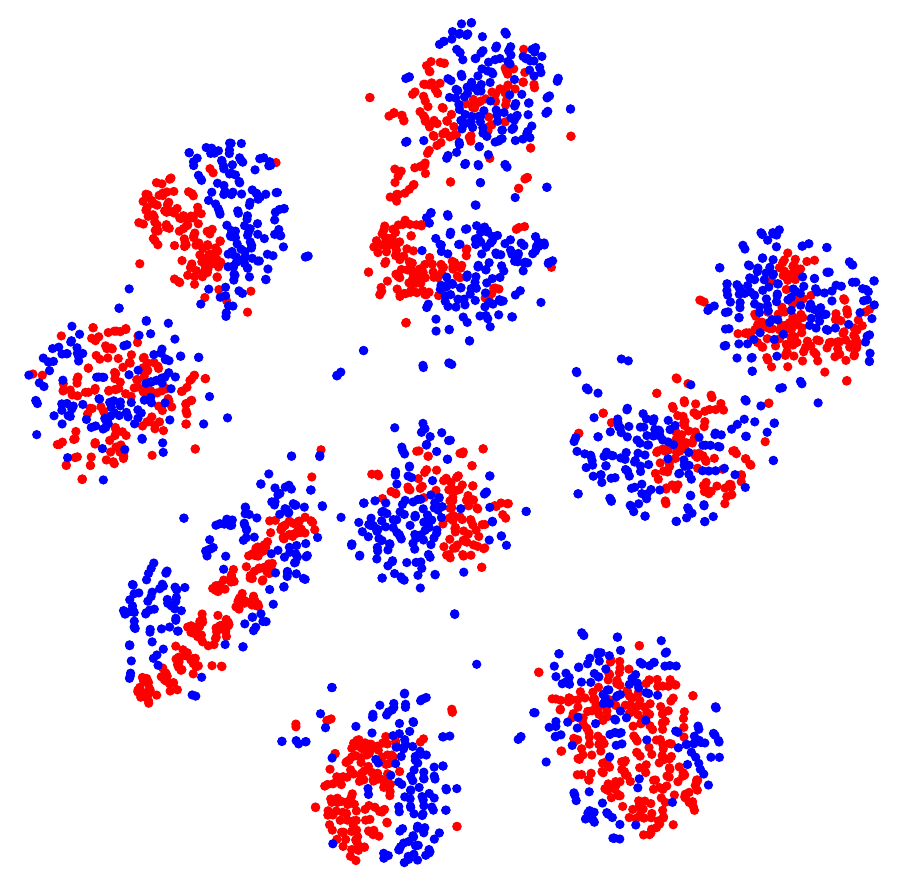}}
 \centerline{(b) MCD+GH on U $\rightarrow$ M }
\end{minipage}
\begin{minipage}{4.4cm}
 \centerline{\includegraphics[scale=0.2]{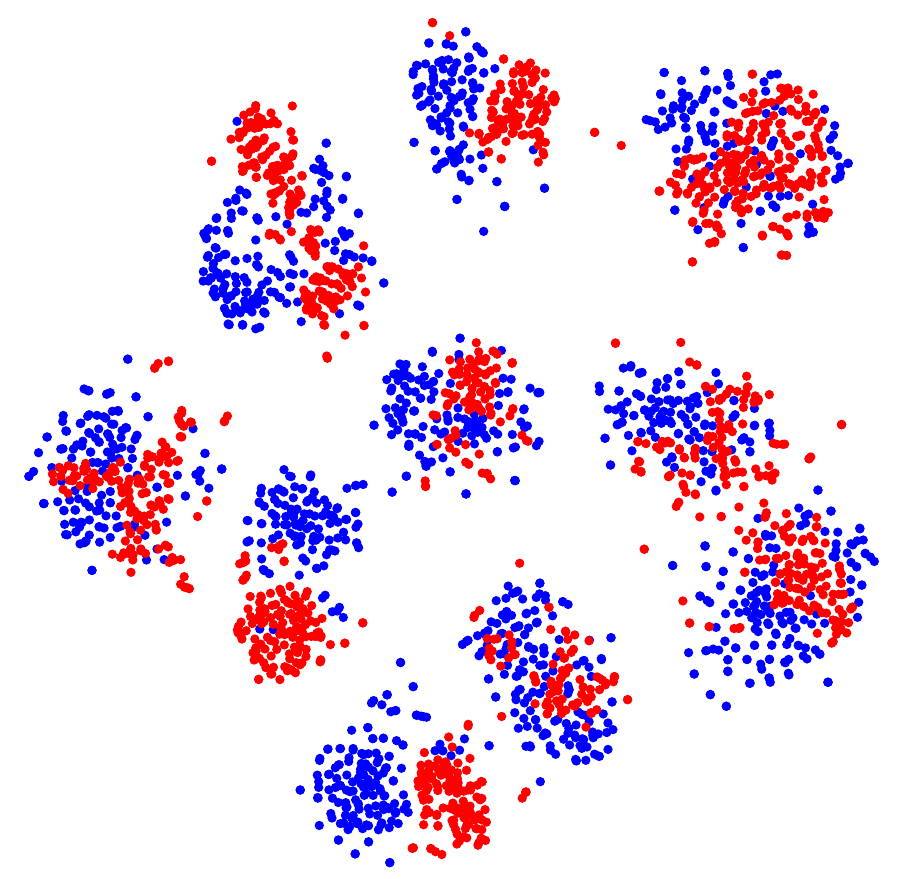}}
 \centerline{(c) MCD on M $\rightarrow$ U}
\end{minipage}
\begin{minipage}{4.4cm}
 \centerline{\includegraphics[scale=0.2]{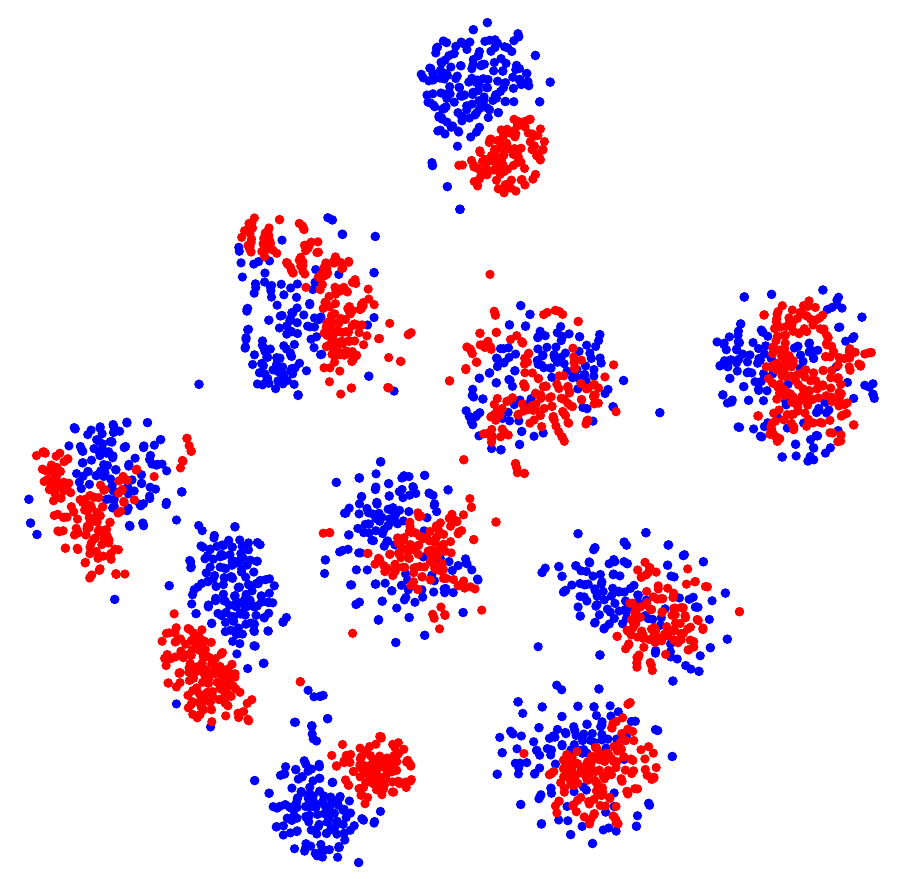}}
 \centerline{(d) MCD+GH on M $\rightarrow$ U}
\end{minipage}
\hfill
  \caption{The t-SNE visualizations of features generated by MCD and MCD+GH. (a) and (b) are feature visualizations on U $\rightarrow$ M. (c) and (d) are feature visualizations on M $\rightarrow$ U. Red and blue points indicate the source and target samples, respectively.}
  \label{fig5}
\end{figure*}

Fig. \ref{fig5} describes the t-SNE~\cite{r:20} visualizations of features learned by MCD (baseline) and MCD+GH on the tasks of U $\rightarrow$ M and M $\rightarrow$ U. Fig. \ref{fig5} (a) and (c) are visualization features generated by MCD. Fig. \ref{fig5} (b) and (d) are visualization features generated by MCD+GH. It can be observed that both features learned by MCD and MCD+GH achieve well-performed global alignment effect with 10 clusters under two tasks. Further, the visualization feature distributions with GH deployed have better clustering effect and have fewer samples distributed across class boundaries, which intuitively boosts the feature discriminability. In addition, visualization results further validate the balance learning ability of our GH approach.

\subsection{Convergence Analysis}  
\begin{figure*}
 \centering
\setlength{\abovecaptionskip}{0.1cm}
\setlength{\belowcaptionskip}{0.1cm}
  \begin{minipage}{4.2cm}
 \centerline{\includegraphics[scale=0.295]{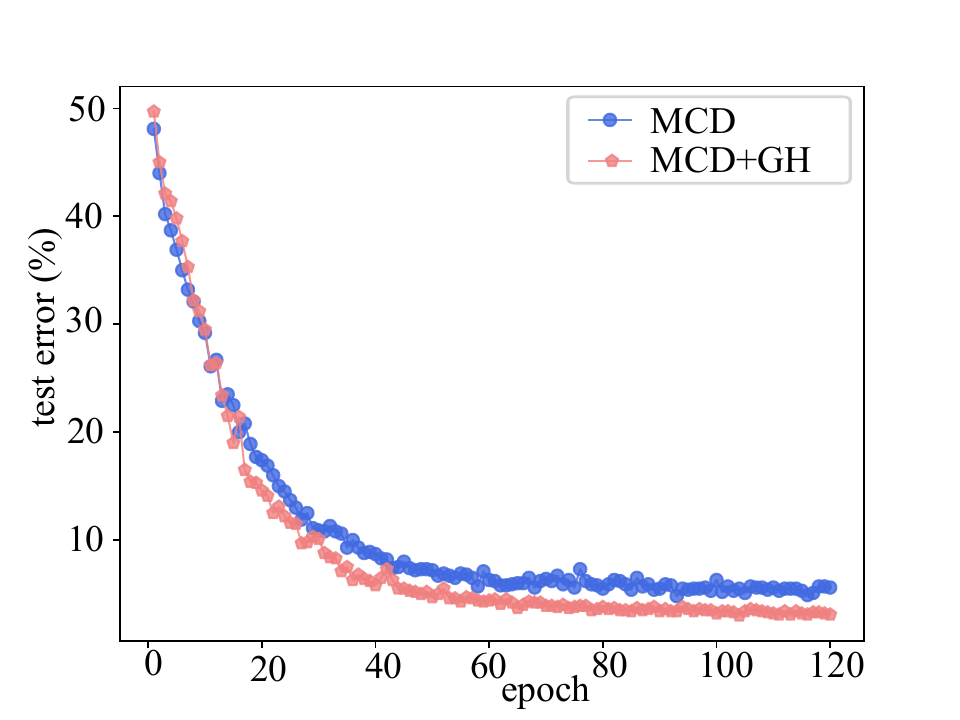}}
 \centerline{(a) U $\rightarrow$ M}
\end{minipage}
\begin{minipage}{4.2cm}
 \centerline{\includegraphics[scale=0.295]{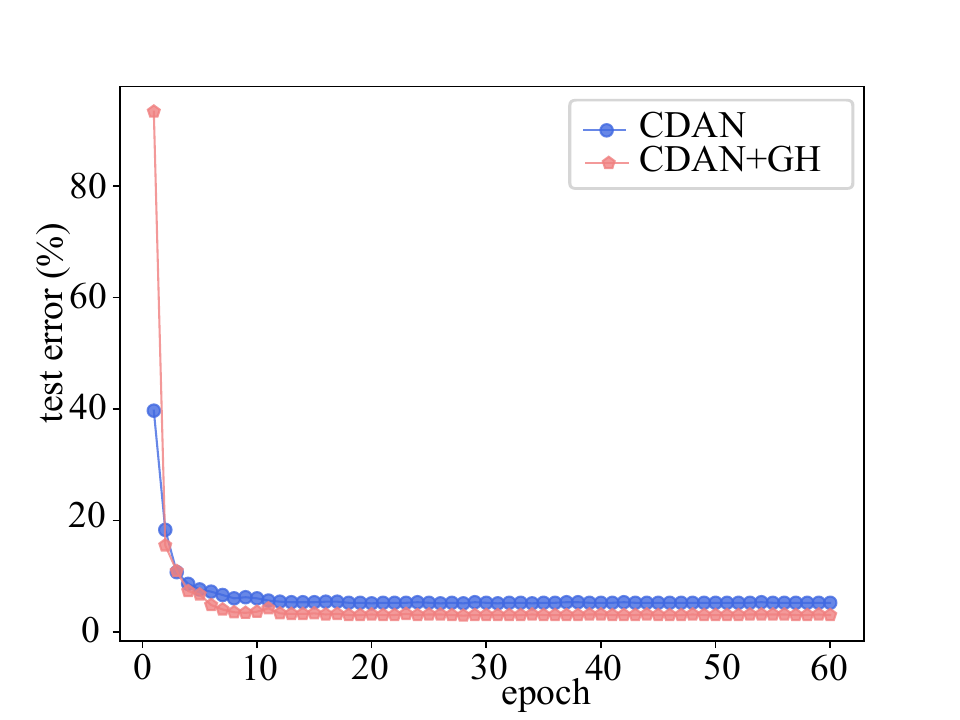}}
 \centerline{(b) U $\rightarrow$ M}
\end{minipage}
\begin{minipage}{4.2cm}
 \centerline{\includegraphics[scale=0.295]{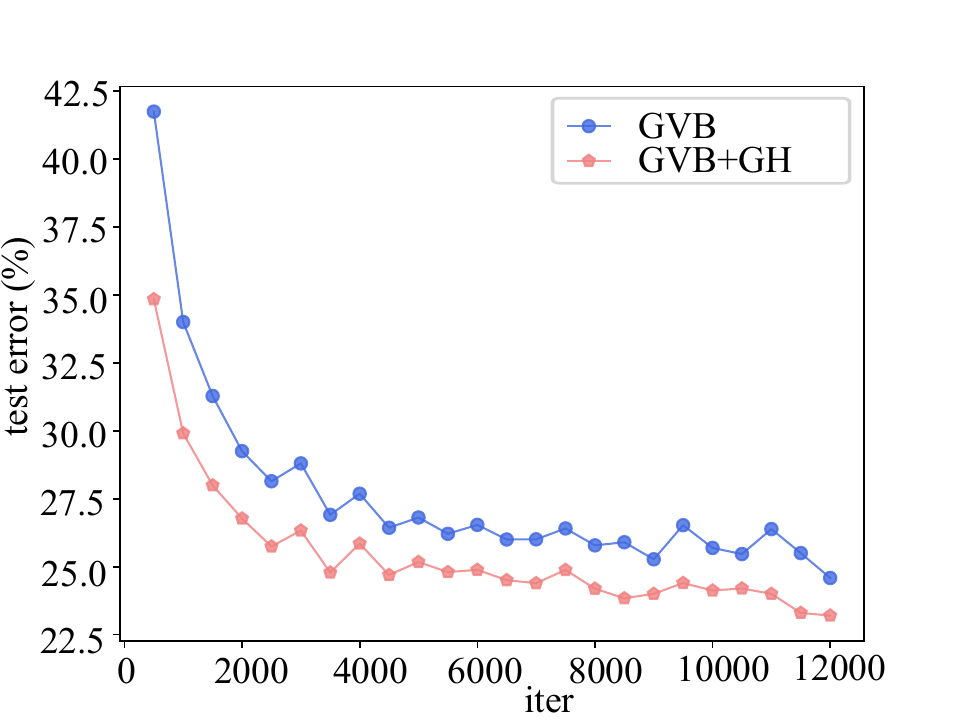}}
 \centerline{(c) Synthetic $\rightarrow$ Real}
\end{minipage}
\begin{minipage}{4.2cm}
 \centerline{\includegraphics[scale=0.295]{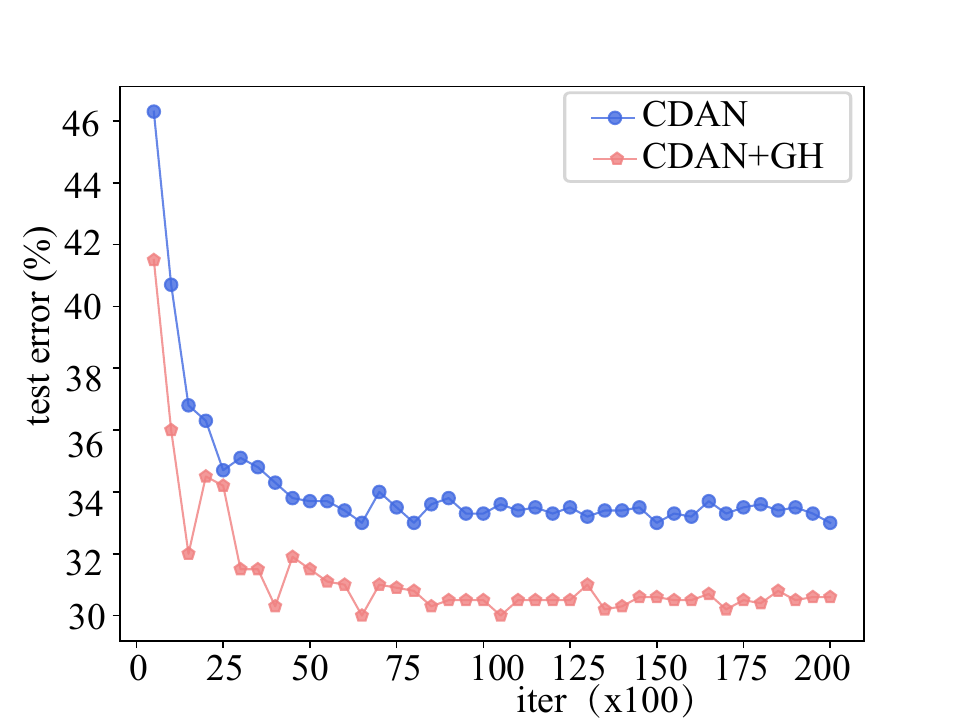}}
 \centerline{(d) Synthetic $\rightarrow$ Real}
\end{minipage}
\hfill
  \caption{Convergence curves of various baselines and baseline+GH on the test error (\%). Clearly, the baselines are improved by GH.}
  \label{fig6}
\end{figure*}

We present the convergence curves of test error with respect to the number of iterations on tasks of U $\rightarrow$ M and Synthetic $\rightarrow$ Real as shown in Fig. \ref{fig6}. For each subfigure, the blue line represents the test error of different baselines, and the red line represents the test error for baseline+GH (e.g., CDAN+GH). Obviously, compared with baselines, the introduction of GH can further improve the test performance and convergence. This fully indicates that GH plays an active coordination role in practical optimization process, which promotes the cooperation of domain alignment task and classification task towards a benign direction as expected. 

\begin{figure*}
 \centering
\setlength{\abovecaptionskip}{0.1cm}
\setlength{\belowcaptionskip}{-0.2cm}
  \begin{minipage}{4.4cm}
 \centerline{\includegraphics[scale=0.16]{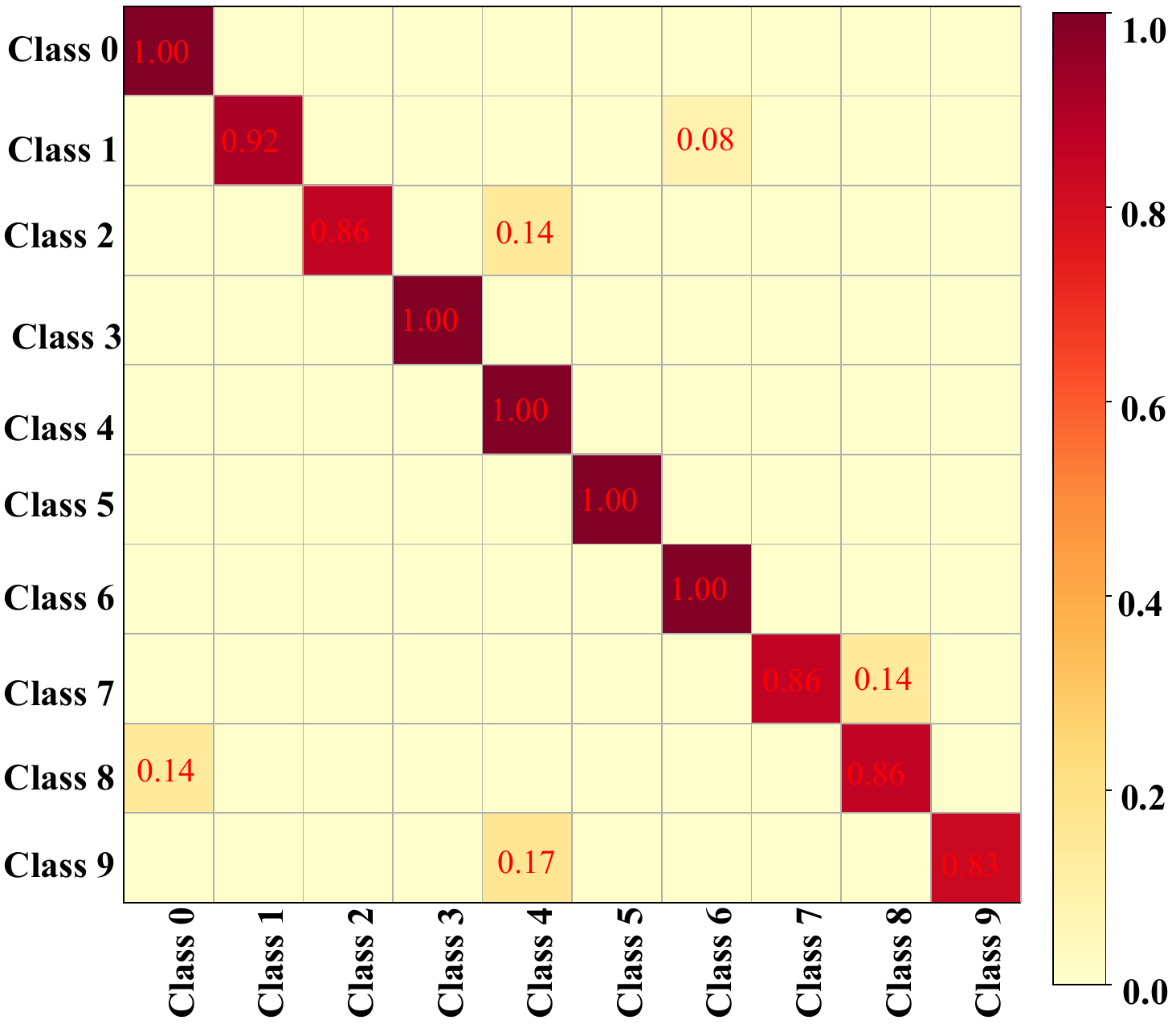}}
 \centerline{(a) DWL (M$\rightarrow$U)}
\end{minipage}
\begin{minipage}{4.4cm}
 \centerline{\includegraphics[scale=0.16]{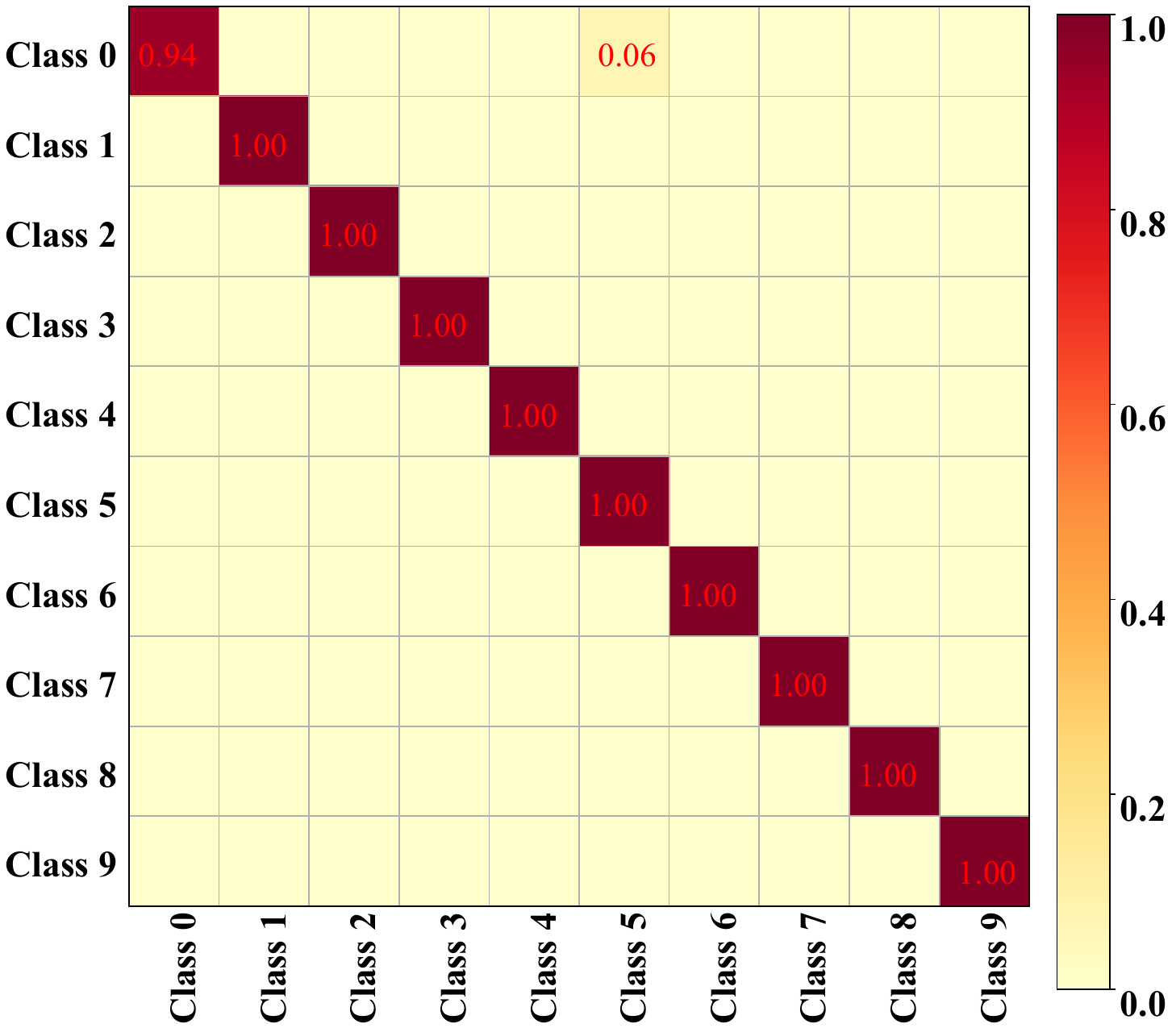}}
 \centerline{(b) DWL+GH (M$\rightarrow$U)}
\end{minipage}
\begin{minipage}{4.4cm}
 \centerline{\includegraphics[scale=0.16]{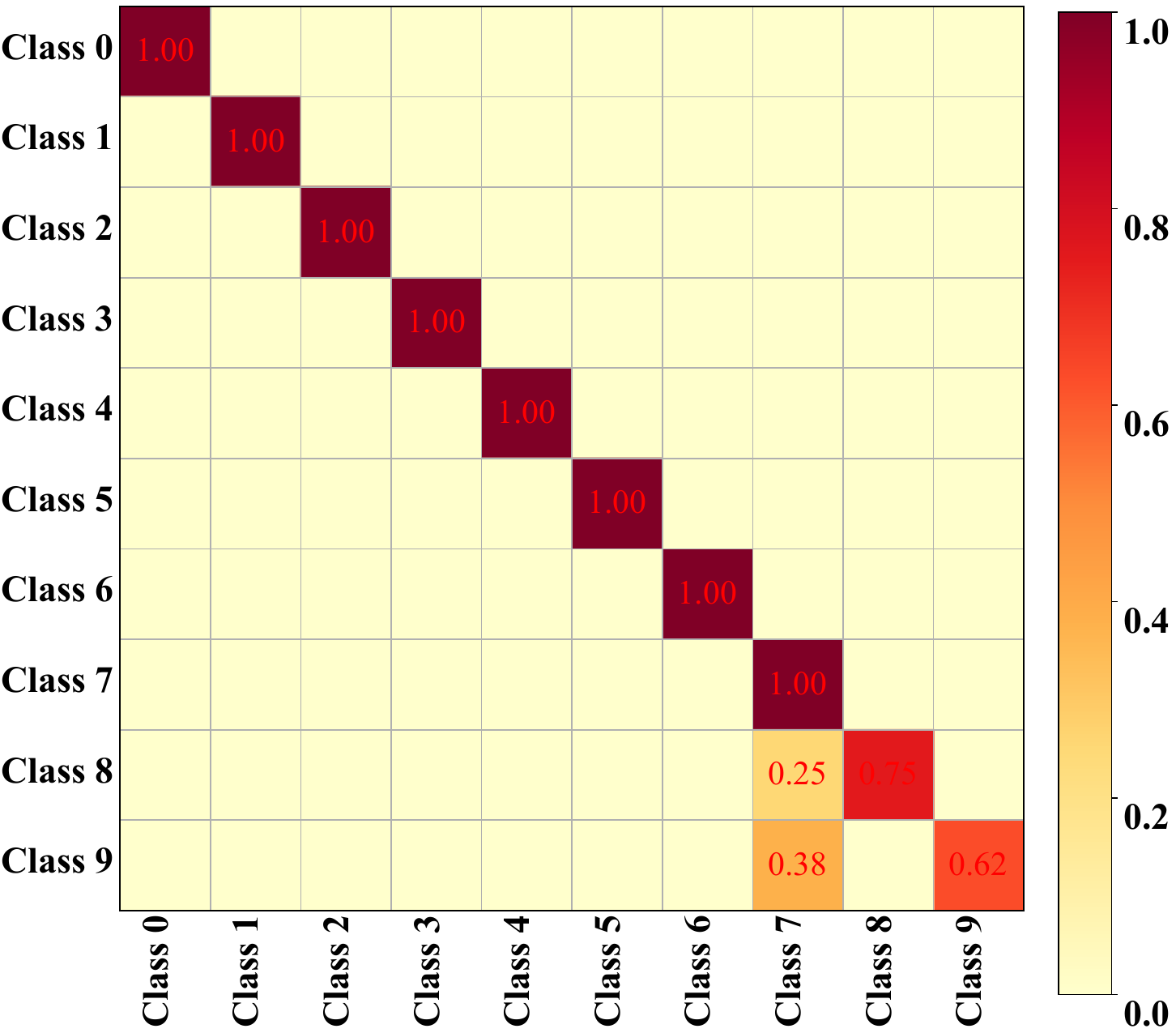}}
 \centerline{(c) DWL (U$\rightarrow$M)}
\end{minipage}
\begin{minipage}{4.4cm}
 \centerline{\includegraphics[scale=0.16]{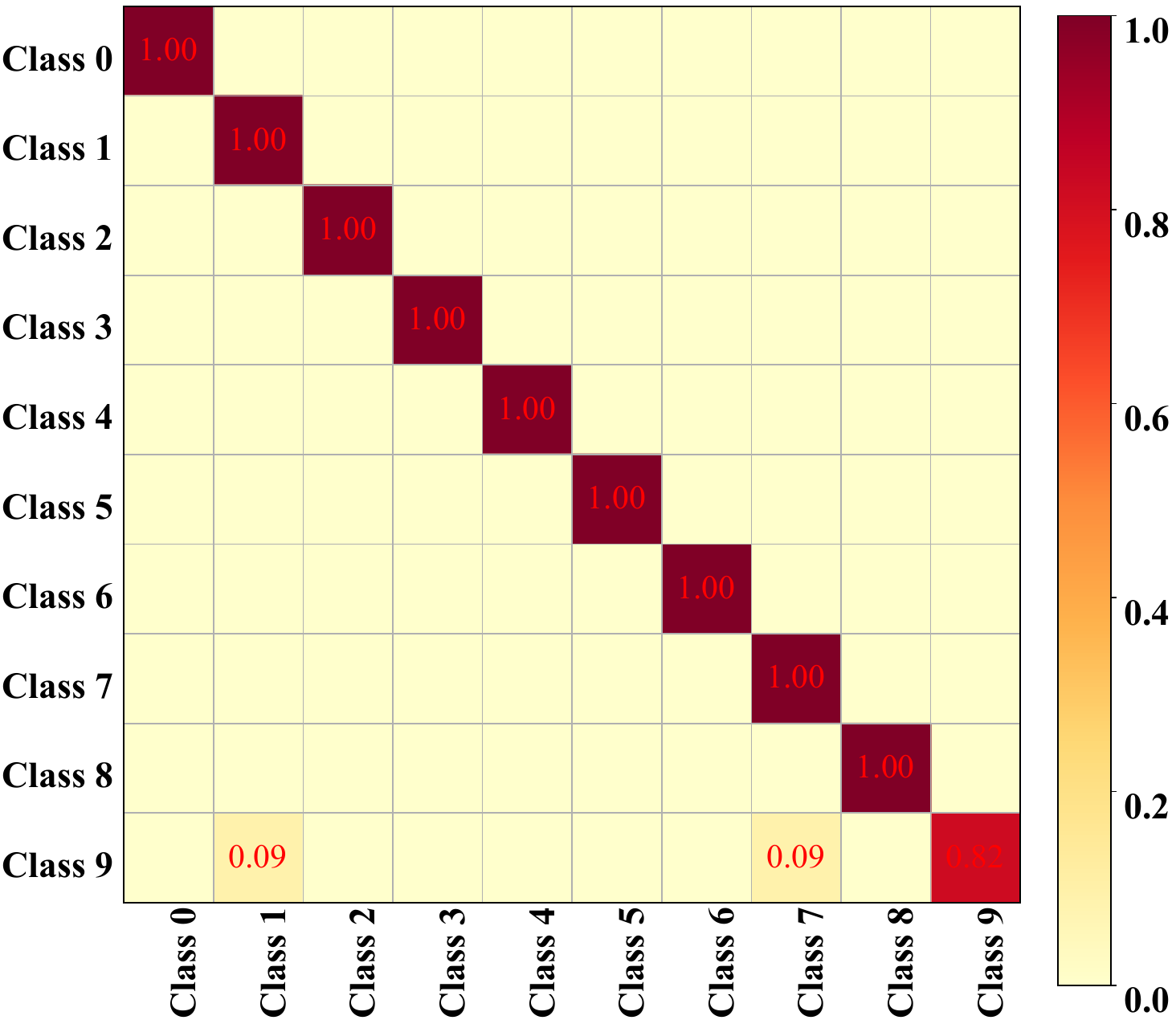}}
 \centerline{(d) DWL+GH (U$\rightarrow$M)}
\end{minipage}
\hfill
  \caption{Confusion matrix visualization Results before and after applying GH (i.e., DWL vs. DWL+GH) on different classification tasks.}
  \label{fig7}
\end{figure*}

\begin{figure}[t]
 \centering
\setlength{\abovecaptionskip}{0.1cm}
\setlength{\belowcaptionskip}{-0.1cm}
  \begin{minipage}{4.2cm}
 \centerline{\includegraphics[scale=0.28]{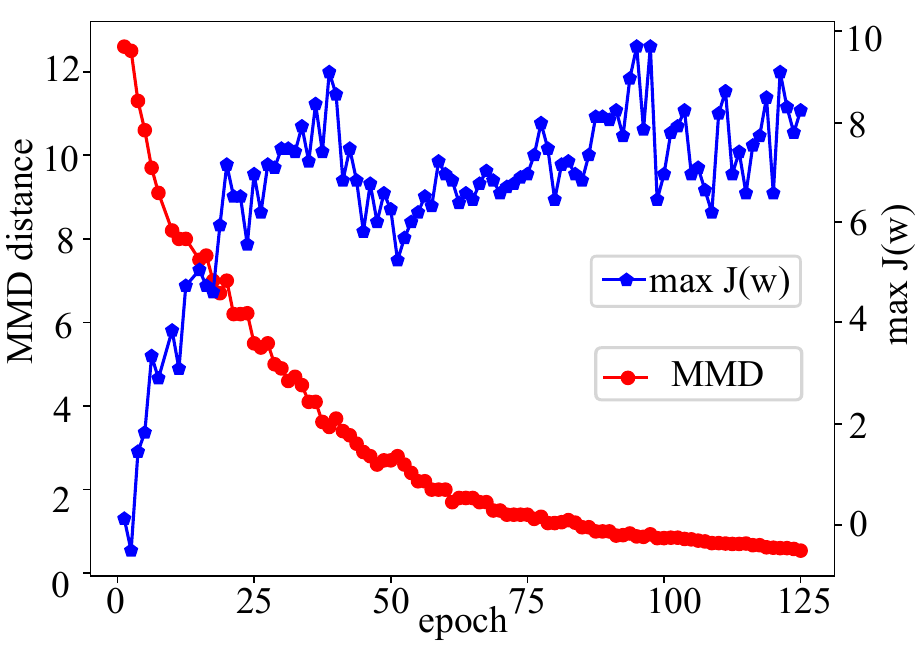}}
 \centerline{(a) M $\rightarrow$ U}
\end{minipage}
\begin{minipage}{4.2cm}
 \centerline{\includegraphics[scale=0.28]{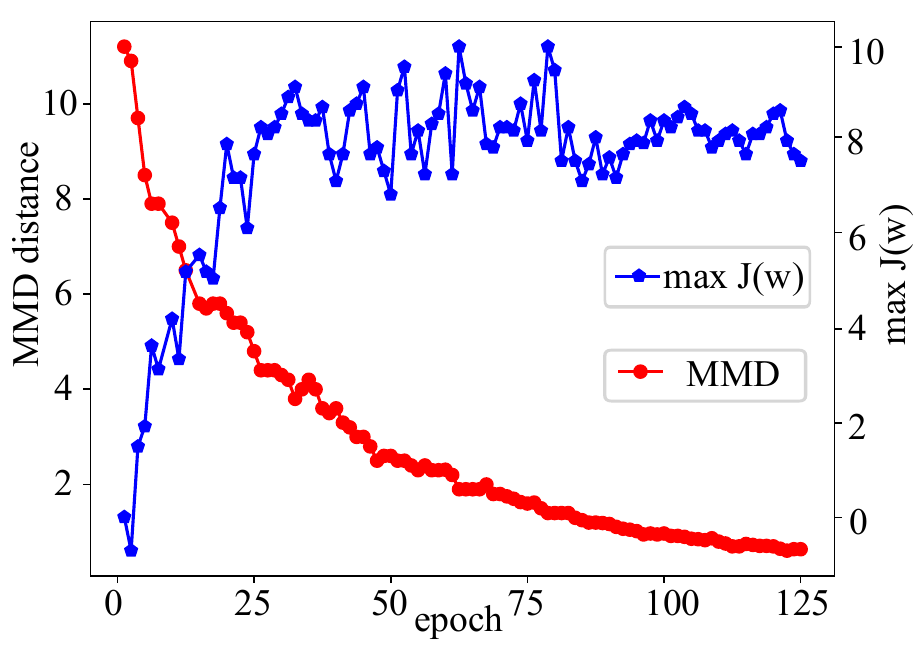}}
 \centerline{(b) U $\rightarrow$ M}
\end{minipage}
\hfill
  \caption{Between-task balance analysis for domain alignment (MMD)~\cite{r:38} and class discrimination (max J(w))~\cite{c:25} of UDA after gradient harmonization. MCD is used in this experiment (i.e., MCD+GH).}
  \label{fig77}
\end{figure}

\begin{figure}[t]
 \centering

  \begin{minipage}{4.4cm}
 \centerline{\includegraphics[scale=0.28]{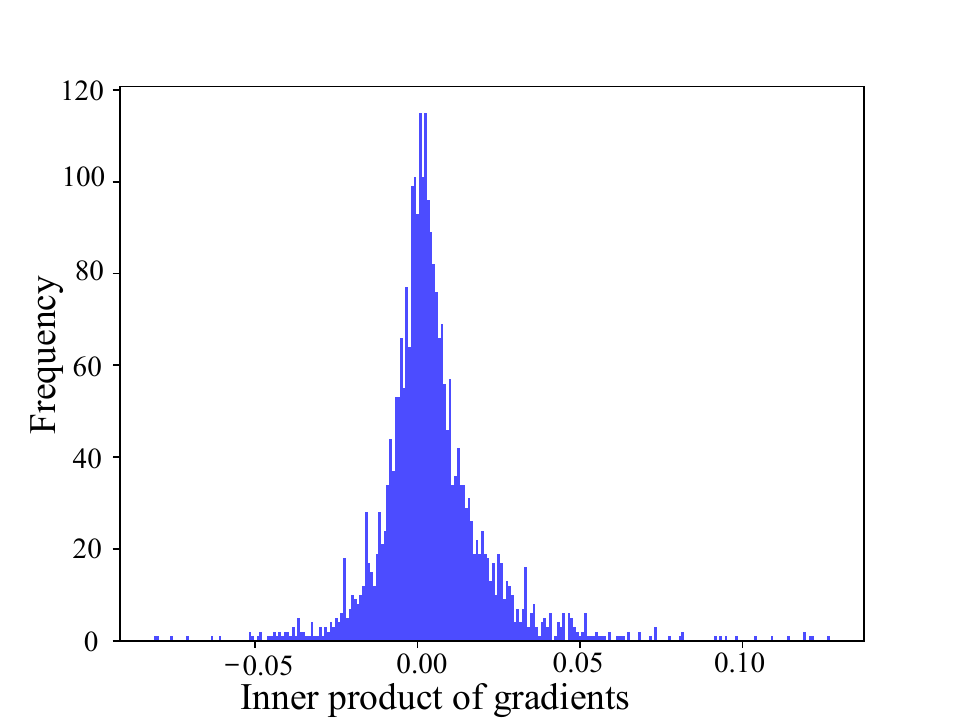}}
 \centerline{(a) Before GH}
\end{minipage}
\begin{minipage}{4.4cm}
 \centerline{\includegraphics[scale=0.28]{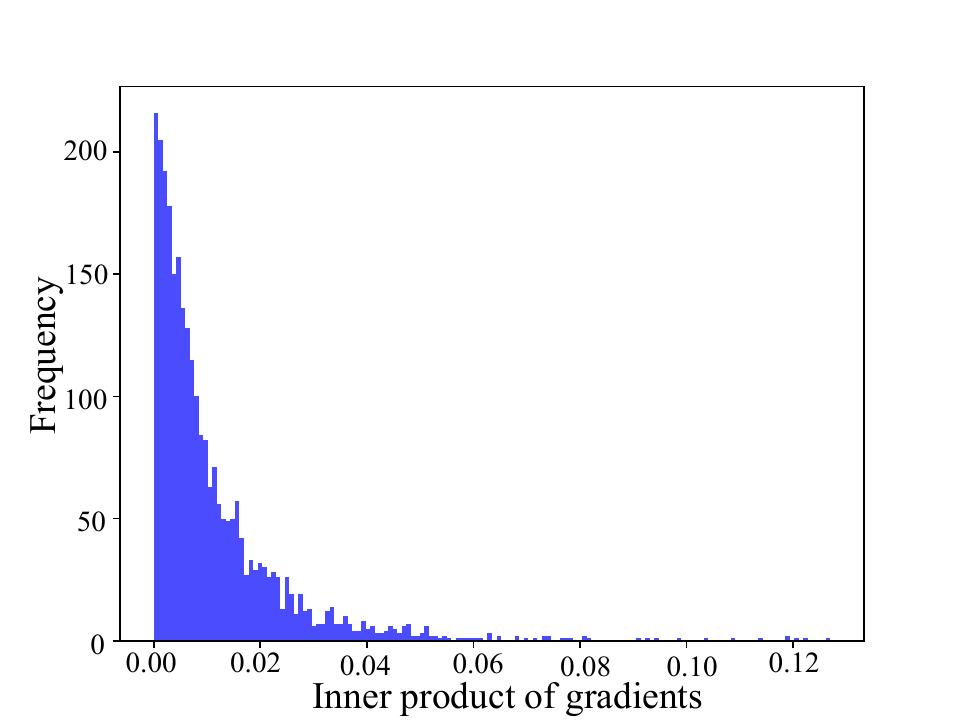}}
 \centerline{(b) After GH}
\end{minipage}
\hfill
\vspace{-4mm}
  \caption{Inner product distributions (histogram) of the two task-specific gradients before and after GH on U $\rightarrow$ M. Clearly, the between-task gradient conflict is eliminated after harmonization.}
  \label{figbaGH}
\end{figure}

\subsection{Confusion Matrix Visualization}

Fig. \ref{fig7} displays the visualizations of confusion matrix for the classifier trained by DWL and DWL+GH. DWL obtains several uncertain predictions with small values while DWL+GH obtains more confident predictions. Comparing with Fig. \ref{fig7} (a) and (b), the confusing ``Class 1, 2, 7, 8, and 9'' are correctly recognized in DWL+GH. From Fig. \ref{fig7} (c) and (d), the confusing ``Class 8'' is corrected in DWL+GH. It implies that optimization conflict during training deteriorates the discriminability in the target domain. Our method preserves the feature discriminability of target samples in the course of harmonic training. DWL+GH improves the between-task balance by coordinating the explicit external task conflict and the implicit internal optimization conflict.

\begin{table}
\renewcommand\arraystretch{1.5}
\centering
\setlength{\abovecaptionskip}{0.cm}
\setlength{\belowcaptionskip}{-0.2cm}
\caption{Analyses about the Training Speed (s/epoch) and Classification Accuracy ($\%$) before and after applying GH.}
\setlength{\tabcolsep}{1.5mm}
\resizebox{\linewidth}{!}{
\begin{tabular}{|c|c|c|c|c|c|c|}
\hline
\multirow{2}{*}{Task} &\multicolumn{2}{c|}{MCD} &\multicolumn{2}{c|}{MCD+GH} &\multicolumn{2}{c|}{Discrepancy}\\
\cline{2-7}
&Times &Acc. &Times &Acc. &$\vartriangle$Times &$\vartriangle$Acc. \\
\hline
M $\rightarrow$ U &0.78540 &94.2 &0.92794 &96.7 &0.14254 &2.5 \\
U $\rightarrow$ M  &0.89765 &94.1 &0.99574 &96.8 &0.09809 &2.7 \\
\hline
\end{tabular}}
\label{tabtraining speed}
\end{table}

\subsection{Balance Analysis of GH-based UDA}  Fig. \ref{fig77} shows MMD distance~\cite{r:38} and the max J(W)~\cite{c:25} values based on the feature representation learned by MCD+GH. The left vertical axis corresponding to the red curve represents the MMD distance used to measure the alignment degree across domains. The smaller value means the better domain alignment. The right vertical axis corresponding to the blue curve represents the degree of discriminability based on Linear Discriminant Analysis (LDA)~\cite{40fukunaga}. The larger max J(W) implies better discriminability, which directly affects the classification result of samples. Fig. \ref{fig77}(a) and (b) reflect the degree of domain alignment and discriminability on task of M$\rightarrow$U and U$\rightarrow$M, respectively. From Fig. \ref{fig77}, MCD+GH has a smaller MMD distance and a larger max J(W) value during training. This shows that GH facilitates the coordinated optimality of alignment and classification tasks while maintaining their task-specific optimality. Enhanced and balanced domain-invariant and class-discriminative feature representations can be obtained.

\subsection{Training Speed and Accuracy} In order to observe the efficiency of the proposed equivalent model more clearly, we present the training speed and classification accuracy before and after applying GH. As shown in Table \ref{tabtraining speed}, for tasks M$\rightarrow$U and U$\rightarrow$M on digits datasets, the training speed of MCD+GH is 0.14254 s/epoch and 0.09809 s/epoch longer than MCD, respectively. In other words, MCD+GH takes less training time than MCD in training process, but can get 2.5\%/2.7\% classification accuracy gain. The computational cost of employing GH is quite low, and thus GH is a powerful and efficient auxiliary tool to facilitate those popular domain adaptation baselines towards more outstanding classification performance.

\subsection{Gradient Inner Product Visualization}  Fig. \ref{figbaGH} presents inner product distributions of two gradients between domain alignment and classification tasks before and after applying Gradient Harmonization for MCD. From Fig. \ref{figbaGH} (a), we observe the acute and obtuse angles between gradients of the two tasks before coordination. The obtuse angles account for about 40\% of the total number, which exhibits the identical property as Fig. \ref{figfig}. In other words, there exists between-task conflict in the model optimization process but paid less attention. After Gradient Harmonization, as shown in Fig. \ref{figbaGH} (b), the inner products of the two gradients are all positive. That is, the gradient angles between the two tasks are coordinated into acute angles. The proposed GH avoids the optimization conflict by separately adjusting the gradients of the two tasks to achieve the purpose of optimal coordination. Experimental results fully illustrate the effectiveness of the proposed GH.

\begin{figure}[t]
 \centering
\includegraphics[width=0.45\textwidth]{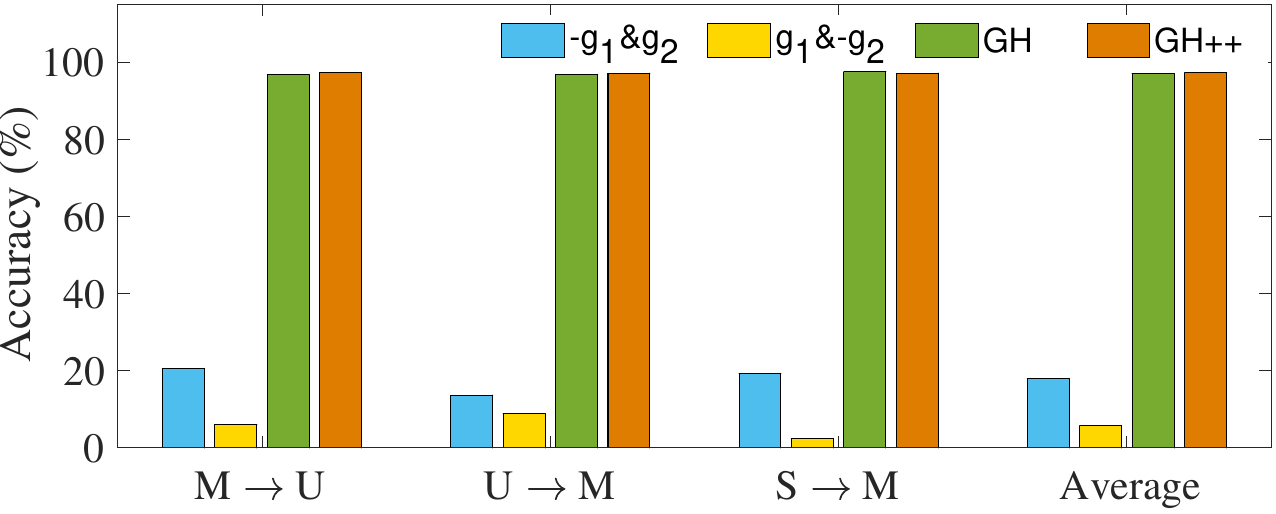}
\setlength{\abovecaptionskip}{0.cm}
\setlength{\belowcaptionskip}{-0.3cm}
\caption{Recognition accuracy ($\%$) on Digits for different gradient harmonization strategies. 
``$-g_1\&g_2$'' represents multiplying a minus sign to $g_1$ and keep $g_2$ unchanged. ``$g_1\&-g_2$'' represents multiplying a minus sign to $g_2$ and keep $g_1$ unchanged. 
  }
  \label{diffgh}
\end{figure}

\subsection{Rationality and Comparison to Other Alternatives}

The proposed GH/GH++ aims at altering the gradient angle between two different tasks from an obtuse angle to an acute/vertical angle. 
Whether the same effect can be achieved with other intuitive gradient correction alternatives or not? For instance, one might consider convert an obtuse angle into an acute angle by simply applying a negative sign to one of the two gradients, i.e., transforming either $g_1$ into $-g_1$ or $g_2$ into $-g_2$. As depicted in Fig. \ref{diffgh}, it is apparent that the results reveal a significant decline in classification accuracy when using the $-g_1\&g_2$ and $g_1\&-g_2$ methods in contrast to the proposed approaches. While applying a negative sign can indeed transit from an obtuse angle to an acute angle and force the two gradients to be positively correlated, it cannot replicate the excellent classification performance achieved by the proposed GH/GH++.

The rationale behind this observation is that directly applying a negative sign to one of the gradients alters the gradient towards the opposite direction of the original optimal gradient descent direction. This significantly undermines the primary objective of the original task. In other words, the performance deteriorates sharply when the gradient harmonization direction significantly deviates from the original direction. GH/GH++ reasonably adjusts the two gradients to achieve the purpose of optimization coordination on the premise of keeping the harmonic gradient as close as possible to the original gradient descent direction. Therefore, the proposed approaches do not break the task-specific optimality, but pursues a better cooperation, which verifies the rationality of the proposed GH/GH++.


\begin{figure}
 \centering
  \begin{minipage}{4.3cm}
 \centerline{\includegraphics[scale=0.28]{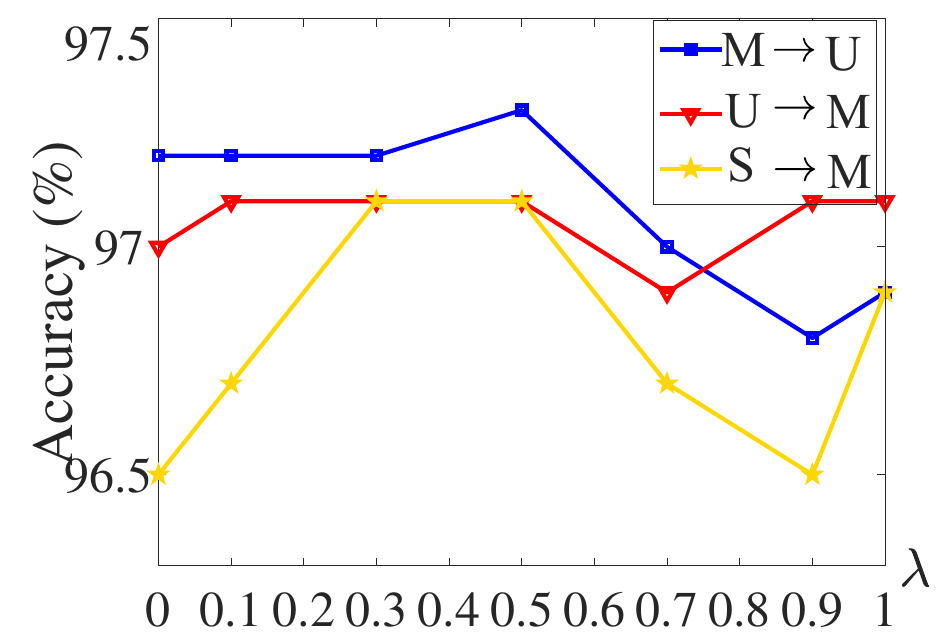}}
 \centerline{(a) MCD}
\end{minipage}
\begin{minipage}{4.3cm}
 \centerline{\includegraphics[scale=0.28]{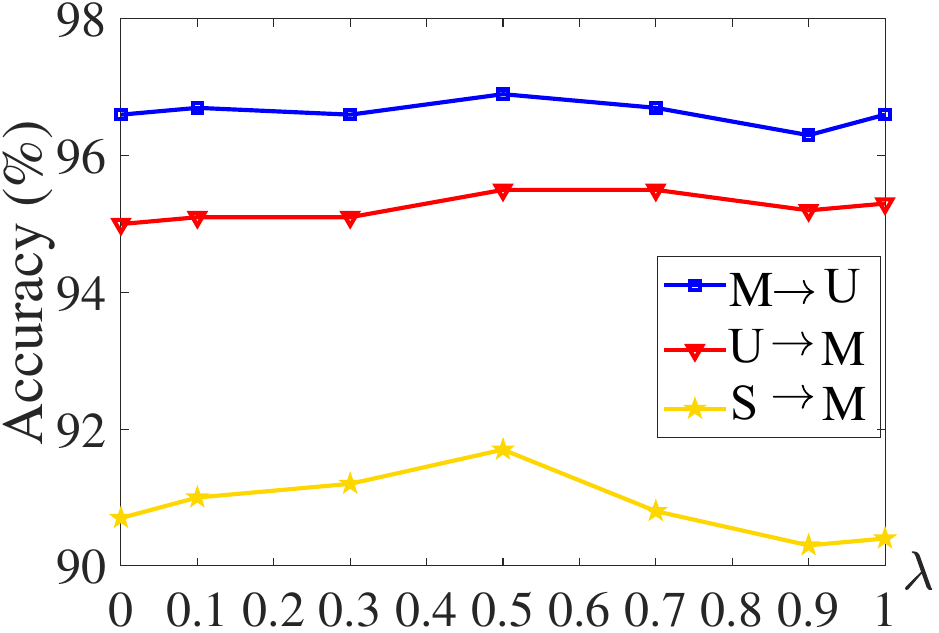}}
 \centerline{(b) GVB }
\end{minipage}
\hfill
\setlength{\abovecaptionskip}{0.1cm}
\setlength{\belowcaptionskip}{-0.1cm}
  \caption{Sensitivity analysis of the parameter $\lambda$ in GH++. MCD and GVB are tested as UDA baselines.}
  \label{figparam}
\end{figure}

\subsection{Parameter Sensitivity Analysis}
To investigate the effect of the parameter $\lambda$ in GH++, we conduct experiments on three tasks (i.e., M$\rightarrow$U, U$\rightarrow$M and S$\rightarrow$M) based on MCD and GVB by varying $\lambda\in\{0, 0.1, 0.3, 0.5,0.7, 0.9, 1\}$. The results are presented in Fig. \ref{figparam}. We can observe GH++ is little sensitive to the scale variety of $\lambda$, which indicates that GH++ is robust across different baselines and tasks. Besides, we observe that when $\lambda=0.5$, models generally achieve the best performance. In other words, when the gradient deviation of the two tasks is relieved, the performance can be largely improved.


\subsection{Scalability to Other Multi-task Problems}
To further demonstrate the universality and scalability of the proposed approaches, we evaluate GH/GH++ in the object detection and multi-modal interactive retrieval fields, which also involves optimization of multiple objectives.


\textbf{Dataset.} We select widely used benchmarks, i.e., \textbf{PASCAL VOC 2007} \cite{everingham2007pascal} and \textbf{CSS} \cite{vo2019composing} for object detection and multi-modal interactive retrieval, respectively.
\textbf{PASCAL VOC 2007} consists of about 5K trainval images and 5K test images over 20 object categories.
\textbf{CSS} consists of 38K synthesized images with different colors, shapes and sizes. It contains about 19K training image-text pairs and 19K testing image-text pairs, respectively. 

\textbf{Implementation Details.} For object detection, we use the DSSD \cite{fu2017dssd} model with $321\times 321$ inputs as the baseline. We follow the same experimental settings and protocol as \cite{wang2021reconcile} and adopt Average Precision (averaged AP at IoUs from 0.5 to 0.9) to measure performance. For multi-modal interactive retrieval, we use TIRG \cite{vo2019composing} and the adversarial training version(AT-TIRG) follows the paper \cite{huang2022adversarial} as the baseline. For evaluation, we follow the same protocols as \cite{huang2022adversarial, huang2023language, vo2019composing} and use retrieval accuracy R@N as our evaluation metric, which computes the percentage of test queries where at least one target or correctly matched image is within the top N retrieved images. Note that we exactly follow the original experimental setups of the baselines. In other words, \emph{DSSD+GH/GH++} and \emph{AT-TIRG+GH/GH++} indicate that GH/GH++ are inserted directly into the DSSD and AT-TIRG, respectively, without any change in either the backbone or hyper-parameters.
\begin{table}[t]
\centering
\setlength{\abovecaptionskip}{0.cm}
\setlength{\belowcaptionskip}{-0.2cm}
\caption{Object detection on PASCAL VOC 2007 test set.}
\setlength{\tabcolsep}{3mm}
  \renewcommand\arraystretch{1.2}
  \resizebox{\linewidth}{!}{
\begin{tabular}{|l|c|c|c|c|c|c|}
\hline
Method &$AP$ &$AP_{50}$ &$AP_{60}$  &$AP_{70}$  &$AP_{80}$  &$AP_{90}$ \\
\hline
DSSD \cite{fu2017dssd} &52.4 &80.0 &75.2 &65.0 &46.3 &16.3 \\
\rowcolor{orange!10}DSSD+GH &{53.0} &{80.1}  &{75.5}  &\textbf{65.5}  &{47.2}  &\textbf{17.3}\\
\rowcolor{orange!20}DSSD+GH++&\textbf{53.1}&\textbf{80.4}&\textbf{75.8}&{65.4}&\textbf{47.6}&\textbf{17.3}\\
\hline

\end{tabular}}
\label{OD}
\end{table}

\begin{table}[t]
\centering
\setlength{\abovecaptionskip}{0.cm}
\setlength{\belowcaptionskip}{-0.2cm}
\caption{Adversarial training (AT) for multi-modal interactive retrieval on CSS.}
\setlength{\tabcolsep}{0.4mm}
  \renewcommand\arraystretch{1.3}
\resizebox{\linewidth}{!}{
\begin{tabular}{|l|cc|cc|cc|cc|cc|}
\hline
\multirow{2}{*}{Method} &\multicolumn{2}{c|}{R@1} &\multicolumn{2}{c|}{R@5} &\multicolumn{2}{c|}{R@10}  &\multicolumn{2}{c|}{R@50}  &\multicolumn{2}{c|}{R@100} \\
\cline{2-11}
&clean &Adv. &clean &Adv. &clean &Adv. &clean &Adv. &clean &Adv. \\
\hline
TIRG &78.8 &1.8 &94.9 &5.9 &97.3 &8.6 &99.1 &20.2 &99.5 &27.9  \\
\hline
{AT-TIRG \cite{huang2022adversarial}} &80.0 &50.4 &96.2 &77.1 &{98.1}&84.8 &{99.6}&95.9 &99.7 &97.9  \\
\hline
\rowcolor{orange!10}{AT-TIRG+GH}  &{80.6} &{53.8}&{96.3}&{80.0} &97.9 &{87.0} &99.5 &{96.4}& {99.8 } & {98.2}\\
\hline
\rowcolor{orange!20}AT-TIRG+GH++&\textbf{81.0}&\textbf{55.6}&\textbf{96.4}&\textbf{82.0}&\textbf{98.2}&\textbf{88.7}&\textbf{99.7}&\textbf{97.0}&\textbf{99.9}&\textbf{98.5}\\
\hline
\end{tabular}}
\label{AT}
\end{table}

\textbf{Results on Object Detection.} Object detection models \cite{fu2017dssd, wang2021reconcile} usually consider two tasks: classification and regression. In this section, we use the proposed GH/GH++ to mitigate the conflict between these two tasks. Table \ref{OD} presents the Average Precision (AP) on the PASCAL VOC 2007 test set. We can observe that GH/GH++ can improve the AP with different bounding boxes, which verifies the effectiveness of the proposed approaches in object detection.


\textbf{Results on Multi-modal Interactive Retrieval.} Recently, \cite{huang2022adversarial} introduces adversarial training \cite{chaturvedi2020mimic, xie2020adversarial} into multi-modal interactive retrieval model (abbreviated as AT-TIRG) to improve model robustness, which aims to train a generalized and robust model suitable for both clean samples and adversarial attack samples. During adversarial training, there is conflict in training the clean samples and adversarial attack samples, which results in the limitation of the final performance. In this section, we try to adopt the proposed approaches to solve this problem. The experimental results are provided in Table \ref{AT}. Comparing \emph{AT-TIRG+GH} with \emph{AT-TIRG}, we can see the proposed GH can enhance the AT. For the adversarial attack samples, \emph{AT-TIRG+GH} outperforms the \emph{AT-TIRG} by 3.4\%, 2.9\%, and 2.2\% on R@1, R@5, and R@10, respectively. \emph{AT-TIRG+GH++} outperforms the \emph{AT-TIRG} by 5.2\%, 4.9\%, and 3.9\% on R@1, R@5, and R@10, respectively. The performance of the clean samples is also improved in most cases. Note that the performance of clean samples is usually close to 100\%, so there is little space for improvement.


\section{Conclusion}
In this paper, we pay attention to the optimization conflict (i.e., imbalance or incoordination) problem between different tasks (i.e., the alignment task and the classification task) in alignment-based unsupervised domain adaptation models. To mitigate this problem, we propose two simple yet efficient Gradient Harmonization approaches, including GH and GH++, which take measures to de-conflict between the gradients of both tasks in optimization. Besides, to facilitate the harmonization during adaptation, we derive the equivalent but more efficient model of \textbf{UDA with GH/GH++}, which becomes a dynamically reweighted loss function of most existing unsupervised domain adaptation models. Further, the essence and insights of the proposed approaches are provided to indicate its rationality. Exhaustive experiments and model analyses demonstrate that the proposed approaches significantly improve the existing UDA models and contribute to achieving state-of-the-art results.
In addition, we have verified that the proposed approaches can be adapted to other problems and areas, such as object detection and multi-modal retrieval, to de-conflict between the gradients of any two tasks in optimization and improve model performance.


%

\appendices


\ifCLASSOPTIONcompsoc

\ifCLASSOPTIONcaptionsoff
  \newpage
\fi



%

\bibliographystyle{IEEEtrans}
\bibliography{IEEEexample}
\end{document}